\documentclass{article} 
\usepackage{iclr2025_conference,times}

\usepackage{amsmath,amsfonts,bm}

\def\eqref#1{equation~\ref{#1}}

\def\1{\bm{1}}

\DeclareMathAlphabet{\mathsfit}{\encodingdefault}{\sfdefault}{m}{sl}
\SetMathAlphabet{\mathsfit}{bold}{\encodingdefault}{\sfdefault}{bx}{n}

\usepackage{hyperref}
\usepackage{url}
\usepackage{enumitem}
\usepackage{subcaption}

\usepackage[pdftex]{graphicx}
\usepackage[utf8]{inputenc} 
\usepackage[T1]{fontenc}    
\usepackage{booktabs}       
\usepackage{tablefootnote} 
\usepackage{makecell}
\usepackage{microtype}
\usepackage{graphicx}
\usepackage{booktabs} %
\usepackage{multirow}
\usepackage{spverbatim} %
\usepackage{adjustbox}
\usepackage{tabularx}
\usepackage{ragged2e}
\usepackage{array}
\usepackage{hyperref}
\usepackage{url}

\usepackage{amsmath}
\usepackage{amssymb}
\usepackage{mathtools}
\usepackage{amsthm}

\usepackage{enumitem}

\usepackage{placeins}
\usepackage{tikz}
\usepackage{bbding}

\usepackage[capitalize,noabbrev]{cleveref}

\usepackage[many]{tcolorbox}
\usepackage{longtable}

\theoremstyle{plain}

\theoremstyle{definition}

\theoremstyle{remark}

\providecommand{\danqi}[1]{
    {\protect\color{purple}{}}
}
\providecommand{\alex}[1]{
    {\protect\color{blue}{}}
}
\providecommand{\aatmik}[1]{
    {\protect\color{orange}{}}
}
\providecommand{\saumya}[1]{
    {\protect\color{orange}{}}
}

\renewcommand{\paragraph}[1]{\vspace{0.2cm}\noindent\textbf{#1}}

\newcommand{\ourmethod}{DataMan}
\newcommand{\ourdata}{DataPajama}

\newtcbox{\hlprimarytab}{on line, rounded corners, box align=base, colback=green!10,colframe=white,size=fbox,arc=3pt, before upper=\strut, top=-2pt, bottom=-4pt, left=-2pt, right=-2pt, boxrule=0pt}
\newtcbox{\hlsecondarytab}{on line, box align=base, colback=red!10,colframe=white,size=fbox,arc=3pt, before upper=\strut, top=-2pt, bottom=-4pt, left=-2pt, right=-2pt, boxrule=0pt}
\newcommand{\dashifted}{\raisebox{0.5\depth}{\tiny$\downarrow$}}
\newcommand{\uashifted}{\raisebox{0.5\depth}{\tiny$\uparrow$}}

\newcommand{\gda}[1]{{\raisebox{0.6ex}{\tiny\hlprimarytab{\dashifted{#1}}}}}   
\newcommand{\gua}[1]{{\raisebox{0.6ex}{\tiny\hlprimarytab{\uashifted{#1}}}}}   
\newcommand{\rda}[1]{{\raisebox{0.6ex}{\tiny\hlsecondarytab{\uashifted{#1}}}}} 
\newcommand{\rua}[1]{{\raisebox{0.6ex}{\tiny\hlsecondarytab{\dashifted{#1}}}}} 
\setlist[itemize]{left=5pt} 

\title{DataMan: Data Manager for Pre-training Large Language Models}

\author{
  Ru Peng$^{1}$\thanks{This work done during internship at Qwen team, Alibaba Group.} \ \ \ 
  Kexin Yang$^{2}$ \ \ 
  Yawen Zeng$^{1}$ \ \ 
  Junyang Lin$^{2}$ \ \ 
  Dayiheng Liu$^{2}$\thanks{Both are corresponding authors.} \ \ \ 
  Junbo Zhao$^{1}$\footnotemark[2] \\[1pt]
  \makebox[\textwidth][c]{
    $^1$Zhejiang University \quad $^2$Alibaba Group
  } \\[1pt]
  \makebox[\textwidth][c]{
    \texttt{\small \{rupeng,j.zhao\}@zju.edu.cn \quad liudayiheng.ldyh@alibaba-inc.com}
  } \\
}

\iclrfinalcopy 
\begin{document}
\maketitle

\begin{abstract}
The performance emergence of large language models (LLMs) driven by data scaling laws makes the selection of pre-training data increasingly important. 
However, existing methods rely on limited heuristics and human intuition, lacking comprehensive and clear guidelines.
To address this, we are inspired by \emph{``reverse thinking''} -- prompting LLMs to self-identify which criteria benefit its performance. 
As its pre-training capabilities are related to perplexity (PPL), we derive 14 quality criteria from the causes of text perplexity anomalies and introduce 15 common application domains to support domain mixing.
In this paper, we train a \textbf{Data} \textbf{Man}ager (\textbf{DataMan}) to learn quality ratings and domain recognition from pointwise rating, and use it to annotate a 447B token pre-training corpus with 14 quality ratings and domain type.
Our experiments validate our approach, using DataMan to select 30B tokens to train a 1.3B-parameter language model, demonstrating significant improvements in in-context learning (ICL), perplexity, and instruction-following ability over the state-of-the-art baseline. 
The best-performing model, based on the \emph{Overall Score l=5} surpasses a model trained with 50\% more data using uniform sampling. 
We continue pre-training with high-rated, domain-specific data annotated by DataMan to enhance domain-specific ICL performance and thus verify DataMan's domain mixing ability. 
Our findings emphasize the importance of quality ranking, the complementary nature of quality criteria, and their low correlation with perplexity, analyzing misalignment between PPL and ICL performance. 
We also thoroughly analyzed our pre-training dataset, examining its composition, the distribution of quality ratings, and the original document sources.
\end{abstract}
\vspace{-5pt}
\section{Introduction}
\vspace{-3pt}
As large language models (LLMs) demonstrate performance emergence driven by data scaling laws, the significance of data has become increasingly evident~\citep{kaplan2020scaling, brown2020fewshotlearners, chowdhery2023palm}. This trend prompts researchers to explore how to select pre-training data, including deduplication~\citep{lee2022deduplicating}, domain mixing~\citep{gao2020pile, shen2023slimpajama}, heuristic-based data selection~\citep{rae2022gopher, wenzek2019ccnet}, and data sampling using LLM quality signals~\citep{gunasekar2023textbooks, wettig2024qurating}.
Although these efforts aim to enhance data quality and diversity, deduplication and domain mixing are solely used as a priori or post-hoc steps in the data selection process. Furthermore, existing data selection methods typically rely on limited heuristics and human intuition, lacking comprehensive and clear criteria for data selection, thus making the selection of ideal pre-training data for LLMs an unresolved challenge.

Following this line, this paper provides guidelines for selecting pre-training data, including quality criteria, application domains, and a \textbf{Data} \textbf{Man}ager (\textbf{DataMan}) with a comprehensive quality rating and domain recognition, equipped by data sampling strategies to enhance the LLM performances.
Firstly, we believe that excellent quality criteria must: 1)-apply to a wide variety of texts; 2)-demonstrate a deep understanding of content, capturing semantic levels; and 3)-complement each other.
However, existing studies~\citep{rae2022gopher, wettig2024qurating} rely on limited heuristics and human intuition, while grounded in empirical findings, lack generality and comprehensiveness.
To address this issue, we are inspired by ``\emph{reverse thinking}'' — \textbf{prompting LLM to self-identify which criteria benefit its performance}, as its pre-training capability is related to perplexity (PPL)~\citep{muennighoff2024scaling,marion2023mdatapruning}.
We extracted documents with the top 2\% and bottom 2\% PPL from different sources and used Super LLM to \textbf{identify the reasons for anomalous in document perplexity}.
Through iterative refinement, we derived 13 quality criteria related to LLM performance: \emph{Accuracy, Coherence, Creativity, Grammatical Diversity, Knowledge Novelty, Language Consistency, Originality, Professionalism, Semantic Density, Sensitivity, Structural Standardization, Style Consistency, and Topic Focus}, and formed comprehensive criterion called \emph{Overall Score}.
Further, we introduced 15 common domain types in the LLM application industry~\citep {naveed2023comprehensive} for domain mixing and developed a full prompt to curate a fine-tuning dataset.
We employ a small-scale LLM to fine-tune DataMan via text generation and apply pointwise rating over pairwise rating~\citep{liu2007letor} for more applicable, cost-effective inference on vast datasets. 
Using DataMan, we annotated 447B tokens in the Slimpajama corpus~\citep{cerebras2023slimpajama} with quality ratings and domain types to create the DataPajama dataset.
While ensuring the diversity of sources and domains, we maximized the representativeness of quality criteria, sampling a 30B token subset from the DataPajama and trained the Sheared-LLaMA-1.3B language model~\citep{xia2023sheared} from scratch.

\begin{figure*}[t]
    \centering
    \centerline{\includegraphics[width=0.92\linewidth]{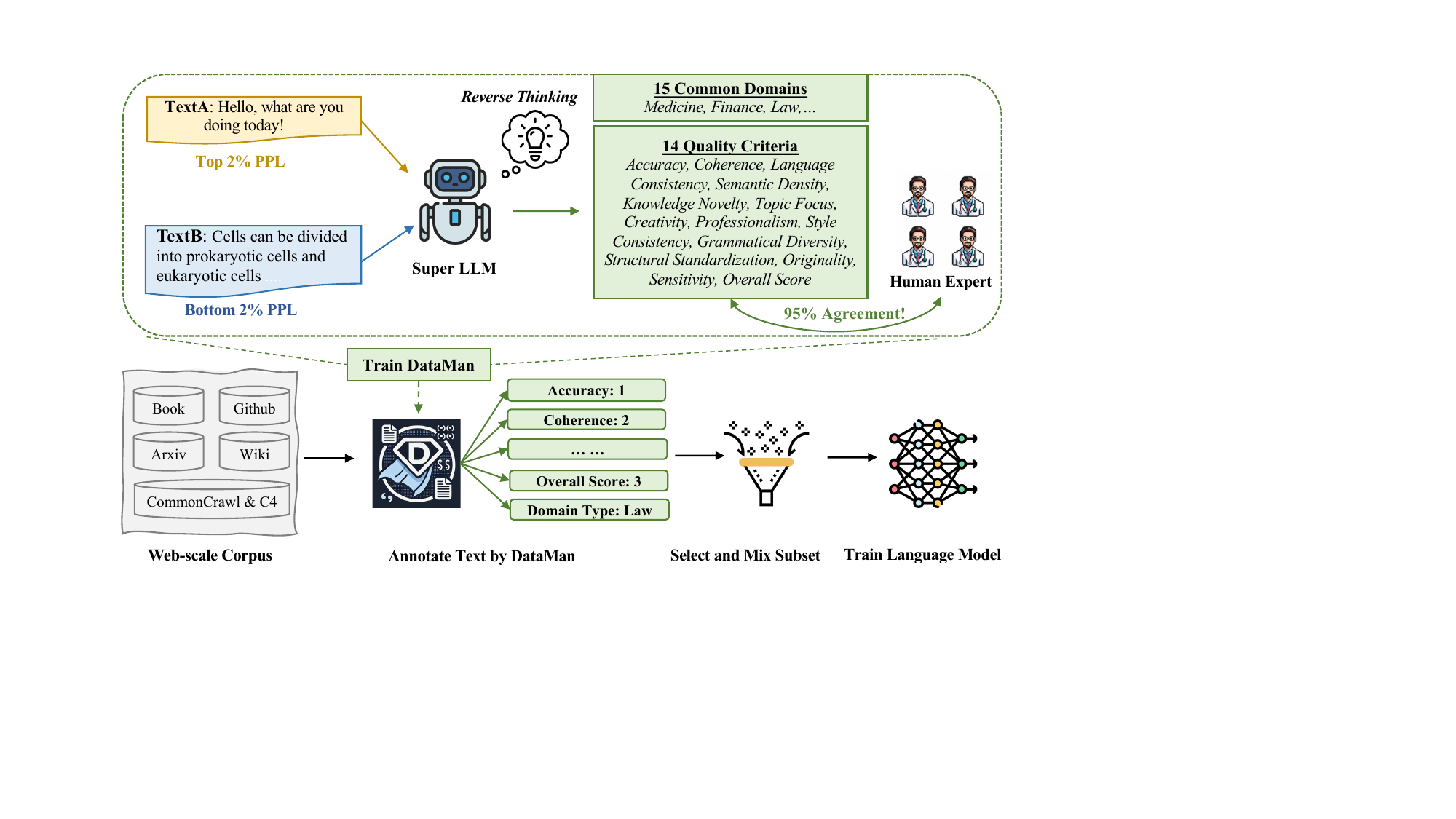}}
    \caption{
The pipeline of the Sample-with-DataMan model: We derived 14 quality criteria from LLMs' \emph{reverse thinking} and used DataMan to annotate the quality rating and domain type of pre-training data. By employing data sampling strategies to select subsets, the performance of trained LLMs outperforms the state-of-the-art data sampling baseline.
    }
    \label{fig:pipeline}
    \vskip -10pt
\end{figure*}

In our experiment, we initially validated the feasibility of using prompt as it achieved over 95\% agreement with human preferences.
We then revisited the limitation of pairwise rating~\citep{wettig2024qurating}, bounded pointwise rating error by its model loss, and fairly ranked the annotation results from both ratings to explain, from three perspectives, why we chose pointwise rating.
While training the DataMan model, we thoroughly analyzed its fine-tuning datasets, hyperparameter search, inference accuracy and efficiency of three model versions, and misclassification issues.\
Further, We evaluated language models trained on 30B tokens sampled with different data selection methods from the DataPajama dataset.
In ten downstream tasks, the Sample-with-DataMan model, trained on data sampled with Dataman's 13 quality criteria, outperformed the existing state-of-the-art (SOTA) baseline (Educational value $\tau=2$) in in-context learning performance by 0.4\% to 4.3\%, showcasing the effectiveness of these criteria.
As the \emph{Overall Score} rose from 1 to 5, both ICL performance and PPL improved significantly, confirming the necessity of quality ranking. 
Our strongest model was achieved at \emph{Overall Score l=5}, even surpassing the model trained on uniform sampling with 50\% more data, as the \emph{Overall Score} encompasses attributes of all criteria, supported by its high correlation with other quality criteria.
To solidify our results, we sampled a larger 60B data subset and compared the strongest Sample-with-DataMan model \emph{Overall Score l=5} against the existing SOTA baseline.
Meanwhile, we found PPL and ICL performance are not aligned strictly. 
Through correlation analysis and visualization, we reveal that PPL reflects general understanding, whereas ICL captures downstream generalization ability.
In the instruction following tasks, all Sample-with-DataMan models overwhelmingly surpassed the existing SOTA baseline with a win rate between 67.1\% and 78.5\%.
Additionally, We continued pre-training the strongest \emph{Overall Score l=5} model using high-rated, domain-specific data annotated by DataMan, achieving superior ICL performance in specific domains, thereby verifying DataMan's capability for domain mixing.
We also conducted an in-depth analysis of the DataPajama dataset, explored the distribution of DataMan's quality ratings from various sources, and inspected the original documents corresponding to each quality rating=1,2,3,4,5.
we also identified complementary relationships among quality criteria, while their low correlation with perplexity further confirms the novelty of DataMan's quality criteria.

Our contributions are summarized as follows:

\thickspace\thinspace 1. Based on the relationship between LLM performance and PPL, ``\emph{reverse thinking}'' allows LLMs to self-identify which criteria benefit its
performance, assemble with typical application domains to guide data selection.

\thickspace\thinspace 2. We introduce DataMan, offering comprehensive quality ratings and domain recognition, along with data sampling strategies to optimize LLM pre-training. Vast experiments validate its efficacy, setting new performance records.

\thickspace\thinspace 3. We will release the code, all models, and the annotated DataPajama dataset, paving the way for the community to explore the guidelines between data and LLMs further.
\vspace{-3pt}
\section{Related Work}
\vspace{-10pt}
\paragraph{Deduplication.}
Deduplicating training data is now standard in managing pre-training data for LLMs, as it greatly impacts performance~\citep{lee2022deduplicating}. While~\cite{kaplan2020scaling} and~\cite{hoffmann2022empirical} examine scaling laws with unique data trained for one epoch, some studies~\citep{hernandez2022scaling,muennighoff2024scaling,xue2024repeat} suggest that repeated data can harm performance, particularly as repetition epochs and model size grow. Additionally, fuzzy or semantic deduplication improves LLM performance~\citep{jiang2022fuzzydedup,abbas2023semdedup,tirumala2023d4}. But deduplication should precede quality-based selection and domain mixing, and it is unsuitable for sampling fixed-size subsets.

\paragraph{Heuristic-based Selection.} This selection approach includes two methods: \textit{Rule-based heuristics}, which apply manually crafted rules—such as mean word length, stop word fraction, and word repetition thresholds~\citep{laurenccon2022bigscience,TogetherAI,penedo2023refinedweb,soldaini2024dolma}—to filter data, exemplified by the C4 filter~\citep{raffel2020exploring} and Gopher rules~\citep{rae2022gopher}. These reduce noise effectively but need complex manual design. In contrast, \textit{Model-based heuristics} use models like binary grammar discriminators~\citep{chowdhery2023palm,touvron2023llama} to select data resembling the target domain, alongside techniques like importance resampling~\citep{xie2023data} and perplexity filtering~\citep{wenzek2019ccnet, muennighoff2024scaling, marion2023less}. However, they require more precise quality ratings for optimal selection.

\paragraph{Domain mixture.}
Most pre-training datasets, like the Pile~\citep{gao2020pile}, comprise mixed data from various sources and domains~\citep{nijkamp2023codegen2,zhang2023llama,yang2024gpt4tools,maini2024rephrasing,li2024synthetic}. As LLMs gain traction, domain-specific data for improved functionalities are increasingly used in model training~\citep{du2022glam,gao2023llama}. Identifying the optimal domain mixture ratio is essential for effective LLM pre-training~\citep{wang2023data}. Early attempts to define this relied on experiments and intuition~\citep{gao2020pile,thoppilan2022lamda}. Recent studies have begun to use automatic methods, such as domain generalization~\citep{xie2023data,xie2024doremi}, domain gradients~\citep{fan2023doge}, and loss evaluation~\citep{xia2023sheared}, to assign domain weights and assess model performance across various mixtures. This inspires our work to introduce domain types to assist in pre-training data mixture.

\paragraph{LLM quality signals.}
Appropriate LLM quality signals are valuable for selecting pre-training data~\citep{gunasekar2023textbooks}. Research shows that data enriched with facts and trivia aids LLMs in accurately addressing niche topics and fictional worlds~\citep{petroni2019language,korbak2023pretraining}. Other studies have emphasized the educational value of data as a key for enhancing LLMs' reasoning capabilities~\citep{wei2022chain,kojima2022large}. Recent efforts have integrated these insights, proposing four quality criteria—writing style, facts and trivia, educational value, and required expertise—to assess pre-training data and enhance LLM capabilities~\citep{wettigqurating}. However, these criteria depend heavily on human intuition and thus lack generality and comprehensiveness.
\vspace{-3pt}
\section{Annotating Text by DataMan}
\vspace{-3pt}
We developed \textbf{DataMan} (see Figure \ref{fig:pipeline}), a data manager that comprehensively annotates 14 quality criteria and domain types, enabling ideal data selection and mixing.

\subsection{Overview of the Annotation}
Let $\mathbf{t} = \{ t_1, \ldots, t_n \}$ represent all documents to be annotated. 
The annotation results of DataMan correspond to querying the ratings of document $t_n$ across all quality criteria and its domain type.
Assume that the quality ratings and domain recognition are in a multi-level annotation format as $\mathcal{L} = \{(l_1^{1}, \ldots, l_1^{C}), \ldots, (l_n^{1},\ldots, l_n^{C})\}$, where $l_n^{C} \in \{1, \ldots, K\}$ is the label for the $C$-th criterion or domain of document $t_n$. 
For this paper, $K$ is 5 for quality ratings and 15 for domain recognition.
Let $F$ be a class of functions, and $f \in F$ be the annotation function. 
We use a super LLM as the annotation function, recording the annotation results for each criterion and domain, expressed as: $ f(t, \mathcal{L}) = \left\{ (t_1, l_1^{1}, \ldots, l_1^{C}), \ldots, (t_n, l_n^{1}, \ldots, l_n^{C}) \right\}. $
Thus, we can quickly create a fine-tuning dataset for DataMan, $\mathcal{S} = \left\{ (t_i, l_i^{1}, \ldots, l_i^{C}) \right\}$. 
Essentially, our method uses pointwise rating by pointwise learning to rank (L2R) model \citep{liu2007letor, liu2009learning}. 
We minimize the loss function based on the document, annotation labels, and function:
\vspace{-5pt}
\begin{equation}\label{eq:1}
L(f; t, \mathcal{L}) = \sum_{i=1}^{n} \left( f(t_i) - l_i \right)^{2}.
\vspace{-3pt}
\end{equation}
This trains DataMan to learn the annotation functions for each quality criterion and domain type.

\subsection{Prompt Curation}
How to define the quality criteria and domain types of texts? We believe that excellent quality criteria should: 1)-apply to a wide variety of texts, 2)-demonstrate a deep understanding of content, capturing semantic levels, and 3)- complement each other. However, prior studies largely relied on human intuition, such as educational value in \citet{gunasekar2023textbooks}, writing style in \citet{wettig2024qurating}, and toxicity and privacy in \citet{korbak2023pretraining}, which lack generality and comprehensiveness.
To address this issue, we undertook an in-depth exploration of text quality, motivated by ``\emph{reverse thinking}''—\emph{prompting the LLM to self-identify which criteria benefit its performance}. Specifically, since LLMs' pre-training abilities are closely related to their PPL~\citep{muennighoff2024scaling,marion2023mdatapruning}, where ``high PPL indicates data is difficult to learn, and vice versa'', we focused on the training text from various sources with the top 2\% and bottom 2\% of PPL. We devised an analytical prompt for a Super LLM\footnote{
The Super LLM can independently refer to any advanced LLM. In this paper, We use the state-of-the-art LLM available at that time, \texttt{GPT-4-0125-preview}.} to investigate the reason behind these perplexity anomalies in documents, aiming to analyze the traits of both easy-to-learn and difficult-to-learn data. Through iterative refinement, we derived 14 quality criteria as shown in Figure~\ref{fig:pipeline}. Additionally, we identified the 15 domain types that need to be assessed from typical LLM application industries~\citep{naveed2023comprehensive}, integrating the \emph{"let's think step by step"} chain-of-thought prompting strategy \citep{wei2022chain} and a thoughtfully designed system prompt. 
Further details of the entire process are in Appendix \ref{app:detailed_prompts}.

\paragraph{Prompt validation.}
We validate prompt effectiveness using clear cases before prompt use. Initially, we gathered a pool of documents preliminarily rated by an independent group, splitting them into two sets of ten documents each—high-rated and low-rated—to ensure a distinct quality gap. Table \ref{tab:prompt_validation_sources} of Appendix \ref{app:detailed_prompts} details these document sources. These 20 documents were randomly shuffled and then rated on a scale of 1-5 based on each quality criterion by five independent human annotators who had not seen them before. Subsequently, using the super LLM with our prompt, we evaluated the same documents and compared super LLM ratings with human majority votes, finding over 95\% agreement. Also, we ensured inter-rater reliability among human annotators by calculating the Kappa coefficient~\citep{mchugh2012interrater}, which validated the consistency of human ratings. However, this human validation remains subjective, we hope the community develops a more rigorous method.

\subsection{Why Use Pointwise Rating?}

Using LLMs to rate text falls into two categories: pointwise rating and pairwise rating. Here, we explain why we chose the pointwise rating from the following three aspects:

\paragraph{Pairwise rating limitations.} 
We revisited pairwise rating~\citep{wettig2024qurating} and observed minimal quality differences among the top-3 documents ranked by \emph{writing style}, as even humans struggle to discern differences (see Table \ref{tab: ten_wriqual_texts}). 
This highlights the limitations of pairwise ratings when quality differences are marginal, as documents meeting an ``acceptable'' quality threshold should be accepted, where pointwise rating aligns well with human judgment.
Appendix B.1 notes that creating a fine-tuning dataset requires Super LLM to make 40 predictions per criterion and document pairs, then use LLM preferences as labels.
Table 5 shows that document pairs need a $\geq$50\% selection probability difference for quality criteria, also required for inference in Table 6, thus questioning the practicality of pointwise ratings.
For $N$ documents, pointwise ratings only need $N$ ratings, whereas pairwise ratings demands $N*(N-1)$ comparisons with 40 predictions each, greatly increasing costs.
Despite both ratings capture different aspects, pointwise rating's broader applicability and cost-effectiveness in fine-tuning dataset curation or pre-training data annotation make it our preferred choice.

\textbf{Bound pointwise rating error.} 
We present the mathematical connection between rating errors and loss in the pointwise rating model~\citep{chen2009ranking}.
\vspace{-5pt}
\begin{gather}~\label{eq:2}
1 - NDCG(f; t, \mathcal{L}) \leq \frac{15\sqrt{2}}{N_n} \left( \left( \sum_{i=1}^{n} D(t_i)^2 \right) - n \prod_{i=1}^{n} D(t_i)^{2/n} \right)^{1/2} \left( L(f; t, \mathcal{L}) \right)^{1/2}, \\
NDCG(f; t, \mathcal{L}) = \frac{1}{N_n} \sum_{i=1}^{n} G(l(\pi_f(t_i))) D(t_i), \\
N_n = \max_{\pi} \sum_{i=1}^{n} G(l(\pi(t_i))) D(t_i),
\vspace{-5pt}
\end{gather}
where NDCG is rating measures defined with respect to \( K \)-level ratings \( \mathcal{L} \), \( G \) is the gain function, \( \pi_f \) is the rating list produced by the rating function \( f \), and \( D \) is the position discount function.
One usually sets \( G(z) = 2^z - 1, D(z) = \frac{1}{\log_2(1+z)} \) if \( z \leq M \), and \( D(z) = 0 \) if \( z > M \) ($M$ is a fixed integer).
Thus, as the pointwise rating model minimizes loss via training, the rating error also reduces, validating the feasibility of pointwise ratings theoretically.

\textbf{Fairly compare both rating results.}
To ensure a fair comparison despite differing evaluation metrics, we ranked annotations from Qurating's pairwise and DataMan's pointwise ratings on the same 58,824 documents from the first split of Qurating's 1B analysis set\footnote{https://huggingface.co/datasets/princeton-nlp/QuRatedPajama-1B\_tokens\_for\_analysis}. We then presented the top, median, and bottom ten documents based on Qurating's four criteria—writing style, facts and trivia, educational value, and required expertise—and DataMan's Overall Score with sum of the remaining criteria. All results are provided here\footnote{https://github.com/pengr/DataMan/blob/main/README.md}. Our conclusions are:
\textit{i)}-Higher Qurating criteria correlate with higher DataMan's Overall Scores for the top and bottom 10 documents, and vice versa.
\textit{ii)}-Median 10 documents show minimal changes in Qurating's criteria, yet their Overall Scores vary from 3 to 5, supporting that ``Pairwise rating struggles with minimal quality differences, while pointwise rating matches human judgment.''
\textit{iii)}-Rankings by Overall Score highlight top-quality documents, especially in STEM fields.
\textit{iv)}-The ``\emph{application\_domain}'' field accurately categorizes domains, aiding in domain-specific continue pre-training, as shown in Table~\ref{tab:domain_specific_results}.

\subsection{Training the DataMan Model} \label{sec:DataMan Rater Training}
After developing the prompt and rating method, we collected documents from both in-source and out-of-source of the SlimPajama corpus~\citep{cerebras2023slimpajama} to train DataMan.
By prompting the Super LLM, we obtained 14 quality ratings and domain types for each document, creating a fine-tuning dataset of 35,700 documents at a cost of \$13,858.
Documents were limited to 2,048 tokens (averaging 810 tokens), surpassing the [256, 512] token range of~\cite{wettig2024qurating} when handling sentences with broader length variations.
Subsequently, we fine-tuned the DataMan model with Qwen2-1.5B \citep{yang2024qwen2} using text generation loss. 
In Table \ref{tab:sft_model_eval}, the DataMan model achieves near 80\% average test accuracy across all criteria, with an \emph{Overall Score} test accuracy of 81.3\% and a \emph{domain recognition} test accuracy of 86\%.
Appendix~\ref{app:dataman_model} also provides details on the fine-tuning dataset statistics, average scores for each quality criterion per domain, correlation among quality criteria, three DataMan model versions, the chat prompt for text generation, the hyperparameter search, reports on inference accuracy, misclassification analysis and inference efficiency.

\vspace{-5pt}
\section{Managing data by DataMan}\label{sec:manage_data_by_dataman}
\vspace{-3pt}
In this section, we apply the DataMan model to manage data by selecting a high-quality, diverse document subset from the pre-training corpus.
Each document \( d_i \) in the corpus \( D \), with source \( s_i \), is annotated by the DataMan model with 14 quality ratings and domain types as \( \mathcal{L} = \{(l_i^{1}, \ldots, l_i^{C-1}, q)\} \), where \( q \) is the domain type. 
Given the source and domain distribution probabilities \( P(s) \) and \( P(q) \) respectively, we perform top-k (k is the selected subset size) sampling without replacement for each quality criterion across source and domain distribution, using the probability:
\begin{equation}~\label{eq:3}
P(d_i) = \frac{P(d_i \mid l^j, s, q) \cdot P(s,q)}{\sum_{d_j \in \text{top-k}(l^{j})} P(d_j \mid l^j, s, q) \cdot P(s,q)}, \text{and} \ P(d_i | l^j) = \frac{l_i^j}{\sum_{d_j \in D} l_i^j}.
\vspace{-5pt}
\end{equation}

This method maximizes sample representativeness based on quality rating while ensuring diversity in source and domain distributions, and sampling without replacement prevents duplicate data. 
To assess if the \emph{Overall Score} covers all criteria, we substitute top-k with uniform sampling based on fixed \emph{Overall Score} ratings. 
These techniques implicitly steer language modeling toward reward-weighted regression~\citep{peters2007reinforcement, korbak2023pretraining}, expanding maximum likelihood estimation loss with a data reward mechanism.
\vspace{-3pt}
\section{Experiments}
\vspace{-3pt}
We empirically validate the \ourmethod{} method by training the language model from scratch.

\vspace{-3pt}
\subsection{Experimental Setup}
\vspace{-5pt}
\paragraph{\ourdata{}.} 
We used the Llama tokenizer \citep{touvron2023llama} to segment Slimpajama corpus~\citep{cerebras2023slimpajama}, a cleaned and deduplicated version of the RedPajama~\citep{together2023redpajama}, into 1024-tokens documents. The \ourmethod{} model then annotated these documents with 14 quality ratings and domain types, creating the 447B token \ourdata{} for pre-training.
Although the annotation process is expensive, it can be reduced via large-scale parallelization and cost-effective \ourmethod{} models.
The quality ratings and domain types in \ourdata{} can serve various purposes, such as data selection, data mixing, or continued pre-training in specific domains. 
Detailed statistics of \ourdata{} can be found in Table \ref{tab:sources_stats} of Appendix \ref{app:training_details}.

\paragraph{Data selection methods.}
For each baseline, we select a 30B token training dataset from \ourdata{} with one of the following methods, while retaining the original source proportion as the overall dataset. 
We leave it as future work to explore combinations of quality criteria to broaden our method.
\begin{itemize}

\item \textit{Uniform}: We select randomly with a uniform probability across documents. For comparison's sake, we train an additional model on 45B tokens, requiring 50\% more compute.

\item \textit{DSIR}: We apply data selection with importance resampling (DSIR) \citep{xie2023data} and select examples that resemble either English Wikipedia or the Book domain \citep{together2023redpajama}---commonly used as proxies for quality \citep{brown2020language, touvron2023llama, xie2023data}. We follow \cite{xie2023data} and train hashed bigram models on \ourdata{} and the target data.

\item \textit{Perplexity Filtering}: We implement perplexity filtering \citep{wenzek2019ccnet, marion2023less} and select the documents with the lowest/highest perplexity scores, as computed by a pre-trained Sheared-Llama-2.7B model \citep{xia2023sheared}---2$\times$ the size of our \ourmethod{} model.

\item \textit{Sample with Qurating}:
We sample 30B tokens according to each of the four criteria described in \cite{wettig2024qurating}: \textit{writing style, facts and trivia, educational value, and required expertise}. 
Specifically, we normalize the variance of the quality ratings to be $1$ and then sample with temperature $\tau \in \{0.0 \text{ (i.e., top-$k$ selection)}, 2.0\}$. 
Additionally, we merge the Qurating-sampled data for $\tau=2.0$ of the four criteria as \textit{criteria mix}, and subsampling is as randomly to 30B tokens, ensuring that we exclude duplicate documents.

\item \textit{Sample with \ourmethod{} (Ours)}: For each of 13 quality criteria, we perform top-k sampling based on the quality ratings and domain type described in \cref{sec:manage_data_by_dataman}. Additionally, we explore sampling 30B tokens with overall scores $l \in \{1, 2, 3, 4, 5\}$  to demonstrate the effectiveness of all criteria.
\end{itemize}

\vspace{-8pt}
\paragraph{Training settings.}
Using different data selection methods, we select a 30B token subset from \ourdata{} and train a randomly initialized Sheared-Llama-1.3B language model \citep{xia2023sheared} for one epoch in a randomly shuffled order.
We train on slightly more data than the compute-optimal ratio \citep{hoffmann2022training} because more training tokens better reflect the performance gains attributed to data quality.
Further details are in \cref{app:training_details}. 

\vspace{-3pt}
\paragraph{Evaluation metrics.}
We provide a holistic evaluation of the language models trained on 30B tokens:
\begin{itemize}[topsep=0pt,parsep=0pt,partopsep=0pt,leftmargin=1em]
    \item We measure the perplexity over SlimPajama's validation set and test set, 500M tokens each.
    
    \item  We evaluate the in-context learning (ICL) performance using \texttt{lm-evaluation-harness} \citep{eval-harness}.
    We study 10 tasks, comprising 5 reading comprehension tasks (ARC-easy/challenge \citep{clark2018think}, SciQA \citep{welbl2017crowdsourcing}, LogiQA \citep{liu2020logiqa}, BoolQ \citep{clark2019boolq}), 3 commonsense reasoning tasks (HellaSwag \citep{zellers2019hellaswag}, PIQA \citep{bisk2020piqa}, WinoGrande \citep{sakaguchi2021winogrande}) and 2 knowledge-intensive tasks (NQ \citep{kwiatkowski2019natural}, MMLU \citep{hendrycks2021measuring}). We report the detailed settings in \cref{app:training_details}.

    \item We evaluate the instruction-following capabilities of our models, borrowing the setting used by \cite{xia2023sheared}. We perform supervised fine-tuning on 10,000 instruction-response pairs from the ShareGPT dataset. We evaluate another 1,000 instructions and use the AlpacaFarm codebase \citep{dubois2024alpacafarm} to judge the responses from two models with \texttt{GPT-4o}.
\end{itemize}

\subsection{Results}
\begin{figure}
  \begin{minipage}[b]{0.55\textwidth}
    \includegraphics[width=\textwidth]{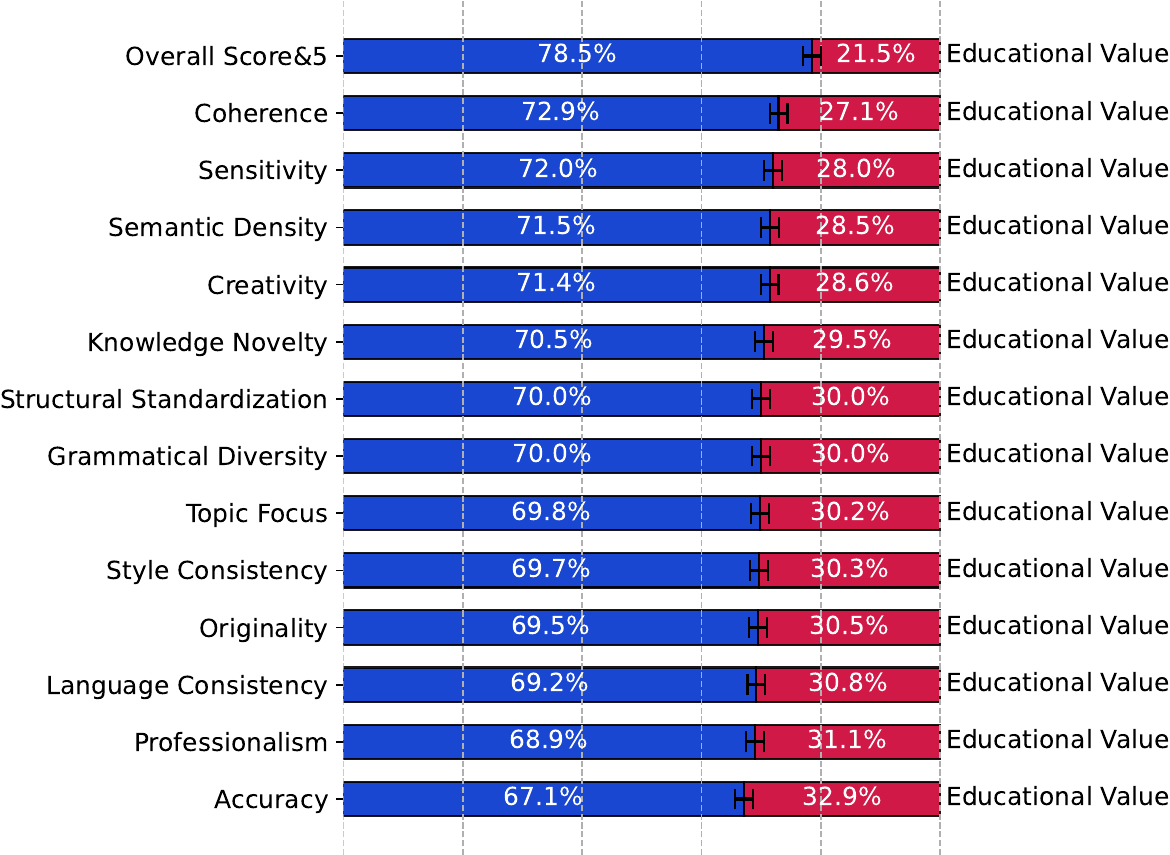}
  \end{minipage} 
  \hspace{0.2cm} 
  \begin{minipage}[b]{0.35\textwidth}
    \caption[Caption]{Instruction following win rates of Sample-with-DataMan models v.s. the state-of-the-art baseline (Educational value $\tau = 2.0$) after instruction fine-tuning on 10K ShareGPT examples. The results indicate that, under the same SFT conditions, our model consistently surpasses the SOTA Baseline, with the \emph{Overall Score l=5} reaching a
impressive win rate at 78.5\%.
}
  \label{fig:instruction_ft_winrates}
  \end{minipage}
  \vspace{-3pt}
\end{figure}

We report the model's perplexity and ICL results in Table~\ref{tab:main_results}, the instruction-following evaluation in Figure~\ref{fig:instruction_ft_winrates}, the domain-specific continue pre-training performance in Table~\ref{tab:domain_specific_results}. 
Appendix~\ref{app:training_details} provides comprehensive results for all models, including validation and test perplexity across sources in Tables~\ref{tab:val_ppl_results},~\ref{tab:test_ppl_results}, and ICL results for individual task in Table~\ref{tab:icl_results}.

\vskip -2pt
\paragraph{Traditional methods perform poorly.} Table~\ref{tab:main_results} indicates that DSIR and perplexity filtering perform worse than random uniform sampling. This indicates that model output–based methods, despite their widespread use, are ineffective for data selection.

\vskip -2pt
\paragraph{Clear quality criteria work, but Qurating’s criteria mix fails.} Table~\ref{tab:main_results} shows that selecting data with Qurating’s four criteria \citep{wettig2024qurating} improves performance compared to uniform sampling, highlighting the importance of clear LLM quality signals. 
Educational value $\tau=2.0$ is the current SOTA baseline. 
However, the criteria mix of Qurating's four criterion did not perform well, possibly due to a lack of complementarity among the Qurating’s criteria.

\vskip -2pt
\paragraph{\ourmethod{}'s criteria surpasses SOTA baseline and \emph{Overall Score} works best.}
In Table~\ref{tab:main_results}, compared to the SOTA baseline (Educational value $\tau=2.0$), our 13 individual quality criteria improved ICL performance by 0.4\% to 4.3\%. 
For example, the sample-with-creativity model achieved an impressive score of 60.6 in commonsense reasoning tasks.
As the comprehensive criterion \emph{Overall Score} increased from 1 to 5, performance gains significantly, highlighting the necessity of quality ranking.
Our \emph{Overall Score l=5} works best, even exceeding the \textit{Uniform +50\% data} baseline, validating the feasibility of combining the 13 quality criteria into a composite criterion via LLM weighting. 
This approach not only avoids the disruption of manual adjustments but also achieves optimal results.
The correlations among all criteria in Figure~\ref{fig:quality_criteria_corr} further confirm this point.

\vskip -2pt
\paragraph{PPL and ICL are not strictly aligned.}
Our results in Table~\ref{tab:main_results} reveal a trend where PPL and ICL metrics correlate to a degree, increasing or decreasing simultaneously, but they are not aligned strictly. This meets with the intuition that PPL implies general understanding, while ICL focuses more on downstream generalization. 
Further analysis see Figure~\ref{fig:icl_vs_ppl} in Appendix~\ref{app:training_details}.
Notably, \ourmethod{}'s high-rated \emph{Overall Score} achieves an optimal trade-off between understanding and generalization.
\begin{table*}[t]
\centering
\caption{
Sample-with-\ourmethod{} models improve perplexity and in-context learning (ICL) results, and the \emph{Overall Score l=5} perform best. 
We report the validation, test perplexity, and ICL performance of 10 downstream tasks. 
We highlight the best result in each column and
improvement over uniform sampling with the 30B token budget.
In \cref{app:training_details}, we report full results of perplexity in Table~\ref{tab:val_ppl_results}, ~\ref{tab:test_ppl_results} and ICL in Table~\ref{tab:icl_results}.
}
\label{tab:main_results}
\vskip -5pt
\setlength{\tabcolsep}{2pt}
\resizebox{1.\textwidth}{!}{
\begin{tabular}{lllccccc}
\toprule
\multicolumn{2}{l}{\textbf{Selection Method}} & \multicolumn{1}{c}{\begin{tabular}[c]{@{}c@{}}\textbf{Val} \\ \textbf{Perplexity} \\\end{tabular}}&\multicolumn{1}{c}{\begin{tabular}[c]{@{}c@{}}\textbf{Test} \\ \textbf{Perplexity} \\\end{tabular}}& \multicolumn{1}{c}{\begin{tabular}[c]{@{}c@{}}\textbf{Reading} \\ \textbf{Comprehension} \\ \textit{(5 tasks)}\end{tabular}} & \multicolumn{1}{c}{\begin{tabular}[c]{@{}c@{}}\textbf{Commonsense} \\ \textbf{Reasoning} \\ \textit{(3 tasks)}\end{tabular}} & \multicolumn{1}{c}{\begin{tabular}[c]{@{}c@{}}\textbf{World} \\ \textbf{Knowledge} \\ \textit{(2 tasks)}\end{tabular}} & \multicolumn{1}{c}{\begin{tabular}[c]{@{}c@{}}\\\textbf{Average} \\ \textit{(10 tasks)}\end{tabular}} \\
\midrule
\multirow{2}{*}{\begin{tabular}[c]{@{}l@{}}Uniform\end{tabular}} & & \multicolumn{1}{c}{10.7}  & \multicolumn{1}{c}{10.75}  & \multicolumn{1}{c}{50.9}  & \multicolumn{1}{c}{55} & \multicolumn{1}{c}{14.9} & \multicolumn{1}{c}{44.9} \\
 & \textit{+50\% data} & 10.09 \gda{0.61} & 10.14 \gda{0.61}  & 52.9 \gua{2.0}  & 57.0 \gua{2.0}  & 15.9 \gua{1.0} & 46.8 \gua{1.9} \\
\midrule
\multirow{2}{*}{DSIR}  & \textit{with Wiki}  & 13.34 \rda{2.64} & 13.37 \rda{2.62}  & 50.1 \rua{0.8}  & 49.8 \rua{5.2}  & 14.7 \rua{0.2} & 42.9 \rua{2.0} \\
 & \textit{with Book}  & 13.60 \rda{2.90} & 13.59 \rda{2.84}  & 47.9 \rua{3.0}  & 56.6 \gua{1.6}  & 14.1 \rua{0.8} & 43.8 \rua{1.1} \\
\cmidrule(lr){1-2}
\multirow{2}{*}{Perplexity}  & \textit{lowest} & 15.98 \rda{5.28} & 16.04 \rda{5.29}  & 48.3 \rua{2.6}  & 49.6 \rua{5.4}  & 13.7 \rua{1.2} & 41.7 \rua{3.2} \\
& \textit{highest}  & 11.32 \rda{0.62} & 11.34 \rda{0.59}  & 49.6 \rua{1.3}  & 53.5 \rua{1.5}  & 13.4 \rua{1.5} & 43.5 \rua{1.4} \\
\cmidrule(lr){1-2}
\multirow{2}{*}{\begin{tabular}[c]{@{}l@{}}Writing Style\end{tabular}}  & \textit{top-k}  & 13.01 \rda{2.31} & 12.97 \rda{2.22}  & 49.3 \rua{1.6}  & 53.3 \rua{1.7}  & 13.5 \rua{1.4} & 43.4 \rua{1.5} \\
& $\tau=2.0$ & 10.60 \gda{0.10} & 10.64 \gda{0.11}  & 51.0 \gua{0.1}  & 55.8 \gua{0.8}  & 14.1 \rua{0.8} & 45.0 \gua{0.1} \\
\cmidrule(lr){1-2}

\multirow{2}{*}{\begin{tabular}[c]{@{}l@{}}Facts \& Trivia\end{tabular}}  & \textit{top-k}  & 14.38 \rda{3.68} & 14.33 \rda{3.58}  & 54.3 \gua{3.4}  & 51.7 \rua{3.3}  & 15.5 \gua{0.6} & 45.8 \gua{0.9} \\
& $\tau=2.0$ & 10.68 \gda{0.02} & 10.72 \gda{0.03}  & 52.7 \gua{1.8}  & 55.6 \gua{0.6}  & 15.6 \gua{0.7} & 46.2 \gua{1.3} \\
\cmidrule(lr){1-2}
\multirow{2}{*}{\begin{tabular}[c]{@{}l@{}}Educational Value\end{tabular}}  & \textit{top-k}  & 13.54 \rda{2.84} & 13.49 \rda{2.74}  & 54.7 \gua{3.8}  & 54.9 \rua{0.1}  & 14.4 \rua{0.5} & 46.7 \gua{1.8} \\
& $\tau=2.0$ & 10.67 \gda{0.03} & 10.72 \gda{0.03}  & 53.3 \gua{2.4}  & 56.3 \gua{1.3}  & 15.7 \gua{0.8} & 46.7 \gua{1.8} \\
\cmidrule(lr){1-2}
\multirow{2}{*}{\begin{tabular}[c]{@{}l@{}}Required Expertise\end{tabular}} & \textit{top-k}  & 14.97 \rda{4.27} & 14.92 \rda{4.17}  & 52.8 \gua{1.9}  & 48.7 \rua{6.3}  & 14.3 \rua{0.6} & 43.9 \rua{1.0} \\
& $\tau=2.0$ & 10.7 & 10.74 \gda{0.01}  & 52.7 \gua{1.8}  & 55.5 \gua{0.5}  & 15.0 \gua{0.1} & 46.0 \gua{1.1} \\
\cmidrule(lr){1-2}
Criteria mix  & $\tau=2.0$ & 10.63 \gda{0.07} & 10.68 \gda{0.07}  & 52.1 \gua{1.2}  & 55.5 \gua{0.5}  & 15.2 \gua{0.3} & 45.7 \gua{0.8} \\
\midrule
Accuracy  & \textit{top-k}  & 10.82 \rda{0.12} & 10.80 \rda{0.05}  & 53.8 \gua{2.9}  & 58.2 \gua{3.2}  & 16.6 \gua{1.7} & 47.7 \gua{2.8} \\
\cmidrule(lr){1-2}
Coherence & \textit{top-k}  & 10.72 \rda{0.02} & 10.71 \gda{0.04}  & 54.9 \gua{4.0}  & 58.8 \gua{3.8}  & 16.1 \gua{1.2} & 48.3 \gua{3.4} \\
\cmidrule(lr){1-2}
Creativity  & \textit{top-k}  & 11.08 \rda{0.38} & 11.00 \rda{0.25}  & 53.0 \gua{2.1}  & \textbf{60.6 \gua{5.6}} & 15.2 \gua{0.3} & 47.7 \gua{2.8} \\
\cmidrule(lr){1-2}
Grammatical Diversity & \textit{top-k}  & 10.87 \rda{0.17} & 10.86 \rda{0.11}  & 55.1 \gua{4.2}  & 58.8 \gua{3.8}  & 16.5 \gua{1.6} & 48.5 \gua{3.6} \\
\cmidrule(lr){1-2}
Knowledge Novelty  & \textit{top-k}  & 11.01 \rda{0.31} & 11.01 \rda{0.26}  & 54.6 \gua{3.7}  & 56.9 \gua{1.9}  & 15.5 \gua{0.6} & 47.5 \gua{2.6} \\
\cmidrule(lr){1-2}
Language Consistency  & \textit{top-k}  & 10.35 \gda{0.35} & 10.35 \gda{0.40}  & 54.1 \gua{3.2}  & 59.3 \gua{4.3}  & 16.7 \gua{1.8} & 48.2 \gua{3.3} \\
\cmidrule(lr){1-2}
Originality & \textit{top-k}  & 10.68 \gda{0.02} & 10.67 \gda{0.08}  & 53.9 \gua{3.0}  & 58.6 \gua{3.6}  & 16.4 \gua{1.5} & 47.8 \gua{2.9} \\
\cmidrule(lr){1-2}
Professionalism  & \textit{top-k}  & 11.27 \rda{0.57} & 11.26 \rda{0.51}  & 54.6 \gua{3.7}  & 54.8 \rua{0.2}  & 15.9 \gua{1.0} & 46.9 \gua{2.0} \\
\cmidrule(lr){1-2}
Semantic Density & \textit{top-k}  & 11.10 \rda{0.40} & 11.09 \rda{0.34}  & 54.4 \gua{3.5}  & 58.1 \gua{3.1}  & 16.7 \gua{1.8} & 48.0 \gua{3.1} \\
\cmidrule(lr){1-2}
Sensitivity & \textit{top-k}  & 10.11 \gda{0.59} & 10.13 \gda{0.62}  & 54.7 \gua{3.8}  & 59.2 \gua{4.2}  & 16.1 \gua{1.2} & 48.3 \gua{3.4} \\
\cmidrule(lr){1-2}
Structural Standardization & \textit{top-k}  & 12.11 \rda{1.41} & 12.11 \rda{1.36}  & 53.7 \gua{2.8}  & 57.0 \gua{2.0}  & 17.1 \gua{2.2} & 47.4 \gua{2.5} \\
\cmidrule(lr){1-2}
Style Consistency  & \textit{top-k}  & 10.74 \rda{0.04} & 10.73 \gda{0.02}  & 55.1 \gua{4.2}  & 59.6 \gua{4.6}  & 16.2 \gua{1.3} & 48.7 \gua{3.8} \\
\cmidrule(lr){1-2}
\begin{tabular}[c]{@{}l@{}}Topic Focus\end{tabular} & \textit{top-k}  & 10.41 \gda{0.29} & 10.41 \gda{0.34}  & 54.6 \gua{3.7}  & 58.4 \gua{3.4}  & 15.6 \gua{0.7} & 47.9 \gua{3.0} \\
\cmidrule(lr){1-2}
\multirow{5}{*}{\begin{tabular}[c]{@{}l@{}}Overall \\ Score\end{tabular}} & \textit{l=1} & 23.83 \rda{13.13}  & 23.95 \rda{13.20} & 43.1 \rua{7.8}  & 47.5 \rua{7.5}  & 13.1 \rua{1.8} & 38.4 \rua{6.5} \\
  & \textit{l=2} & 12.84 \rda{2.14} & 12.91 \rda{2.16} & 50.3 \rua{0.6} & 50.9 \rua{4.1} & 14.7 \rua{0.2} & 43.4 \rua{1.5} \\
  & \textit{l=3} & 11.75 \rda{1.05} & 11.78 \rda{1.03} & 50.7 \rua{0.2} & 54.1 \rua{0.9} & 15.2 \gua{0.3} & 44.6 \rua{0.3} \\
  & \textit{l=4} & \textbf{10.21} \textbf{\gda{0.49}} & 10.22 \gda{0.53}  & 53.5 \gua{2.6}  & 60.1 \gua{5.1}  & 16.0 \gua{1.1} & 47.9 \gua{3.0} \\
  & \textit{l=5} & 10.52 \gda{0.18} & \textbf{10.50 \gda{0.25}} & \textbf{55.2 \gua{4.3}} & 60.2 \gua{5.2}  & \textbf{17.4 \gua{2.5}} & \textbf{49.1 \gua{4.2}} \\
\bottomrule
\end{tabular}
}
\vskip -10pt
\end{table*}

\vskip -2pt
\paragraph{\ourmethod{}'s instruction-following abilities also well.}
As shown in Figure~\ref{fig:instruction_ft_winrates}, we compare the instruction-following win rates between the Sample-with-\ourmethod{} model and the SOTA baseline (Educational value $\tau=2.0$).
The results indicate that, under the same supervised fine-tuning conditions, our model consistently surpasses the SOTA Baseline, with the \emph{Overall Score l=5} reaching an impressive win rate at 78.5\%, further validating the superior performance of our method.

\vskip -2pt
\begin{table*}[t]
\centering
\renewcommand{\arraystretch}{1.1} 
\setlength{\tabcolsep}{4pt}
\caption{The Sample-with-\ourmethod{} model further improves perplexity and in-context learning (ICL) results when continuing pre-training on high \emph{Overall Score},  domain-specific data. We report the validation, test perplexity, and ICL performance of corresponding MMLU subtasks.}
\label{tab:domain_specific_results}
\resizebox{1.\textwidth}{!}{
\begin{tabular}{cccccc}
\toprule
 & \textbf{Val Perplexity} & \textbf{Test Perplexity}  & \textbf{Anatomy}  &  \textbf{College Medicine} & \textbf{Medical Genetics} \\
\hline
Overall Score \emph{l=5} & 8.10 & 8.05 & 28.1 & 24.3 & 22.0 \\ 
+ Medicine CPT    & 8.11 & 8.11 & 30.4 & 26.6 & 30.0 \\ 
\hline
 & \textbf{Val Perplexity} & \textbf{Test Perplexity}  & \textbf{International Law}  & \textbf{Professional Law} & \textbf{Jurisprudence} \\
\hline
Overall Score \emph{l=5} & 7.92 & 8.13 & 25.5 & 33.9 & 22.2 \\ 
+ Law CPT    & 8.06 & 8.34 & 35.5 & 24.7 & 24.1 \\ 
\hline
 & \textbf{Val Perplexity} & \textbf{Test Perplexity}  & \textbf{Econometrics}  &  \textbf{High School Macroeconomics} & \textbf{Marketing} \\
\hline
Overall Score \emph{l=5} & 9.51 & 9.49 & 23.7 & 33.3 & 22.2 \\ 
+ Finance CPT    & 9.59 & 9.63 & 25.4 & 34.9 & 23.9 \\ 
\bottomrule
\end{tabular}
}
\vskip -2pt
\end{table*}
\paragraph{Training domain-specific models.}
While the \emph{Overall Score l=5} achieves the best performance, its performance in specific domains can still be improved, as shown in Table~\ref{tab:domain_specific_results}. 
To address this, we applied \ourmethod{}'s domain recognition to filter data with high \emph{Overall Score} in the \textit{medical, law, and financial} domains, and continued pre-training domain-specific models, achieving ICL performance gains in the corresponding MMLU subtasks. This validates \ourmethod{}'s capability for domain mixing.
\vskip -2pt
\subsection{Analysis of Quality Ratings} \label{sec:analysis}
\begin{figure*}[t]
    \centering
    \centerline{\includegraphics[width=1.\linewidth]{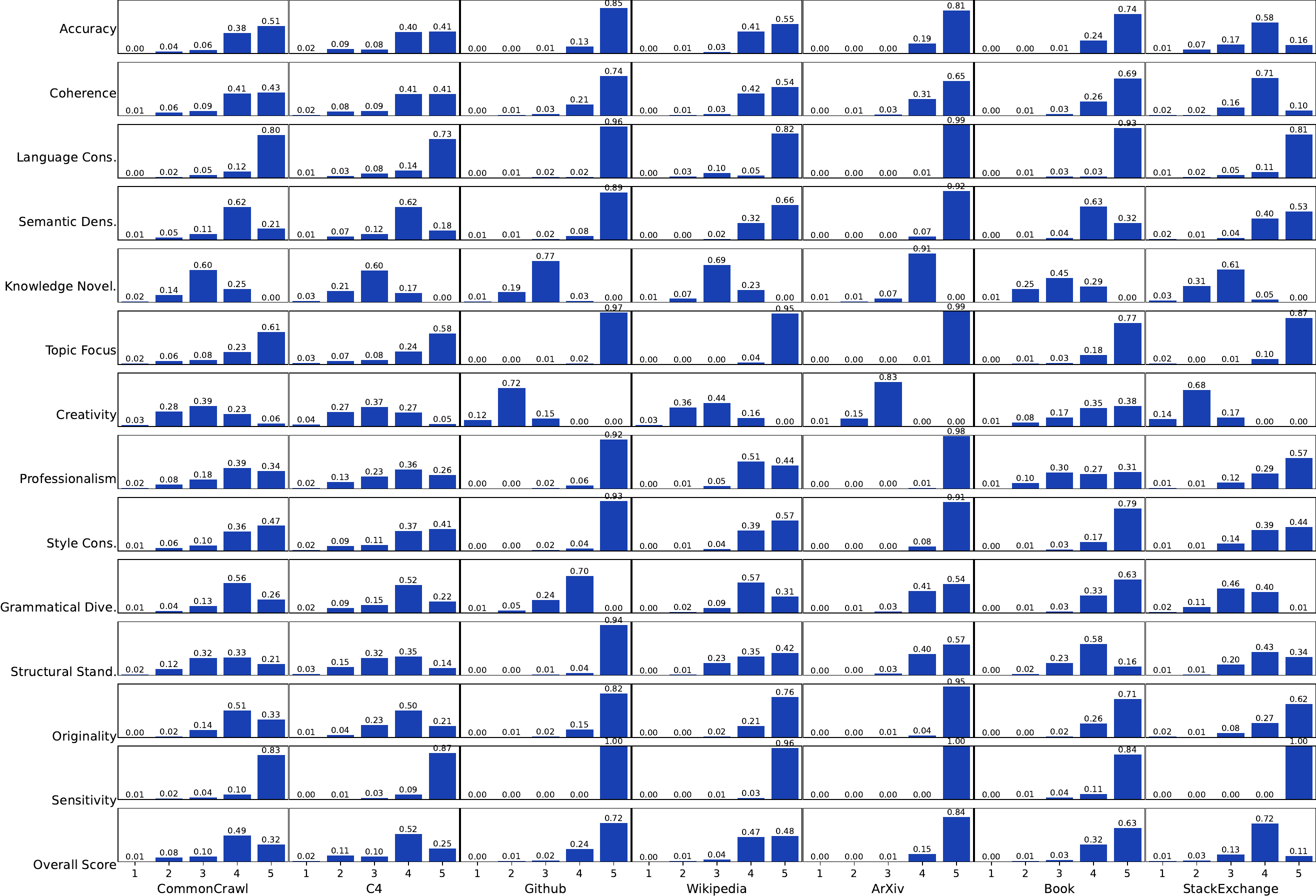}}
    \caption{The distribution of quality ratings across different sources in DataPajama}
    \label{fig:quality_rating_distribution}
\end{figure*}

\paragraph{Distribution of quality ratings.}
In Figure \ref{fig:quality_rating_distribution}, we present the distribution of quality ratings across different sources in DataPajama. 
Overall, the quality ratings for each source are primarily concentrated at 4 and 5, indicating generally high sample quality. 
This may be related to the fact that DataPajama is a subset of the curated and deduplicated slimpjama corpus. 
However, for the criteria of \emph{Knowledge Novelty} and \emph{Creativity}, there is a higher proportion of samples scoring 2 and 3, which is consistent with the lower average scores for these two criteria found in Table \ref{tab:sft_avgscore_domains}. 
Across all domains, only a few scientific domains like mathematics and medicine have \emph{Knowledge Novelty} scores above 3, while in \emph{Creativity}, only culture and entertainment scores were high at 3.64 and 3.56, respectively. 
Nevertheless, in DataPajama, the combined share of domains like mathematics and medicine is only 11.5\%, and similarly, the combined share of culture and entertainment is only 25\%, both of which are relatively small. This explains the modest ratings of the DataPajama dataset in terms of \emph{Knowledge Novelty} and \emph{Creativity}.

\begin{figure}[t]
    \centering
    \centerline{\includegraphics[width=1.\linewidth]{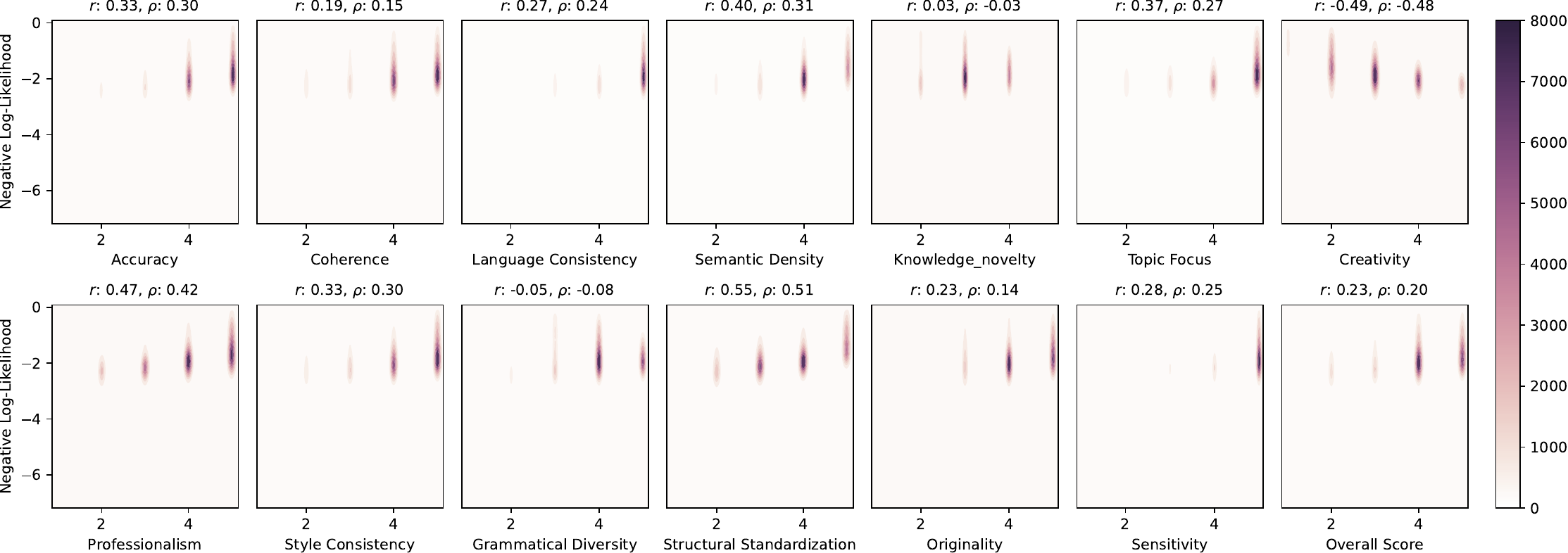}}
    \caption{Correlations of quality ratings and negative log-likelihood scores by Llama-2-7B \citep{touvron2023llama2} over 30B tokens training documents. The negative log-likelihoods are averaged over the number of tokens, and are the logarithm of the perplexity score of a single sequence. We observe that perplexity scores are not good approximations for any quality criteria.}
    \label{fig:quality_rating_nll_corr}
    \vskip -10pt
\end{figure}

\paragraph{Correlation between quality ratings and log-likelihood.}
In Figure \ref{fig:quality_rating_nll_corr}, we illustrate the correlation between quality ratings and the log-likelihood scores computed by Llama-2-7b \citep{touvron2023llama2}. Most quality criteria do not show a significant correlation with perplexity, except for the criteria of \emph{Structural standardization, Professionalism, and Creativity}, which have Spearman correlation coefficients ranging from 0.47 to 0.55, indicating a weak correlation. 
This indicates that our 14 quality criteria and sample-with-dataman method are independent of traditional perplexity metrics and filtering, indirectly showcasing the sophistication of the \emph{``reverse thinking''}.

\subsection{Data Inspection} \label{sec:inspection}
Furthermore, we examined examples of original documents from each source under Dataman's quality criteria and ratings. 
Specifically, for each criterion, we randomly selected samples with ratings ranging from 1 to 5 from different sources and presented them in the Appendix~\ref{app:raw_documents}.
Notably, these samples represent only a small snippet; nonetheless, they exhibit significant quality differences. 
We invite readers to review these differences in detail, which compares high and low ratings.
\section{Conclusion}
In this work, we introduce DataMan, a comprehensive data manager capable of quality rating and domain recognition, developed to facilitate the pre-training data selection and mixing. 
By using DataMan to annotate pre-training datasets and sampling data based on quality ratings while balancing domain diversity, our trained models demonstrate significant improvements in language modeling, task generalization, and instruction following. Through extensive experiments, we provide valuable insights for the community studying the relationship between data and large language models.
\section{Limitations} 
We acknowledge several limitations in our work. First, DataMan's reliance on LLMs for text quality assessment and domain categorization may inherit biases from these models. Second, DataMan's inference accuracy is not yet optimal, sometimes causing misclassification, highlighting the need for a large-scale collection of documents with diverse quality differences for fine-tuning. Third, using SlimPajama alone as a pre-training corpus limits result reliability, incorporating additional corpora would be better. Fourth, the model size is restricted by data and training resources, resulting in models with only 1.3B parameters, whereas increasing parameters might reveal interesting phenomena. Lastly, the considerable costs of developing data managers, data filtering, and pre-training experiments could hinder further research in this domain. 
We aim to address these in future work. Despite these limitations, DataMan remains a powerful tool for data selection and mixing.

\section*{Acknowledgements}
This work is supported by the Fundamental Research Funds for the Central Universities (226-2024-00049), the NSFC Grants (No. 62206247) and the Pioneer R\&D Program of Zhejiang (No. 2024C01035). 
\bibliography{iclr2025_conference}
\bibliographystyle{iclr2025_conference}

\clearpage
\appendix
\section{Prompt Curation}\label{app:detailed_prompts}
\newtcolorbox{promptbox}[1]{
        boxrule = 1.5pt,
        fontupper = \footnotesize,  
        fonttitle = \bf\color{black},
        arc = 5pt,
        rounded corners,
        colframe = black,
        colbacktitle = white!97!blue,
        colback = white!97!blue,
        title = #1,
}

\begin{minipage}[t]{1.0\linewidth}
    \vspace{-10pt} %
    \begin{promptbox}{Full Prompt}\label{fig:full_prompt_template}
Please carefully read and analyze the following text, score it based on fourteen evaluation criteria and their respective scoring definitions. Additionally, select the most appropriate category from the fifteen domain types that best matches the content of the text. Let's think step by step.\\
   
\textbf{Text}:\{text\}\\
    
\textbf{Domain Types:}
[A]Medicine [B]Finance [C]Law [D]Education [E]Technology [F]Entertainment [G]Mathematics [H]Coding [I]Government [J]Culture [K]Transportation [L]Retail E-commerce [M]Telecommunication [N]Agriculture [O]Other\\

\textbf{The Higher The Score, The Evaluation Criteria}:

[1]Accuracy: the fewer grammar, referential, and spelling errors the text contains, and the more accurate its expression. \_/5

[2]Coherence: the more fluent the content is expressed, and the stronger its logical coherence. \_/5

[3]Language Consistency: the more consistent the use of language in the text, with less mixing of languages. \_/5

[4]Semantic Density: the greater the proportion of valid information in the text, with less irrelevant or redundant information. \_/5

[5]Knowledge Novelty: the more novel and cutting-edge the knowledge provided by the text, with more insightful views on the industry or topic. \_/5

[6]Topic Focus: the more the text content focuses on the topic, with less deviation from the main theme. \_/5

[7]Creativity: the more creative elements are shown in the text's expression. \_/5

[8]Professionalism: the more professional terminology appears in the text, with more accurate use of terms and more professional domain-specific expression. \_/5

[9]Style Consistency: the more consistent the style of the text, with proper and appropriate style transitions. \_/5

[10]Grammatical Diversity: the more varied and correct the grammatical structures used in the text, showing a richer language expression ability. \_/5

[11]Structural Standardization: the clearer the structure followed by the text and the more standardized its format. \_/5

[12]Originality: the fewer repetitions and similar content in the text. \_/5

[13]Sensitivity: the more appropriately sensitive topics are handled in the text, with less inappropriate content. \_/5

[14]Overall Score: the better the comprehensive evaluation of the text, with superior performance in all aspects.\_/5
    \end{promptbox}
    \vspace{5pt}
\end{minipage}%

Our full prompt is shown above, where \text{\{text\}} represents the text to be annotated. Next, we will elaborate on the entire prompt curation process, including obtaining all quality criteria, domain types, and system prompts, which were conducted by experts with the assistance of Super LLM:

\paragraph{Initializing quality criteria.} 
Inspired by ``\emph{reverse thinking}''—\emph{prompting the LLM to self-identify which criteria benefit its performance}, as its pre-training capabilities are closely related to perplexity (PPL)~\citep{muennighoff2024scaling,marion2023mdatapruning}. To this end, we devised an analytical prompt for Super LLM to investigate the reasons behind textual PPL anomalies (in the top and bottom 2\%) from each source, and extract the initial quality criteria below. For clarity, Table~\ref{tab:initial_quality_criteria_case} provides several examples illustrating how these initial quality criteria are derived.

\paragraph{Enhancing quality criteria.} 
Next, we utilize Super LLM to rank the initial quality criteria by importance, eliminating, merging, and supplementing them. The resulting second-step criteria include: \textit{[1] Text accuracy (grammar, references, spelling), [2] Semantic coherence and consistency, [3] Language consistency, [4] Effective semantic content ratio, [5] Knowledge novelty, [6] Topic focus, [7] Creative expression ratio, [8] Proportion of technical terms, [9] Style variability, [10] Complexity of grammatical structures, [11] Content regularity, [12] Content redundancy, [13] Proportion of sensitive topics}. 
We then revised the prompts under the principle that higher scores indicate better criteria.
With Super LLM's assistance, we simplified the criterion names, detailed the criteria for each rating level, and ultimately derived 14 quality criteria in the paper.

\newtcolorbox{initialpromptbox}[1]{
        boxrule = 1.5pt,
        fontupper = \small,  
        fonttitle = \bf\color{black},
        arc = 5pt,
        rounded corners,
        colframe = black,
        colbacktitle = white!97!blue,
        colback = white!97!blue,
        title = #1,
}

\begin{minipage}[t]{1.\linewidth}
    \begin{initialpromptbox}{Initial Quality Criteria}\label{fig:initial_quality_criteria}

[1] Semantic Fluency/Coherence/Logic: Evaluate whether the text is smooth and easy to read, whether the content is coherent, and whether the logic is clear.

[2] Content Consistency/Variability in Language Style: Evaluate if the information within the text is contradictory and if the language style is diverse.

[3] Topic Diversity: Determine the richness and variety of topics addressed in the text.

[4] Content Regularity/Formatting: Consider whether the text adheres to a certain structure or format.

[5] Content Redundancy: Analyze the extent of information repetition within the text.

[6] Proportion of Domain-Specific Vocabulary: Measure the frequency of professional terms or specific vocabulary used in the text (such as proper nouns, technical terms, or Classical Chinese).

[7] Proportion of Sensitive Topics: Examine the percentage of content that addresses sensitive topics (e.g., involving politics, toxicity).

[8] Proportion of Creative Expression: Assess the degree of creative or innovative expression in the text (e.g., use of rhetorical techniques).

[9] Degree of Language Mixing: Analyze the extent to which different languages are used within the text (i.e., the ratio of text in various languages).

[10] Complexity of Text Structure: Evaluate the overall complexity of the text's structure.

[11] Proportion of Long Sentences: Assess the ratio of long sentences within the text.

[12] Proportion of Grammatical, Reference, and Spelling Errors: Evaluate the ratio of grammatical errors (e.g., incorrect punctuation, unclear sentence breaks), reference errors, and spelling mistakes in the text.

[13] Proportion of Content Lacking Semantics: Determine the ratio of parts within the text that lack meaningful content (e.g., garbled text, HTML tags, XML elements, navigation bars, incomplete chart numbers, or disjointed citations).
    \end{initialpromptbox}%
\end{minipage}%

\paragraph{Identifying domains.} 
We identified the 15 domain types that require assessment, based on factors such as typical application LLM industries~\citep{naveed2023comprehensive}, the number of existing industry LLMs, and the level of attention they have received, as indicated by metrics like GitHub stars\footnote{https://github.com/HqWu-HITCS/Awesome-Chinese-LLM}.

\paragraph{Chain-of-though and system prompts.}
We incorporated the chain-of-thought prompting\citep{wei2022chain}—\textit{``let's think step by step''}—while avoiding using few-shot example prompting, as the diverse sources of text could introduce rating biases. We allowed the Super LLM to generate system prompts to enhance the accuracy and confidence of quality ratings, \textit{``You are an expert to evaluate the text quality with high accuracy and confidence. Don't hesitate to use the full range of the score scale, including extreme scores if the text warrants it.''}  Results indicated that this prompt facilitated more nuanced text annotations compared to the typical prompt, \textit{``You are a helpful assistant.''}. To ensure stable output from Super LLM, we set the temperature to 0.

{\tiny
\begin{longtable}{lp{65mm}p{55mm}}
\caption{We provide several examples of excessively high and low perplexity, use analytical prompts to investigate the reasons for these anomalies, and extract initial quality criteria from the LLM's responses. The analytical prompt is:
\textit{``Read the following document, which has a \{high/low\} perplexity of \text{\{ppl\_value\}} for LLM inference. Please analyze the reasons for the PPL anomaly}.''}\label{tab:initial_quality_criteria_case}\\
\toprule
\textbf{PPL} & \textbf{Text} & \textbf{LLM's Response}  \\
\midrule
\endfirsthead

\multicolumn{3}{r}{}\\
\toprule
\textbf{PPL} & \textbf{Text} & \textbf{LLM's Response}  \\
\midrule
\endhead

\bottomrule
\multicolumn{3}{c}{Continued on next page}\\
\endfoot

\endlastfoot
    1.19 & \texttt{March: 1.77 inches- April: 5.31 inches- May: 5.9 inches- June: 5.9 inches- July: 5.9 inches- August: 2.36 inches- September: 5.31 inches- October: 5.9 inches- November: 2.36 inches- December: 2.36 inches2009, Senneville QC Precipitation: 56- January: 1.68 inches- February: 2.24 inches- March: 2.24 inches- April: 5.04 inches- May: 5.6 inches- June: 5.6 inches- July: 4.48 inches- August: 2.24 inches- September: 5.6 inches- October: 4.48 inches- November: 2.8 inches- December: 2.24 inches2008, Senneville QC Precipitation ...} & \texttt{Structural consistency: Data is presented in a fixed format. \textbf{([4] Content Regularity/Formatting)} 
    Redundancy: A large number of repeating elements. \textbf{([5]Content Redundancy)}
    Lack of complex syntax: The text lacks complex sentence structures. \textbf{([10]Complexity of Text Structure)}
    Minimal context dependence. \textbf{([1]Semantic Fluency/Coherence/Logic)}}\\
    \midrule
    1.07 & \texttt{4But the wise took oil in their vessels with their lamps. 5While the bridegroom tarried, they all slumbered and slept. 6And at midnight there was a cry made, Behold, the bridegroom cometh; go ye out to meet him. 7Then all those virgins arose, and trimmed their lamps. 8And the foolish said unto the wise, Give us of your oil; for our lamps are gone out. 9But the wise answered, saying, Not so; lest there be not enough for us and you: but go ye rather to them that sell, and buy for yourselves.  ...} & 
    \texttt{Textual Standardization: The text is an excerpt from the Bible. \textbf{([3]Topic Diversity)}
    Structural and Redundant: The narrative has structure and redundancy. \textbf{([4]Content Regularity/Formatting) ([5]Content Redundancy)} Contextual Consistency: The parts of the text have internal coherence. \textbf{([2]Content Consistency/Variability in Language Style)} Lexical and Grammatical Consistency: The Bible use fixed vocabulary and grammatical structures. \textbf{([10]Complexity of Text Structure)}}\\
    \midrule
    185.1 & \texttt{\detokenize{An rugent--bas,~d c~li:bic Hmt lakes ca,ring as, ils nwst 011ernrching mol:ival:ional ideal avoids. thfa prolbl!ern, since ii: ent11ils that peisorn, who are conll'nt 1:hrou.ghmit their lifetime to simply receive cm\"<\" from others. without acknowied!ging anid/ or recipmrnbing thrnt rnre 1in any way are jm:tnfo11bly criticized fo1r exhibitirng; a d!eplornble latck of otlr1er-regaird, and in particufar, a fad. of enga.gem.,~11t with and gratilude tow,ird the s.perifir persons who crnntribule to their Hves itl varim1s ways. Nod!dings's own view Is tbait it takes s1uprisingly little effort to disclluuge our duties to receive or \"complet,~\" 11nother p,~rso111's care: we nrny respond with an approprriale aUih1de such as recognition and/ rnr gratitude (][bid.: 19; 65) ...}} & 
    \texttt{Complex Content: The text deeply discusses philosophy and ethics. Terminology Dense: Understanding requires in-depth domain knowledge. \textbf{([6] Proportion of Domain-Specific Vocabulary)}
    Incoherent Text: Multiple parts show incoherence. \textbf{([3]Topic Diversity)}
    Random Characters: There may be encoding errors. \textbf{([12]Proportion of Grammatical, Reference, and Spelling Errors)}}\\
    \midrule
    167.5 & \texttt{\detokenize{The Indian Express is now on Telegram. Click here to join our channel (@indianexpress) and stay updated with the latest headlinesFor all the latest Kolkata News, download Indian Express App. Mönchengladbach Michael Cuisance ist vorzeitig wieder ins Training eingestiegen. Er ist fit und will in der neuen Saison mehr spielen als in der abgelaufenen. Kurzum: Der 19-Jährige will sich bei Borussia Mönchengladbach durchsetzen.Michael Cuisance ärgert sich über das frühe Aus mit Frankreich bei der U20-WM. Foto: AP/Darko VojinovicCuisance startet bei Borussia seinen Angriff auf die StartelfbGüte rwued mi tnaShcte des eh-IFtnrteoslann uaf med wudh,eelgcaNhsäcn da erd eansR eds tatlztiHnagrnispeuaps chontges reenwd lsle. ...}} & 
    \texttt{Mixed Languages: A mix of various languages and spelling systems. \textbf{([9]Degree of Language Mixing)} Incoherent Sentences: Poor connectivity between sentences. \textbf{([2]Content Consistency/Variability in Language Style)} Ambiguous Context: Lack of clear contextual information. \textbf{([13]Proportion of Content Lacking Semantics)} Formatting Errors: Contains numerous formatting issues, typos, and garbled text. \textbf{([12]Proportion of Grammatical, Reference, and Spelling Errors)}}\\
    \midrule
    72.19  &  \texttt{\detokenize{property on sand\\\\in the autumn\\\\rinsing the token and ridding it of dispiritedness a cold\\\\foam bath and icy smiles the lips permitted and the knight\\\\was refreshed then also the mountains had reviving airs\\\\nesting in the hollows from peak to golden peak and there were\\\\lairs for robbers which altered the loneliness\\\\for always he wished for his swan even its shadow even\\\\a shadow on stones that once were heavy and warm\\\\and the swan's story\\\\he cherished along with the memory of his ride that as a basin\\\\is filled then emptied yet its curve remains and its depth\\\\he would never forget nor the exact three quarter of his waltz\\\\there on the fringes of clouds and the embrace in the chasm\\\\with snow and down erasing any doubt the act was modern\\\\as mountain climbing or looking for gentians\\\\any ruin has its surplus or wrinkle there on the mask\\\\fed with dishes of rain\\\\the knight was no exception he even wore his hair like a legend\\\\whistling a lot from a need for quaintness he plied himself\\\\back and forth pretending he had an occupation like speech\\\\of course it was the same\\\\even the military and the religious wherever there was a crowd}}   & 
    \texttt{Text content is complex and varied: The text you provide looks like it is part of a series of poems or literary works, with lots of metaphors, symbols, and abstract expressions.
\textbf{([1]Semantic Fluency/Coherence/Logic)}
Irregular format and structure: Poetry often has a different format and structure than regular prose, such as line breaks, irregular punctuation, blank lines, and indentation.
\textbf{([10]Complexity of Text Structure)}
Proper nouns and archaisms: The text may contain proper nouns, archaisms, or words coined by the author, which may not be in the model's vocabulary.
\textbf{([6] Proportion of Domain-Specific Vocabulary)}
Creative expression in text: Because poetry often contains a lot of creative expression and experimental use of language, this is beyond the scope of the model's predictions based on conventional language usage.
\textbf{([8]Proportion of Creative Expression)}}\\
    \midrule
1.91 & \texttt{\detokenize{a former 1970s antiwar radical fugitive wanted for murder. After living for more than 30 years underground, Grant must now go on the run. With the FBI in hot pursuit, he sets off on a cross-country journey to track down the one person that can clear his name.\\ Shepard knows the significance of the national news story he has exposed and, for a journalist, this is an opportunity of a lifetime. Hell-bent on making a name for himself, he is willing to stop at nothing to capitalize on it. He digs deep into Grant’s past. Despite warnings from his editor and threats from the FBI, Shepard relentlessly tracks Grant across the country.\\ As Grant reopens old wounds and reconnects with former members of his antiwar group, the Weather Underground, Shepard realizes something about this man is just not adding up.}} & 
\texttt{As an AI model, I can't talk about topics involving politics, religion, pornography, violence, etc.
\textbf{([7]Proportion of Sensitive Topics)}}\\
    \midrule
    118.75  &  \texttt{\detokenize{75@dsocz7lˋp qar||-！:-！q-#,-#g- - c-(d-o 75@dsocz7lˋp qar||-！:-！q-#,-#g- - c-(d-o  75@dsocz7lˋp qar||-！:-！q-#,-#g- - c-(d-o  75@dsocz7lˋp qar||-！:-！q-#,-#g- - c-(d-o -),y-1-1p-1z-7n-8p-9d,  ar||-！:-！q-#,-#g- - c-(d-o   ar||-！:-！q-#,-#g- - c-(d-o    ar||-！:-！q-#,-#g- - c-(d-o    ar||-！:-！q-#,-#g- - c-(d-o   ar||-！:-！q-#,-#g- - c-(d-o }}   & 
    \texttt{Non-standard characters and symbols: the text contains a large number of words, such as "@"," @"," =", ";"   Such special characters and punctuation marks that are not common in normal language expression.
\textbf{([13]Proportion of Content Lacking Semantics)}
Garbled characters and random strings: occurrences in the text similar to "dsocz7l, 'p qar||-!"   Such garbled or random strings are unpredictable to the language model because they are meaningless in normal language representation.
\textbf{([13]Proportion of Content Lacking Semantics)}}\\
    \bottomrule
\end{longtable}
}
\begin{table*}[ht]
\centering
\caption{
For each quality criterion, we collected a set of documents initially rated by an independent panel and divided them into two sets of ten documents each—high-rated and low-rated—to
ensure a distinct quality gap. We used this data for prompt tuning and to validate the agreement between the prompt uses and human preferences. This table describes the sources of the documents.
}
\label{tab:prompt_validation_sources}
\setlength{\tabcolsep}{4pt}
\resizebox{1.\textwidth}{!}{
\begin{tabular}{lll}
\toprule
\multicolumn{2}{l}{Criterion} &  Sources\\

\midrule
Accuracy & \textit{High} & Academic journals, technical manuals, professional reports. \\
\cmidrule(lr){2-3}
& \textit{Low} & Social media posts, personal blogs, informal emails.\\
 
\cmidrule(lr){1-3}
Coherence & \textit{High} & News reports, research papers, essays. \\
\cmidrule(lr){2-3}
& \textit{Low} &  Forum comments, random lists, random collections of paragraphs.\\
 
\cmidrule(lr){1-3}
Language Consistency & \textit{High} & Business documents, legal contracts, academic essays.
\\
\cmidrule(lr){2-3}
& \textit{Low} & Bilingual posts, casual conversations, mixed-language blogs.\\

\cmidrule(lr){1-3}
Semantic Density & \textit{High} & Research reports, market analyses, white papers.\\
\cmidrule(lr){2-3}
& \textit{Low} & Advertising copy, social media posts, forum Q\&A.
\\

\cmidrule(lr){1-3}
Knowledge Novelty & \textit{High} & Cutting-edge research papers, conference presentations, expert interviews.\\
\cmidrule(lr){2-3}
& \textit{Low} & Common tutorials, listicles, outdated press.\\

\cmidrule(lr){1-3}
Topic Focus & \textit{High} & Specialized textbooks, academic papers on specific topics, focused industry reports.\\
\cmidrule(lr){2-3}
& \textit{Low} & Miscellaneous blog articles, off-topic comments and social media content, unthemed discussion drafts.
\\

\cmidrule(lr){1-3}
Creativity & \textit{High} & Poetry, creative writing, artistic critiques. \\
\cmidrule(lr){2-3}
& \textit{Low} & Technical documents, routine business communications, standard emails, lengthy legal texts.
\\

\cmidrule(lr){1-3}
Professionalism & \textit{High} & Formal technical reports, industry white papers, legal documents. \\
\cmidrule(lr){2-3}
& \textit{Low} & Personal blogs, informal tweets, children's literature.
\\

\cmidrule(lr){1-3}
Style Consistency & \textit{High} & Published novels, professional speeches, magazine articles. \\
\cmidrule(lr){2-3}
& \textit{Low} & Articles with mixed styles, drafts of letters, hastily written online reviews.
\\

\cmidrule(lr){1-3}
Grammatical Diversity & \textit{High} & Literary works, academic articles, formal speeches.\\
\cmidrule(lr){2-3}
& \textit{Low} & Emails composed of simple sentences, children's reading materials, transcriptions of oral presentations.
\\

\cmidrule(lr){1-3}
Structural Standardization & \textit{High} & Formal reports, standard operating procedures, structured proposals. \\
\cmidrule(lr){2-3}
& \textit{Low} & Free-form writings, scattered notes, rough drafts.
\\

\cmidrule(lr){1-3}
Originality & \textit{High} & Research reviews, detailed analyses, varied essays. \\
\cmidrule(lr){2-3}
& \textit{Low} & repetitive comments, simplistic online articles, redundant advertising copy.
\\

\cmidrule(lr){1-3}
Sensitivity & \textit{High} & generic content, informed articles on sensitive topics, guidelines. \\
\cmidrule(lr){2-3}
& \textit{Low} & Crude social media content, unthoughtful internet jokes, superficial news headlines.\\

\bottomrule
\end{tabular}
}
\end{table*}
\begin{table*}[hb]
\tiny
\caption{We follow Qurating's Table 4 \citep{wettig2024qurating} by using 10 documents from different sources, ranking them by \emph{writing style}, and using them to analyze pointwise and pairwise ratings.}
\label{tab: ten_wriqual_texts}
\vskip -0.05in
\begin{tabularx}{\textwidth}{lXp{35mm}}
\toprule
\textbf{Rank} & \textbf{Text} & \textbf{DataMan’s Annotation}  \\
\midrule
1 & \texttt{Amory Blaine inherited from his mother every trait, except the stray inexpressible few, that made him worth while. His father, an ineffectual, inarticulate man with a taste for Byron and a habit of drowsing over the Encyclopedia Britannica, grew wealthy at thirty through the death of two elder brothers, successful Chicago brokers, and in the first flush of feeling that the world was his, went to Bar Harbor and met Beatrice O'Hara. In consequence, Stephen Blaine handed down to posterity his height of ...} & accuracy: 5 coherence: 4 language\_consistency: 5 semantic\_density: 4 knowledge\_novelty: 2 topic\_focus: 5 creativity: 4 professionalism: 3 style\_consistency: 5 grammatical\_diversity: 4 structural\_standardization: 3 originality: 5 sensitivity: 5 overall\_score: 4 domain: culture \\
\midrule
2 & \texttt{Technologies for making and manipulating DNA have enabled advances in biology ever since the discovery of the DNA double helix. But introducing site-specific modifications in the genomes of cells and organisms remained elusive. Early approaches relied on the principle of site-specific recognition of DNA sequences by oligonucleotides, small molecules, or self-splicing introns. More recently, the site-directed zinc finger nucleases (ZFNs) and TAL effector nucleases (TALENs) using the principle of site-specific ...} & accuracy: 5 coherence: 5 language\_consistency: 5 semantic\_density: 5 knowledge\_novelty: 4 topic\_focus: 5 creativity: 3 professionalism: 5 style\_consistency: 5 grammatical\_diversity: 5 structural\_standardization: 4 originality: 5 sensitivity: 5 overall\_score: 5 domain: technology \\
\midrule
3 & \texttt{The winter of 1906-07 was the coldest in Alberta's history and was exacerbated by a shortage of coal. One cause of this shortage was the strained relationship between coal miners and mine operators in the province. At the beginning of April 1907, the Canada West Coal and Coke Company locked out the miners from its mine near Taber. The same company was also facing a work stoppage at its mine in the Crow's Nest Pass, where miners were refusing to sign a new contract. The problem spread until by April ...}  & accuracy: 5 coherence: 5 language\_consistency: 5 semantic\_density: 5 knowledge\_novelty: 3 topic\_focus: 5 creativity: 3 professionalism: 4 style\_consistency: 5 grammatical\_diversity: 4 structural\_standardization: 4 originality: 5 sensitivity: 5 overall\_score: 4 domain: other \\
\midrule
4 & \texttt{On December 3, Venezuela held a controversial referendum over a claim to the oil-rich Essequibo region controlled by Guyana. That same day, the Vice President of Venezuela, Delcy Rodríguez, shared a video on X, formerly Twitter, showing a group of Indigenous people lowering a Guyanese flag and hoisting a Venezuelan flag in its stead over the territory, which is also known as Guayana Esequiba. 'Glory to the brave people!' she wrote, which is the first line of the country's national anthem. The post came ...}  & accuracy: 5 coherence: 5 language\_consistency: 5 semantic\_density: 4 knowledge\_novelty: 3 topic\_focus: 5 creativity: 3 professionalism: 4 style\_consistency: 5 grammatical\_diversity: 4 structural\_standardization: 4 originality: 5 sensitivity: 5 overall\_score: 4 domain: government \\
\midrule
5 & \texttt{The Godfather is one of the most praised movies in cinema history. It gives everything that critics and audiences alike ask for in movies. In my opinion it gets all the attention it gets for being one of, or the best movies ever. One of the best things The Godfather does is its incredible casting and its iconic performances from each and every one of its characters. The actors are so convincing that it won the movie several academy awards. It also jumpstarted several actors, acting careers, and gave an ...}  & accuracy: 4 coherence: 4 language\_consistency: 5 semantic\_density: 4 knowledge\_novelty: 3 topic\_focus: 5 creativity: 4 professionalism: 3 style\_consistency: 4 grammatical\_diversity: 4 structural\_standardization: 3 originality: 5 sensitivity: 5 overall\_score: 4 domain: entertainment \\
\midrule
6 & \texttt{The food is good, but not a great value. Up front, I will just say, do not waste your time getting traditional sushi here because tbh it's not really that much better. For example, we ordered some maki and nigiri and while it was good, it wasn't that much better than our fave sushi places.   Instead, come here for their signature dishes and you'll probably be happier. We really enjoyed some of their signature dishes.   We dined as a party of 4 and we had:   Spicy edamame:  tasty and spicy!  Yellowtail ...}  & accuracy: 4 coherence: 4 language\_consistency: 5 semantic\_density: 4 knowledge\_novelty: 2 topic\_focus: 5 creativity: 3 professionalism: 2 style\_consistency: 4 grammatical\_diversity: 3 structural\_standardization: 3 originality: 4 sensitivity: 5 overall\_score: 4 domain: other \\
\midrule
7 & \texttt{My Father worked for a Forbes 500 company since the 70s. Moved up the ranks as a software engineer and management, has patents for the company that saved it millions of dollars. He's almost to pension age and suddenly HR starts making his life miserable. He noticed this trend was happening to some of his coworkers when they were getting close to age 60 as well.  HR Lady calls him into the office and says that he was not punching in and out at the correct time. My Father, an engineer, is very very ...
}  & accuracy: 4 coherence: 4 language\_consistency: 5 semantic\_density: 4 knowledge\_novelty: 3 topic\_focus: 4 creativity: 3 professionalism: 4 style\_consistency: 4 grammatical\_diversity: 4 structural\_standardization: 3 originality: 5 sensitivity: 5 overall\_score: 4 domain: technology \\
\midrule
8 & \texttt{THE ADVENTURE OF LINA AND HER ADVENTUROUS DOG SHERU Lina was a normal girl like any girl.She lived in the hills.She went to the top of the hills and she looked behind a special bush under the rearest of pine trees.She saw many pines behind it,but when she moved the pines she found a large piece of paper in which something was writen.Lina, Lina said her mother.GET UP!!You're late for school!!Oh mom!I'm too tired.Come on you have to go,no arguements.Lina was from a rich family.She lived in Los Anjilous ...
}  & accuracy: 2 coherence: 3 language\_consistency: 4 semantic\_density: 3 knowledge\_novelty: 1 topic\_focus: 4 creativity: 3 professionalism: 1 style\_consistency: 3 grammatical\_diversity: 2 structural\_standardization: 2 originality: 4 sensitivity: 5 overall\_score: 2 domain: other \\
\midrule
9 & \texttt{"Sunshine Quiz Wkly Q! Win a top Sony DVD player if u know which country the Algarve is in? Txt ansr to 82277. Â£1.50 SP: Tyrone Customer service annoncement. You have a New Years delivery waiting for you. Please call 07046744435 now to arrange delivery You are a winner U have been specially selected 2 receive Â£1000 cash or a 4* holiday (flights inc) speak to a live operator 2 claim 0871277810810 URGENT! We are trying to contact you. Last weekends draw shows that you have won a
Â£900 prize ...
}  & accuracy: 2 coherence: 3 language\_consistency: 2 semantic\_density: 3 knowledge\_novelty: 1 topic\_focus: 4 creativity: 2 professionalism: 2 style\_consistency: 2 grammatical\_diversity: 2 structural\_standardization: 2 originality: 3 sensitivity: 5 overall\_score: 2 domain: retail e-commerce \\
\midrule
10 & \texttt{cRjp7tQcwHoNERPRhj7HbiDuessoBAkl8uM0GMr3u8QsHfyGaK7x0vC3L0YGGLA7Gh240
GKhDjNwoaBtQubP8tbwrKJCSmRkUbg9aHzOQA4SLWbKcEVAiTfcQ68eQtnIF1IhOoQXLM
7RlSHBCqibUCY3Rd0ODHSvgiuMduMDLPwcOxxHCCc7yoQxXRr3qNJuROnWSuEHX5WkwNR
Sef5ssqSPXauLOB95CcnWGwblooLGelodhlLEUGI5HeECFkfvtNBgNsn5En628MrUyyFh
rqnuFNKiKkXA61oqaGe1zrO3cD0ttidD ...} & accuracy: 1 coherence: 1 language\_consistency: 1 semantic\_density: 1 knowledge\_novelty: 1 topic\_focus: 1 creativity: 1 professionalism: 1 style\_consistency: 1 grammatical\_diversity: 1 structural\_standardization: 1 originality: 1 sensitivity: 5 overall\_score: 1 domain: other \\
\bottomrule
\end{tabularx}
\end{table*}

\FloatBarrier
\section{\ourmethod{} Model} \label{app:dataman_model}
\subsection{Fine-tuning dataset}
To create the fine-tuning dataset of DataMan, we collected documents from both in-source and out-of-source within the large pre-training corpus SlimPajama \citep{cerebras2023slimpajama}. For each document, we used the full prompt to instruct the Super LLM to generate scalar scores ($l \sim [1-5]$) across 14 quality criteria, along with an $[A-O]$ letter grade to indicate its domain type. 
The fine-tuning dataset has 357k documents. 
Each document was limited to 2,048 tokens, averaging 810 tokens. It outperforms the [256, 512] token range of \cite{wettig2024qurating} when handling documents with broader length variations. 
Table~\ref{tab:sft_data_stats} lists the number and proportion of documents categorized by domain, \emph{Overall Score}, and source in the fine-tuning dataset.

\paragraph{Domain analysis.} Domain \emph{Other} accounts for nearly 25\% indicating that the fine-tuning dataset encompasses domains outside the existing 15 domains, providing \ourmethod{} with rich domain-specific prior knowledge.
Domains that account for between 15\% and 3\% involve texts related to web crawling (such as \emph{entertainment} and \emph{culture}) as well as typical vertical domains (like \emph{medicine} and \emph{coding}), enabling \ourmethod{} to better address both general and specialized knowledge.
Finally, the collection of data with high barriers to entry, such as \emph{mathematics} and long-tail \emph{telecom} data, remains a challenge for data management.

\paragraph{Overall Score analysis.} Considering the imbalance in the collected documents between high and low scores, we performed up-sampling on low-scoring documents ($< 3$) to avoid biases in the quality ratings for \ourmethod{}.
In practice, we divided the sources into five equal parts based on the difference between high- and low-scoring documents and performed a fourfold up-sampling on low-scoring documents, ultimately reaching a total dataset size of 425,794.

\paragraph{Source analysis.} While ensuring adequate data within the SlimPajama domain, we also introduced 19\% of out-of-domain data (\emph{Other}) to enhance \ourmethod{}'s source generalization capability.
\begin{table*}[h]
\centering
\caption{The number and proportion of documents categorized by domain, \emph{Overall Score}, and source in the fine-tuning dataset.}
\label{tab:sft_data_stats}
\vskip -5pt
\resizebox{0.95\linewidth}{!}{
\begin{tabular}{lrr|lrr}
\toprule
\textbf{Domains} & \# Documents &  Proportion & \textbf{Overall Score} & \# Documents & Proportion \\
\midrule
Other & 84,373     &    24.83\%    & 5.0     &      100,242     &      29.50\% \\
Technology   &  45,094     &    13.27\%  & 4.0     &      161,225     &      47.45\% \\
Entertainment   &  40,696     &    11.98\%   & 3.0     &      51,571      &     15.18\% \\
Culture   &  31,595     &     9.30\% & 2.0     &      22,423      &    6.60\% \\
Government  &   24,075     &     7.09\% & 1.0     &      4,293       &    1.26\% \\ \cline{4-6} 
Medicine   &  21,146     &     6.22\%   & \textbf{Sources} & \# Documents &  Proportion \\ \cline{4-6} 
Coding  &   19,861     &     5.85\%   & CommonCrawl & 228,000 & 63.8\%    \\
Retail E-commerce  &  16,880     &     4.97\%   & C4   &  8,000     &    2.24\%    \\
Law  &   15,989     &     4.71\%   & Wikipedia (English)   &  10,227     &     2.87\%    \\
Education   &  13,629     &     4.01\%   & Book   &  12,000     &    3.36\%    \\
Finance  &    8,915     &     2.62\%  & StackExchange  &   10,348     &     2.90\%    \\
Transportation  &    6,891     &     2.03\%  & Github  &  10,386     &    2.91\%    \\
Mathematics  &   4,875     &     1.43\%   & ArXiv  &   10,152     &     2.85\%    \\
Agriculture  &    4,627     &     1.36\%  &  --  &    --  &     --   \\ \cline{4-6} 
Telecommunication  &    1,132     &     0.33\%  & Overall  &   356,978     &     100\%    \\
\bottomrule
\end{tabular}
}
\vskip -0.1in
\end{table*}

We present the average score for quality criteria across each domain in Table \ref{tab:sft_avgscore_domains}, with analyses below:

\begin{itemize}
\item \textbf{Knowledge Novelty} excels in \emph{mathematics} and \emph{medicine}, closely linked to cutting-edge scientific research, enhancing the model’s scientific abilities.

\item \textbf{Creativity} is highest in \emph{culture} and lowest in \emph{legal} domains, reflecting the openness of literary works versus the stability of legal texts, thereby improving the model's literary skills.

\item High \textbf{professionalism} indicates data from specialized fields like\emph{mathematics}, \emph{law}, \emph{medicine}, and \emph{finance}, enhancing the model’s performance in these areas.

\item \textbf{Coding} exhibits the least \emph{grammatical diversity} and high \emph{structural standardization} due to its fixed grammatical formats. In contrast, low values in \emph{retail e-commerce} for these two criteria suggest that they lack correlation.

\item Data from specialized domains showcases strong \emph{originality} and \emph{semantic density}, with low content redundancy and meaningful content, improving the model's performance in vertical fields.

\item The \textbf{government} and \textbf{entertainment} domains exhibit lower \emph{sensitivity}, likely related to free speech on social media and politically sensitive topics, aiding the model in filtering harmful speech and sensitive content.

\item Other criteria perform well across all domains, ensuring basic requirements are met and enhancing the model’s general capabilities.

\item In general, specialized domains tend to achieve higher \emph{Overall Score}, while long-tail and general domains are relatively lower.
\end{itemize}
\begin{table*}[t]
\centering
\caption{The average score for quality criteria across each domain in the fine-tuning dataset.}
\label{tab:sft_avgscore_domains}
\vskip -5pt
\resizebox{1\linewidth}{!}{
\begin{tabular}{lcccccccccccccc}
\toprule
\textbf{Domains} & \rotatebox{90}{\textbf{Accuracy}} & \rotatebox{90}{\textbf{Coherence}} & \rotatebox{90}{\textbf{\begin{tabular}[c]{@{}l@{}}Language\\ Consistency\end{tabular}}} & \rotatebox{90}{\textbf{\begin{tabular}[c]{@{}l@{}}Semantic\\ Density\end{tabular}}} & \rotatebox{90}{\textbf{\begin{tabular}[c]{@{}l@{}}Knowledge\\ Novelty\end{tabular}}} & \rotatebox{90}{\textbf{Topic Focus}} & \rotatebox{90}{\textbf{Creativity}} & \rotatebox{90}{\textbf{Professionalism}} & \rotatebox{90}{\textbf{\begin{tabular}[c]{@{}l@{}}Style\\ Consistency\end{tabular}}} & \rotatebox{90}{\textbf{\begin{tabular}[c]{@{}l@{}}Grammatical\\ Diversity\end{tabular}}} & \rotatebox{90}{\textbf{\begin{tabular}[c]{@{}l@{}}Structural\\ Standardization\end{tabular}}} & \rotatebox{90}{\textbf{Originality}} & \rotatebox{90}{\textbf{Sensitivity}} & \rotatebox{90}{\textbf{Overall Score}} \\
\midrule
Mathematics & 4.66 & 4.59 & 4.91 & 4.82 & 3.94 & 4.90 & 2.80 & 4.86 & 4.77 & 4.26 & 4.59 & 4.84 & 5.00 & 4.71 \\
Law & 4.68 & 4.62 & 4.85 & 4.53 & 2.79 & 4.74 & 1.94 & 4.59 & 4.63 & 4.15 & 4.38 & 4.59 & 4.86 & 4.40 \\
Medicine & 4.50 & 4.50 & 4.76 & 4.40 & 3.36 & 4.65 & 2.73 & 4.39 & 4.47 & 4.21 & 4.14 & 4.47 & 4.87 & 4.33\\
Coding & 4.31 & 4.33 & 4.83 & 4.62 & 2.63 & 4.88 & 1.98 & 4.52 & 4.52 & 3.22 & 4.31 & 4.68 & 4.99 & 4.21 \\
Culture & 4.49 & 4.43 & 4.68 & 4.18 & 3.13 & 4.37 & 3.64 & 3.69 & 4.38 & 4.21 & 3.74 & 4.50 & 4.82 & 4.20\\
Agriculture & 4.50 & 4.41 & 4.78 & 4.37 & 3.14 & 4.61 & 2.83 & 4.09 & 4.42 & 3.97 & 3.92 & 4.52 & 4.96 & 4.19  \\
Education & 4.43 & 4.39 & 4.74 & 4.15 & 2.90 & 4.52 & 2.89 & 3.97 & 4.36 & 3.94 & 3.85 & 4.39 & 4.93 & 4.07\\
Government & 4.48 & 4.33 & 4.78 & 4.11 & 2.86 & 4.44 & 2.49 & 3.99 & 4.34 & 3.98 & 3.80 & 4.36 & 4.68 & 4.01 \\
Finance & 4.40 & 4.26 & 4.71 & 4.07 & 2.90 & 4.48 & 2.41 & 4.20 & 4.23 & 3.82 & 3.78 & 4.28 & 4.91 & 3.99 \\
Technology & 4.26 & 4.16 & 4.64 & 4.10 & 3.17 & 4.46 & 2.69 & 4.07 & 4.16 & 3.77 & 3.67 & 4.33 & 4.93 & 3.99 \\
Transportation & 4.34 & 4.22 & 4.70 & 4.13 & 2.73 & 4.56 & 2.57 & 3.83 & 4.21 & 3.73 & 3.66 & 4.34 & 4.95 & 3.91 \\
Telecommunication & 4.29 & 4.16 & 4.66 & 4.05 & 2.90 & 4.55 & 2.44 & 4.00 & 4.17 & 3.70 & 3.73 & 4.22 & 4.89 & 3.90\\
Entertainment & 4.16 & 4.13 & 4.46 & 3.87 & 2.68 & 4.28 & 3.56 & 3.22 & 4.07 & 3.80 & 3.37 & 4.26 & 4.62 & 3.82\\
Other & 4.11 & 4.02 & 4.47 & 3.86 & 2.60 & 4.09 & 3.08 & 3.32 & 4.01 & 3.70 & 3.34 & 4.17 & 4.65 & 3.71\\
Retail E-commerce & 4.20 & 4.02 & 4.59 & 3.91 & 2.38 & 4.43 & 2.86 & 3.46 & 4.02 & 3.52 & 3.41 & 4.14 & 4.95 & 3.70\\
\bottomrule
\end{tabular}
}
\vskip -0.1in
\end{table*}

\begin{figure*}[ht]
    \centering
    \vskip -0.1in
     \centerline{\includegraphics[width=0.75\linewidth]{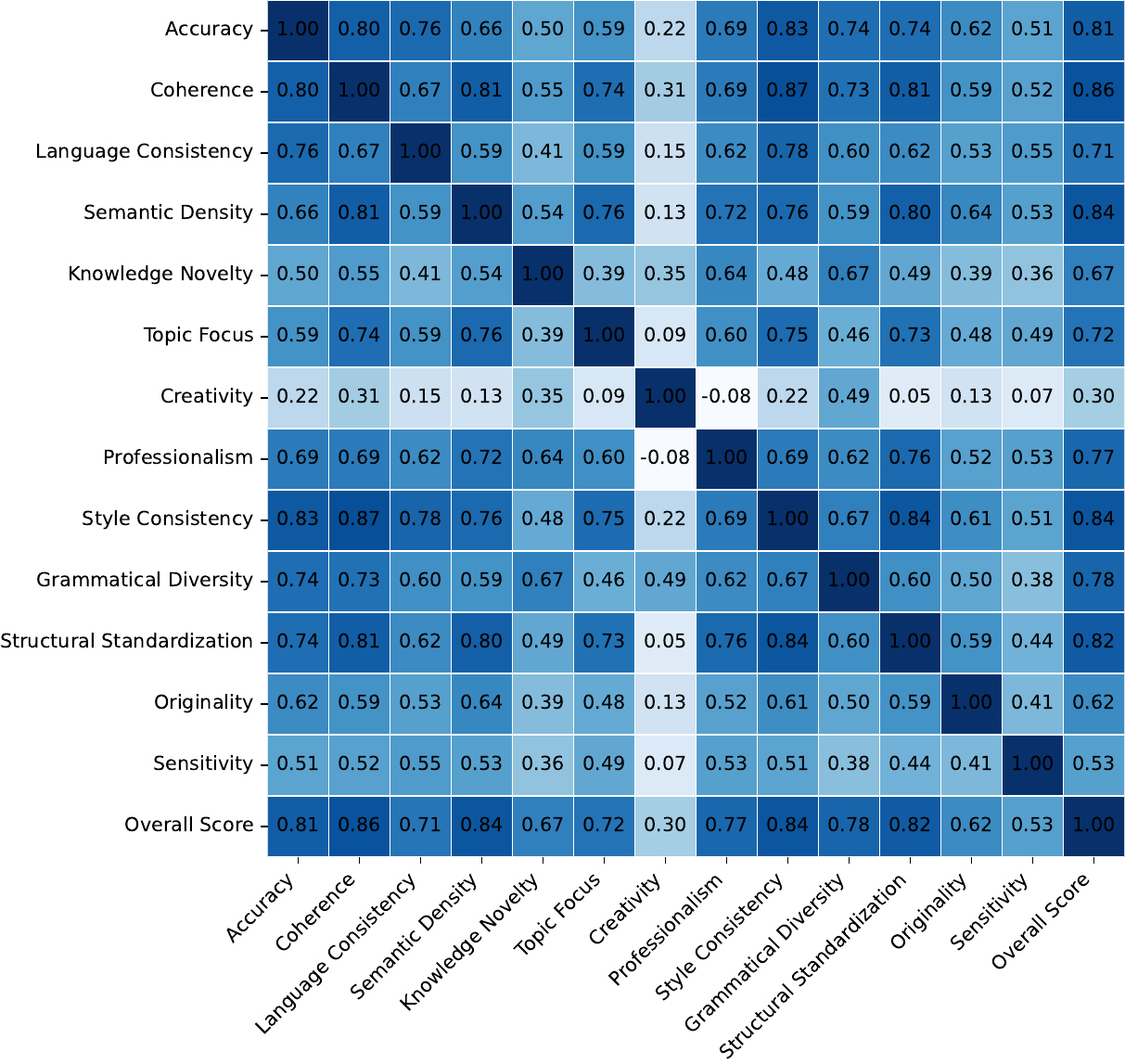}}
    \caption{Pearson correlation heatmap between 14 quality criteria in the fine-tuning dataset.}
    \vskip -0.1in
    \label{fig:quality_criteria_corr}
\end{figure*}
Figure~\ref{fig:quality_criteria_corr} shows the Pearson correlation heatmap among various quality criteria. All criteria are positively correlated, with Pearson correlation coefficients generally below 0.8, except for \emph{Style Consistency} and \emph{Structural Standardization}, which we speculate may adapt as basic requirements with other criteria. Notably, since the \emph{Overall Score} is derived from the remaining 13 quality criteria, it highly correlates to each individual criterion.

\FloatBarrier 
\subsection{\ourmethod{} Training}
We fine-tune the DataMan model using Qwen2-1.5B \citep{yang2024qwen2}, an advanced open-source 1.5B parameter language model, based on text generation loss. To meet diverse application needs, we offer three DataMan model versions, named according to their chat prompts and applicable scenarios:

\newtcolorbox{chatbox}[1]{
        boxrule = 1.5pt,
        fontupper = \small,  
        fonttitle = \bf\color{black},
        arc = 5pt,
        rounded corners,
        colframe = black,
        colbacktitle = white!97!blue,
        colback = white!97!blue,
        title = #1,
}

\begin{minipage}[t]{1.\linewidth}
    \begin{chatbox}{Chat Prompts}

\textbf{Score-only}: Please give an overall score for the text:
Text: \{text\}
Overall Score:\_/5\\

\textbf{Domain-only}: Please specify an domain type for the text:
Text: \{text\}
Domain:\_\\

\textbf{All-rating}: Please score the text on fourteen evaluation criteria and specify its domain:
Text: \{text\}
Domain:\_
[1]Accuracy:\_/5 
[2]Coherence:\_/5 
[3]Language Consistency:\_/5 
[4]Semantic Density:\_/5 
[5]Knowledge Novelty:\_/5 
[6]Topic Focus:\_/5 
[7]Creativity:\_/5 
[8]Professionalism:\_/5 
[9]Style Consistency:\_/5 
[10]Grammatical Diversity:\_/5 
[11]Structural Standardization:\_/5 
[12]Originality:\_/5 
[13]Sensitivity:\_/5 
[14]Overall Score:\_/5 
    \end{chatbox}%
\end{minipage}%

\begin{itemize}
    \item \textbf{Score-only DataMan} rates the \emph{Overall Score} of text (1 token), ideal for large-scale dataset filtering.
    \item \textbf{Domain-only DataMan} identifies text domain (1 token), ideal for large-scale data mixing.
    \item \textbf{All-rating DataMan} rates the 14 quality criteria of text and identifies the text domain (15 tokens), ideal for refined data selection and mixing.
\end{itemize}

\paragraph{Hyperparameter search.} Next, we conduct a hyperparameter search using a validation set of 8.6k documents. The search grid included: seed $\epsilon \{42, 1024, 3407\}$, learning rate $\epsilon \{1 \times 10^{-6}, 7 \times 10^{-6}, 1 \times 10^{-5}, 2 \times 10^{-5}, 5 \times 10^{-5}\}$, number of epochs $\epsilon \{2, 3, 4, 5\}$, batch size $\epsilon \{256, 512, 1024\}$, data size $\epsilon \{82k, 164k, 246k, 312k, 357k\}$, up-sampling fold $\epsilon \{1, 2, 3, 4, 5\}$, model size $\epsilon \{0.5B, 1.5B\}$, and inference temperature $\epsilon \{0.0 \text{(greedy decoding)}, 0.1, 0.3, 0.5, 0.8, 1.0\}$. 
Model selection was based on which model achieves the best accuracy on the criterion of \emph{Overall Score}.
The selected model parameters were: seed 1024, learning rate $1 \times 10^{-5}$, trained for 5 epochs, batch size 512, data size 357k, 4-fold up-sampling ratio, 1.5B model size, and utilizing greedy decoding for inference.

\paragraph{Inference accuracy.} Subsequently, we evaluated the accuracy of three DataMan model versions on a test set comprising 8.6k documents, as shown in Table~\ref{tab:sft_model_eval}.
Leveraging gold-labeled fine-tuning data from Super LLM, all model versions exhibited excellent performance. 
All-rating DataMan achieved nearly 80\% accuracy across quality criteria, with \emph{grammatical diversity} and \emph{structural standardization} being the most challenging criteria to predict.
We found performance limitations in quality rating using the \emph{Overall Score} criterion, which showed a five-class accuracy of 81.3\% and a binary accuracy of 97.5\%. The accuracy for high-quality documents reached 98.5\%, but for low-quality documents, it was only 81.6\%, due to insufficient samples. We aim to collect more low-quality documents to improve DataMan's accuracy in this area.  

\paragraph{Misclassification analysis.} In Table~\ref{tab:detailed_analysis_overall_score}, we analyze the test accuracy of the \emph{Overall Score} criterion to verify that DataMan rarely makes unreasonable decisions, making it unlikely to cause a ``snowball effect''. 
In addition to the 5-level classification accuracy (5-level Acc) used in the paper, we classify samples with a score of 3 or above as positive samples, and vice versa as negative samples, thus obtaining 2-level classification accuracy: (<3, $\geq$3 Acc). We detail the accuracy for positive samples ($\geq$3 Acc) and negative samples (<3 Acc), as well as error rates for specific misclassification cases: extreme false negative samples (<2 but $\geq$3 Error Rate), moderate false negative samples ($\geq$2, <3 but >3 Error Rate), marginal false negative samples ($\geq$2, <3 but =3 Error Rate), extreme false positive samples ($\geq$4 but <3 Error Rate), and marginal false positive samples ($\geq$3, <4 but <3 Error Rate). Results show the error rate for DataMan in the two extreme misclassification cases of the \emph{Overall Score} is very low, at just 0.2\%. This indicates that DataMan rarely mistakes poor-quality documents for high-quality ones, and vice versa. Considering the strong fault tolerance of pre-training, the snowball effect will not become a bottleneck.

\begin{table*}[t]
\centering
\renewcommand{\arraystretch}{1.2} 
\setlength{\tabcolsep}{2pt}
\caption{
The test accuracy of the three DataMan model versions. Here, \XSolidBrush indicates not applicable.
}
\label{tab:sft_model_eval}
\vskip 0.05in
\resizebox{1.\linewidth}{!}{
\begin{tabular}{l|l|lll|l|lll}
\toprule
& \textbf{\begin{tabular}[c]{@{}l@{}}Domain\\ Avg. Acc\end{tabular}}& \multicolumn{3}{c|}{\textbf{Domain Accuracy}}& \textbf{\begin{tabular}[c]{@{}l@{}}Quality\\ Avg. Acc\end{tabular}}& \multicolumn{3}{c}{\textbf{Quality Accuracy}} \\

\midrule
\multirow{10}{*}{All-rating}  & \multirow{10}{*}{86.0} & \textbf{Medicine} & \textbf{Finance} & \textbf{Law} & \multirow{10}{*}{79.2} & \textbf{Accuracy}& \textbf{Coherence}& \textbf{Language Consistency} \\ 
&& 95.8    & 89.8 & 93.4    && 78.8 & 84.1 & 76.7\\
&& \textbf{Education}& \textbf{Technology}& \textbf{Entertainment}&& \textbf{Semantic Density} & \textbf{Knowledge Novelty} & \textbf{Topic Focus} \\ 
&& 86.8    & 89.0 & 86.2    && 82.2 & 78.4 & 78.6\\
&& \textbf{Mathematics}& \textbf{Coding}& \textbf{Government} && \textbf{Creativity}   & \textbf{Professionalism}   & \textbf{Style Consistency}    \\
&& 90.4    & 90.6 & 81.0    && 79.5 & 76.8 & 76.4\\
&& \textbf{Culture}  & \textbf{Transportation} & \textbf{Retail E-commerce} && \textbf{Grammatical Diversity} & \textbf{Structural Standardization} & \textbf{Originality} \\
&& 80.2    & 77.8 & 84.9    && 73.9 & 74.8 & 92.3\\
&& \textbf{Telecommunication} & \textbf{Agriculture}    & \textbf{Other}    && \textbf{Sensitivity}  &  \textbf{Overall Score}    & -- \\
&& 87.1    & 85.8   & 83.4    && 75.6 &  81.3    & -- \\ \hline
\multirow{10}{*}{Domain-only} & \multirow{10}{*}{85.9} & \textbf{Medicine} & \textbf{Finance} & \textbf{Law} & \multirow{10}{*}{--}    & \multicolumn{3}{c}{\multirow{10}{*}{\begin{tikzpicture}
   \node[anchor=center] (r) {};
   \draw[thick] (-6,5) -- (5,9.4);
   \draw[thick] (-6,9.4) -- (5,5);
 \end{tikzpicture}}}\\
&    & 91.7    & 79.9 & 91.5    && \multicolumn{3}{l}{}   \\
&    & \textbf{Education}& \textbf{Technology}& \textbf{Entertainment}&& \multicolumn{3}{l}{}   \\
&    & 90.0    & 88.8 & 86.7    && \multicolumn{3}{l}{}   \\
&    & \textbf{Mathematics}& \textbf{Coding}& \textbf{Government} && \multicolumn{3}{l}{}   \\
&    & 87.0    & 71.6 & 86.8    && \multicolumn{3}{l}{}   \\
&    & \textbf{Culture}  & \textbf{Transportation} & \textbf{Retail E-commerce} && \multicolumn{3}{l}{}   \\
&    & 79.1    & 75.9 & 81.4    && \multicolumn{3}{l}{}   \\
&    & \textbf{Telecommunication} & \textbf{Agriculture}    & \textbf{Other}    && \multicolumn{3}{l}{}   \\
&    & 67.9    & 82.6 & 85.9    && \multicolumn{3}{l}{}   \\ \hline
\multirow{2}{*}{Score-only} & \multirow{2}{*}{--} & \multirow{2}{*}{--} & \multirow{2}{*}{--} & \multirow{2}{*}{--} & \multirow{2}{*}{77.3}    & \multirow{2}{*}{--}& \multirow{1}{*}{\textbf{Overall Score}} & \multirow{2}{*}{--} \\
 &  &  &  &  &     & & \multirow{1}{*}{77.3} & \\

\bottomrule
\end{tabular}
}
\end{table*}
\begin{table*}[t]
\centering
\renewcommand{\arraystretch}{1.2} 
\setlength{\tabcolsep}{4pt}
\caption{The detailed analysis of testing accuracy of the \emph{Overall Score} criterion.}
\label{tab:detailed_analysis_overall_score}
\resizebox{1.\textwidth}{!}{
\begin{tabular}{cccccccccc}
\toprule
 & \multicolumn{4}{c}{Accuracy} & \multicolumn{5}{c}{Error Rate} \\
\cmidrule(lr){2-5}\cmidrule(lr){6-10}
Overall Score & 5-level & $<$3, $\geq$3 & $\geq$3 & $<$3 & $<$2 but $\geq$3 &  $\geq$2, $<$3 but $>$3 & $\geq$2, $<$3 but =3 & $\geq$4 but $<$3 & $\geq$3, $<$4 but $<$3  \\
\cmidrule(lr){2-5}\cmidrule(lr){6-10}
& 81.3 & 97.5 & 98.5 & 81.6 & 0.2 & 2.7 & 15.5 & 0.2 & 1.4 \\
\bottomrule
\end{tabular}
}
\end{table*}
\begin{table*}[t]
\centering
\renewcommand{\arraystretch}{1.2} 
\caption{The inference FLOPs and memory usage of three DataMan model versions.}
\label{tab:inference_FLOPS}
\resizebox{1.\textwidth}{!}{
\begin{tabular}{ccccc}
\toprule
 & Input Speed (Toks/S) & Output Speed (Toks/S) & Processing Speed (Docs/S)  & Memory (G) \\ 
\midrule
All-rating & 31822 & 868 & 30 & 72.9 \\
Score-Only & 63644 & 1736 & 60 & 72.9 \\
Domain-Only & 63644 & 1736 & 60 & 72.9 \\
 \bottomrule
\end{tabular}
}
\end{table*}

\paragraph{Inference efficiency.} Finally, Table~\ref{tab:inference_FLOPS} presents the inference FLOPs and memory usage for three DataMan model versions evaluated on a single A800 GPU using vLLM~\citep{kwon2023efficient}. To reduce DataMan annotation costs, we recommend cost-effective models like Score-only DataMan or fine-tuned Qwen2-0.5B and suggest changing the training objective from text generation to multi-task classification. 
However, to avoid parameter transfer issues due to the different learning paradigms between pre-training and fine-tuning \citep{wang2019characterizing}, we continue using text generation loss. Furthermore, heuristic pre-processing, such as deduplication using Fuzzydedup~\citep{jiang2022fuzzydedup} and Semdedup~\citep{abbas2023semdedup}, or rule-based and model-based selection methods like C4 filter~\citep{raffel2020exploring}, Gopher rules~\citep{rae2022gopher}, and binary grammar discriminators~\citep{chowdhery2023palm}, can pre-reduce data annotation needs.
\FloatBarrier

\section{Experimental Details} \label{app:training_details}
\paragraph{\ourdata{} statistics.} Table~\ref{tab:sources_stats} shows the domain, \emph{Overall Score}, and source statistics of the 447B \ourdata{} token corpus, from which we select 30B tokens using different data selection methods.
Firstly, from a domain perspective, the proportion of the mathematics domain in \ourdata{} has significantly increased compared to the fine-tuning dataset, while the coding domain has seen a slight rise. 
In contrast, the proportions of all long-tail domains (such as Transportation, Agriculture, Retail E-commerce, and Telecommunications) still remain at the lowest levels. 
Secondly, regarding overall scores, \ourdata{}, as a subset of Slimpajama, has undergone extensive cleaning and deduplication, resulting in a high proportion of samples rated 5 and 4. 
Conversely, low-quality texts (rated below 3) account for only 7.86\%. We chose to retain these low-quality texts to allow researchers for in-depth analysis.
\textit{\ourdata{} is a curated subset of SlimPajama, which is itself a subset of RedPajama. Both SlimPajama and RedPajama are released on HuggingFace under the Apache 2.0 License.}

\paragraph{Training details.} 
Using different data selection methods, we select the 30B token subset from 
\ourdata{} and train a language model from scratch for one epoch in a randomly shuffled order.
We employ a Sheared-Llama-1.3B transformer architecture with RoPE embedding \citep{su2024roformer} and SwiGLU activations \citep{shazeer2020glu}.
This model is trained using a global batch size of 2048 sequences and a learning rate of $5\times10^{-4}$ with a cosine learning rate decay to $5\times10^{-5}$ and a linear warmup for the first $5\%$ of training steps.
We use a weight decay of $0.1$ and train with Adam \citep{kingma2014adam} with hyperparameters $\beta = (0.9, 0.95)$.
Finally, we save a checkpoint every 1,000 steps and merge the last three using mergekit \citep{goddard2024arcee} as the desired LLM, eliminating biases from step fluctuations.
Each model is trained on 32x NVIDIA A800 over 228 GPU hours. 
\begin{table*}[t]
\centering
\caption{
The number and proportion of documents categorized by domain, \emph{Overall Score}, and source in the 447B token pre-training corpus, DataPajama.
}
\label{tab:sources_stats}
\vskip 0.05in
\resizebox{0.99\linewidth}{!}{
\begin{tabular}{lrr|lrr}
\toprule
\textbf{Domains} & \# Documents &  Proportion & \textbf{Overall Score} & \# Documents & Proportion \\
\midrule
Other & 100,395,132     &    22.97\%    & 5.0     &      169,558,482     &      38.80\% \\
Culture   &  64,774,739     &     14.82\%  & 4.0     &      198,088,168     &      45.33\% \\
Technology   &  44,947,278     &    10.29\%  & 3.0     &      36,156,824      &     8.27\% \\
Entertainment   &  43,543,874     &    9.96\%  & 2.0     &      29,504,959      &    6.75\% \\
Government  &   38,157,053     &     8.73\% & 1.0     &      3,681,879       &    0.84\% \\ \cline{4-6} 
Coding  &   31,900,509     &     7.30\%   & \textbf{Sources} & \# Documents &  Proportion \\ \cline{4-6} 
Medicine   &  30,021,105     &     6.87\%   & CommonCrawl & 263,494,321 & 60.30\%    \\
Mathematics  &   20,108,505     &     4.60\%   & C4   &  70,289,855     &    16.08\%    \\
Law  &   19,463,871    &     4.45\%   & Wikipedia (English)   &  13,282,740     &     3.04\%    \\
Education   &  14,663,298     &     3.36\%   & Book   &  27,674,520     &    6.33\%    \\
Finance  &    10,138,552     &     2.32\%  & StackExchange  &   8,518,050     &     1.95\%    \\
Transportation  &    6,430,573     &     1.47\%  & Github  &  10,386     &    2.91\%    \\
Agriculture  &   5,739,330     &     1.31\%   & ArXiv  &   10,152     &     2.85\%    \\
Retail E-commerce  &  5,355,667     &     1.23\%  & Other   &  67,865    &     19.05\%    \\ \cline{4-6} 
Telecommunication  &    1,350,806    &     0.31\%  & Overall  &   436,990,312     &     100\%    \\
\bottomrule
\end{tabular}
}
\end{table*}

\paragraph{In-context learning settings.}
We choose a different number of few-shot examples per task to ensure that all demonstrations fit within the context window of 1024 tokens.
We use the following number of demonstrations (given in parentheses): ARC-easy (15), ARC-challenge (15), SciQA (2), LogiQA (2), BoolQ (0), HellaSwag (6), PIQA (6), WinoGrande (15), NQ (10), MMLU (10). We report accuracy for all tasks, except for NQ, where we report EM.
When available, we use the normalized accuracy metric provided by \texttt{lm-evaluation-harness}.

\paragraph{Full results of perplexity and ICL.} 
In Tables \ref{tab:val_ppl_results} and \ref{tab:test_ppl_results}, we report the full validation and test perplexity results for all models, including those for each RedPajama source. Table \ref{tab:icl_results} contains the ICL performance of all models across 10 downstream tasks.
The lowest perplexity reveals how quality criteria enhance LLM performance in specific data sources: 
i)-\emph{Sensitivity} excelled in web domains (CommonCrawl, C4), emphasizing the importance of avoiding sensitive topics to improve web content adaptability.
ii)-\emph{Semantic Density}, \emph{Originality}, and \emph{Topic Focus} showed superior performance in Wikipedia, suggesting the benefit of informative, original content for world knowledge absorption.
iii)-\emph{Creativity} scored lowest in book sources, indicating its role in enhancing the understanding of literature.
In ICL tasks, the criteria's influence is as follows:
i)-High \emph{Semantic Density} improved performance in elementary science tasks (ARC-E, ARC-C) and, alongside high \emph{Professionalism}, in more advanced questions (SciQ).
ii)-High \emph{Creativity} aided in summarization and subtitle tasks (HellaSw., W.G.), while complex reasoning tasks (PIQA) benefited from a mix of high \emph{Semantic Density}.
iii)-High \emph{Originality} was effective for Wikipedia-related tasks, where redundant knowledge was counterproductive.

\begin{table*}[t]
\centering
\caption{Validation per-token perplexity per RedPajama source across all our models. We highlight the best result in each column. 
Abbreviations: CC = CommonCrawl, Wiki = Wikipedia, StackEx = StackExchange.
}
\label{tab:val_ppl_results}
\setlength{\tabcolsep}{2pt}
\resizebox{1.\textwidth}{!}{
\begin{tabular}{llcccccccc}
\toprule
\multicolumn{2}{l}{\textbf{Selection Method}}  & \multicolumn{1}{c}{\textbf{CC}} & \multicolumn{1}{c}{\textbf{C4}}  & \multicolumn{1}{c}{\textbf{Github}} & \multicolumn{1}{c}{\textbf{Wiki}}  & \multicolumn{1}{c}{\textbf{ArXiv}} & \multicolumn{1}{c}{\textbf{StackEx}}  & \multicolumn{1}{c}{\textbf{Book}} & \multicolumn{1}{c}{\textbf{Overall}} \\
\midrule
\multirow{2}{*}{\begin{tabular}[c]{@{}l@{}}Uniform\end{tabular}} & & \multicolumn{1}{c}{11.09}  & \multicolumn{1}{c}{\textbf{13.93}} & \multicolumn{1}{c}{3.04}  & \multicolumn{1}{c}{10.41} & \multicolumn{1}{c}{5.69}  & \multicolumn{1}{c}{6.15}  & \multicolumn{1}{c}{12.42}  & \multicolumn{1}{c}{10.7} \\
& \textit{+50\% data} & 10.47 \gda{0.62} & 13.12 \gda{0.81}  & 2.88 \gda{0.16} & 9.43 \gda{0.98} & 5.42 \gda{0.27} & 5.84 \gda{0.31} & 11.70 \gda{0.72} & 10.09 \gda{0.61} \\
\midrule
\multirow{2}{*}{DSIR}  & \textit{with Wiki}  & 13.02 \rda{1.93} & 18.66 \rda{4.73}  & 3.62 \rda{0.58} & 24.07 \rda{13.66} & 6.63 \rda{0.94} & 7.28 \rda{1.13} & 15.39 \rda{2.97} & 13.34 \rda{2.64} \\
  & \textit{with Book } & 13.11 \rda{2.02} & 18.16 \rda{4.23}  & 3.50 \rda{0.46} & 38.97 \rda{28.56} & 6.55 \rda{0.86} & 6.83 \rda{0.68} & 13.18 \rda{0.76} & 13.60 \rda{2.90} \\
\cmidrule(lr){1-2}
\multirow{2}{*}{Perplexity} & \textit{lowest} & 16.20 \rda{5.11} & 21.51 \rda{7.58}  & 4.41 \rda{1.37} & 18.26 \rda{7.85}  & 7.12 \rda{1.43} & 9.10 \rda{2.95} & 20.26 \rda{7.84} & 15.98 \rda{5.28} \\
  & \textit{highest}  & 11.92 \rda{0.83} & 14.34 \rda{0.41}  & 3.21 \rda{0.17} & 11.38 \rda{0.97}  & 5.90 \rda{0.21} & 6.20 \rda{0.05} & 12.38 \gda{0.04} & 11.32 \rda{0.62} \\
\cmidrule(lr){1-2}
\multirow{2}{*}{\begin{tabular}[c]{@{}l@{}}Writing \\ Style\end{tabular}} & \textit{top-k} & 12.77 \rda{1.68} & 18.87 \rda{4.94}  & 3.40 \rda{0.36} & 25.61 \rda{15.20} & 5.82 \rda{0.13} & 7.03 \rda{0.88} & 12.48 \rda{0.06} & 13.01 \rda{2.31} \\
  & $\tau=2.0$ & 10.94 \gda{0.15} & 14.09 \rda{0.16}  & 2.99 \gda{0.05} & 10.32 \gda{0.09}  & 5.60 \gda{0.09} & 5.60 \gda{0.55} & 12.01 \gda{0.41} & 10.60 \gda{0.10} \\
\cmidrule(lr){1-2}
\multirow{2}{*}{\begin{tabular}[c]{@{}l@{}}Facts \& \\ Trivia\end{tabular}}  & \textit{top-k} & 12.60 \rda{1.51} & 19.15 \rda{5.22}  & 3.52 \rda{0.48} & 64.82 \rda{54.41} & 5.91 \rda{0.22} & 7.23 \rda{1.08} & 15.90 \rda{3.48} & 14.38 \rda{3.68} \\
  & $\tau=2.0$ & 10.98 \gda{0.11} & 14.25 \rda{0.32}  & 3.00 \gda{0.04} & 10.65 \rda{0.24}  & 5.56 \gda{0.13} & 6.11 \gda{0.04} & 12.32 \gda{0.10} & 10.68 \gda{0.02} \\
\cmidrule(lr){1-2}
\multirow{2}{*}{\begin{tabular}[c]{@{}l@{}}Educational \\ Value\end{tabular}}  & \textit{top-k} & 13.26 \rda{2.17} & 18.84 \rda{4.91}  & 3.45 \rda{0.41} & 27.20 \rda{16.79} & 5.63 \gda{0.06} & 6.90 \rda{0.75} & 15.45 \rda{3.03} & 13.54 \rda{2.84} \\
  & $\tau=2.0$ & 11.02 \rda{0.03} & 14.10 \rda{0.17}  & 2.98 \gda{0.06} & 10.49 \rda{0.08}  & 5.53 \gda{0.16} & 6.09 \gda{0.06} & 12.34 \gda{0.08} & 10.67 \gda{0.03} \\
\cmidrule(lr){1-2}
\multirow{2}{*}{\begin{tabular}[c]{@{}l@{}}Required \\ Expertise\end{tabular}} & \textit{top-k} & 15.13 \rda{4.04} & 21.83 \rda{7.90}  & 3.59 \rda{0.55} & 18.87 \rda{8.46}  & 5.54 \gda{0.15} & 7.63 \rda{1.48} & 16.38 \rda{3.96} & 14.97 \rda{4.27} \\
  & $\tau=2.0$ & 11.06 \rda{0.97} & 14.17 \rda{0.24}  & 2.98 \gda{0.06} & 10.25 \gda{0.16}  & 5.54 \gda{0.15} & 6.10 \gda{0.05} & 12.29 \gda{0.13} & 10.7 \\
\cmidrule(lr){1-2}
Criteria mix & $\tau=2.0$ & 10.97 \gda{0.12} & 14.10 \rda{0.17}  & 2.99 \gda{0.05} & 10.57 \rda{0.16}  & 5.56 \gda{0.13} & 6.10 \gda{0.05} & 12.19 \gda{0.23} & 10.63 \gda{0.07} \\
\midrule
Accuracy & \textit{top-k} & 10.73 \gda{0.36} & 16.59 \rda{2.66}  & 2.94 \gda{0.10} & 9.96 \gda{0.45} & 5.28 \gda{0.41} & 6.15  & 11.64 \gda{0.78} & 10.82 \rda{0.12} \\
\cmidrule(lr){1-2}
Coherence  & \textit{top-k} & 10.70 \gda{0.39} & 16.36 \rda{2.43}  & 2.90 \gda{0.14} & 9.32 \gda{1.09} & 5.27 \gda{0.42} & 6.01 \gda{0.14} & 11.48 \gda{0.94} & 10.72 \rda{0.02} \\
\cmidrule(lr){1-2}
Creativity & \textit{top-k} & 11.27 \rda{0.18} & 16.41 \rda{2.48}  & 3.19 \rda{0.15} & 9.70 \gda{0.71} & 5.38 \gda{0.31} & 6.26 \rda{0.11} & 10.87 \gda{1.55} & 11.08 \rda{0.38} \\
\cmidrule(lr){1-2}
\begin{tabular}[c]{@{}l@{}}Grammatical \\ Diversity\end{tabular} & \textit{top-k} & 10.85 \gda{0.24} & 16.72 \rda{2.79}  & 2.92 \gda{0.12} & 9.84 \gda{0.57} & 5.25 \gda{0.44} & 5.91 \gda{0.24} & 11.27 \gda{1.15} & 10.87 \rda{0.17} \\
\cmidrule(lr){1-2}
\begin{tabular}[c]{@{}l@{}}Knowledge \\ Novelty\end{tabular} & \textit{top-k} & 11.06 \rda{0.97} & 16.42 \rda{2.49}  & 2.86 \gda{0.18} & 9.59 \gda{0.82} & 5.18 \gda{0.51} & 5.87 \gda{0.28} & 12.33 \gda{0.09} & 11.01 \rda{0.31} \\
\cmidrule(lr){1-2}
\begin{tabular}[c]{@{}l@{}}Language \\ Consistency\end{tabular}  & \textit{top-k} & 10.34 \gda{0.75} & 15.43 \rda{1.50}  & 2.90 \gda{0.14} & 9.33 \gda{1.08} & 5.28 \gda{0.41} & 5.84 \gda{0.31} & 11.36 \gda{1.06} & 10.35 \gda{0.35} \\
\cmidrule(lr){1-2}
Originality & \textit{top-k} & 10.73 \gda{0.36} & 16.16 \rda{2.23}  & 2.84 \gda{0.20} & 8.98 \gda{1.43} & 5.25 \gda{0.44} & 5.82 \gda{0.33} & 11.29 \gda{1.13} & 10.68 \gda{0.02} \\
\cmidrule(lr){1-2}
Professionalism & \textit{top-k} & 11.23 \rda{0.14} & 17.00 \rda{3.07}  & 2.88 \gda{0.16} & 9.52 \gda{0.89} & 5.23 \gda{0.46} & 5.94 \gda{0.21} & 13.26 \rda{0.84} & 11.27 \rda{0.57} \\
\cmidrule(lr){1-2}
\begin{tabular}[c]{@{}l@{}}Semantic \\ Density\end{tabular} & \textit{top-k} & 11.22 \rda{0.13} & 16.61 \rda{2.68}  & 2.84 \gda{0.20} & 8.96 \gda{1.45} & 5.22 \gda{0.47} & 5.85 \gda{0.30} & 12.13 \gda{0.29} & 11.10 \rda{0.40} \\
\cmidrule(lr){1-2}
Sensitivity & \textit{top-k} & 10.30 \gda{0.79} & 14.16 \rda{0.23}  & 2.83 \gda{0.21} & 9.01 \gda{1.40} & 5.29 \gda{0.40} & 5.76 \gda{0.39} & 11.42 \gda{1.00} & \textbf{10.11 \gda{0.59}} \\
\cmidrule(lr){1-2}
\begin{tabular}[c]{@{}l@{}}Structural \\ Standardization\end{tabular} & \textit{top-k} & 12.15 \rda{1.06} & 17.95 \rda{4.02}  & 2.91 \gda{0.13} & 10.40 \gda{0.01}  & 5.37 \gda{0.32} & 6.07 \gda{0.08} & 14.72 \rda{2.30} & 12.11 \rda{1.41} \\
\cmidrule(lr){1-2}
\begin{tabular}[c]{@{}l@{}}Style \\ Consistency\end{tabular} & \textit{top-k} & 10.71 \gda{0.38} & 16.39 \rda{2.46}  & 2.92 \gda{0.12} & 9.65 \gda{0.76} & 5.28 \gda{0.41} & 5.94 \gda{0.21} & 11.42 \gda{1.00} & 10.74 \rda{0.04} \\
\cmidrule(lr){1-2}
\begin{tabular}[c]{@{}l@{}}Topic \\ Focus\end{tabular}   & \textit{top-k} & 10.48 \gda{0.61} & 15.39 \rda{1.46}  & 2.84 \gda{0.20} & 8.97 \gda{1.44} & 5.27 \gda{0.42} & 5.81 \gda{0.34} & 11.40 \gda{1.02} & 10.41 \gda{0.29} \\
\cmidrule(lr){1-2}
\multirow{5}{*}{\begin{tabular}[c]{@{}l@{}}Overall \\ Score\end{tabular}} & \textit{l=1} & 22.22 \rda{11.13}  & 24.93 \rda{11.00} & 15.58 \rda{12.54} & 69.54 \rda{59.13} & 21.71 \rda{16.02} & 19.95 \rda{13.80} & 25.41 \rda{13.00}  & 23.83 \rda{13.13}  \\
  & \textit{l=2} & 12.80 \rda{1.71} & 15.01 \rda{1.08}  & 5.40 \rda{2.36} & 18.89 \rda{8.48}  & 10.73 \rda{5.04}  & 8.57 \rda{2.42} & 14.97 \rda{2.55} & 12.84 \rda{2.14} \\
  & \textit{l=3} & 11.61 \rda{0.52} & 15.83 \rda{1.90}  & 4.05 \rda{1.01} & 14.25 \rda{3.84}  & 8.78 \rda{3.09} & 6.54 \rda{0.39} & 13.20 \rda{0.78} & 11.75 \rda{1.05} \\
  & \textit{l=4} & \textbf{10.16 \gda{0.93}} & 14.93 \rda{0.99}  & 3.15 \rda{0.11} & 9.02 \gda{1.39} & 6.05 \rda{0.36} & \textbf{5.67 \gda{0.48}} & 11.36 \gda{1.06} & 10.21 \gda{0.49} \\
  & \textit{l=5} & 10.56 \gda{0.53} & 16.36 \rda{2.43}  & \textbf{2.66 \gda{0.38}} & \textbf{8.51 \gda{1.90}} & \textbf{4.74 \gda{0.95}} & 5.92 \gda{0.23} & \textbf{10.79 \gda{1.63}} & 10.52 \gda{0.18} \\
\bottomrule
\end{tabular}
}
\end{table*}
\begin{table*}[t]
\centering
\caption{Test per-token perplexity per RedPajama source across all our models. We highlight the best result in each column. 
Abbreviations: CC = CommonCrawl, Wiki = Wikipedia, StackEx = StackExchange.}
\label{tab:test_ppl_results}
\setlength{\tabcolsep}{2pt}
\resizebox{1.\textwidth}{!}{
\begin{tabular}{llcccccccc}
\toprule
\multicolumn{2}{l}{\textbf{Selection Method}} & \textbf{CC} & \textbf{C4} & \textbf{Github} & \textbf{Wiki}  & \textbf{ArXiv} & \textbf{StackEx} & \textbf{Book} & \textbf{Overall} \\
\midrule
\multirow{2}{*}{\begin{tabular}[c]{@{}l@{}}Uniform\end{tabular}} & & 11.1  & \textbf{13.82} & 2.97 & 10.29 & 5.26 & 5.75 & 13.05 & 10.75 \\
& \textit{+50\% data} & 10.47 \gda{0.63} & 13.03 \gda{0.79} & 2.82 \gda{0.15} & 9.33 \gda{0.96} & 5.47 \rda{0.21} & 4.94 \gda{0.81} & 12.29 \gda{0.76} & 10.14 \gda{0.61} \\
\midrule
\multirow{2}{*}{DSIR} & \textit{with Wiki} & 12.97 \rda{1.87} & 18.50 \rda{4.68} & 3.50 \rda{0.53} & 23.91 \rda{13.62} & 6.71 \rda{1.45} & 6.36 \rda{0.61} & 16.11 \rda{3.06} & 13.37 \rda{2.62} \\
 & \textit{with Book} & 13.08 \rda{1.98} & 17.96 \rda{4.14} & 3.41 \rda{0.44} & 39.03 \rda{28.74} & 6.62 \rda{1.36} & 5.93 \rda{0.18} & 13.38 \rda{0.33} & 13.59 \rda{2.84} \\
\cmidrule(lr){1-2}
\multirow{2}{*}{Perplexity}  & \textit{lowest} & 16.15 \rda{5.05} & 21.34 \rda{7.52} & 4.21 \rda{1.24} & 18.14 \rda{7.85} & 7.21 \rda{1.95} & 7.43 \rda{1.68} & 21.51 \rda{8.46} & 16.04 \rda{5.29} \\
 & \textit{highest} & 11.91 \rda{0.81} & 14.27 \rda{0.45} & 3.14 \rda{0.17} & 11.24 \rda{0.95} & 5.96 \rda{0.70} & 5.34 \gda{0.41} & 12.74 \gda{0.31} & 11.34 \rda{0.59} \\
\cmidrule(lr){1-2}
\multirow{2}{*}{\begin{tabular}[c]{@{}l@{}}Writing \\ Style\end{tabular}} & \textit{top-k} & 10.95 \gda{0.15} & 18.65 \rda{4.83} & 3.33 \rda{0.36} & 25.18 \rda{14.89} & 5.85 \rda{0.59} & 6.10 \rda{0.35} & 12.81 \gda{0.24} & 12.97 \rda{2.22} \\
 & $\tau=2.0$ & 10.95 \gda{0.15} & 13.96 \rda{0.14} & 2.93 \gda{0.04} & 10.20 \gda{0.09} & 5.60 \rda{0.34} & 5.22 \gda{0.53} & 12.62 \gda{0.43} & 10.64 \gda{0.11} \\
\cmidrule(lr){1-2}
\multirow{2}{*}{\begin{tabular}[c]{@{}l@{}}Facts \& \\ Trivia\end{tabular}} & \textit{top-k} & 12.50 \rda{1.40} & 18.89 \rda{5.07} & 3.40 \rda{0.43} & 64.81 \rda{54.52} & 5.95 \rda{0.69} & 6.30 \rda{0.55} & 15.91 \rda{2.86} & 14.33 \rda{3.58} \\
 & $\tau=2.0$ & 10.98 \gda{0.12} & 14.11 \rda{0.29} & 2.93 \gda{0.04} & 10.52 \rda{0.23} & 5.60 \rda{0.34} & 5.20 \gda{0.55} & 12.99 \gda{0.06} & 10.72 \gda{0.03} \\
\cmidrule(lr){1-2}
\multirow{2}{*}{\begin{tabular}[c]{@{}l@{}}Educational \\ Value\end{tabular}}  & \textit{top-k} & 13.18 \rda{2.08} & 18.61 \rda{4.79} & 3.29 \rda{0.32} & 26.33 \rda{16.04} & 5.69 \rda{0.43} & 5.92 \rda{0.17} & 15.86 \rda{2.81} & 13.49 \rda{2.74} \\
 & $\tau=2.0$ & 11.03 \rda{0.03} & 13.97 \rda{0.15} & 2.91 \gda{0.06} & 10.36 \rda{0.07} & 5.58 \rda{0.32} & 5.17 \gda{0.58} & 12.97 \gda{0.08} & 10.72 \gda{0.03} \\
\cmidrule(lr){1-2}
\multirow{2}{*}{\begin{tabular}[c]{@{}l@{}}Required \\ Expertise\end{tabular}} & \textit{top-k} & 15.04 \rda{3.94} & 21.58 \rda{7.76} & 3.46 \rda{0.49} & 18.37 \rda{8.08} & 5.59 \rda{0.33} & 6.54 \rda{0.79} & 16.70 \rda{3.65} & 14.92 \rda{4.17} \\
 & $\tau=2.0$ & 11.07 \rda{0.97} & 14.04 \rda{0.22} & 2.91 \gda{0.06} & 10.12 \gda{0.17} & 5.59 \rda{0.33} & 5.17 \gda{0.58} & 12.91 \gda{0.14} & 10.74 \gda{0.01} \\
\cmidrule(lr){1-2}
Criteria mix  & $\tau=2.0$ & 10.97 \gda{0.13} & 13.97 \rda{0.15} & 2.92 \gda{0.05} & 10.44 \rda{0.15} & 5.61 \rda{0.35} & 5.26 \gda{0.49} & 12.82 \gda{0.23} & 10.68 \gda{0.07} \\
\midrule
Accuracy & \textit{top-k} & 10.68 \gda{0.42} & 16.42 \rda{2.60} & 2.82 \gda{0.15} & 9.84 \gda{0.45} & 5.33 \rda{0.07} & 5.20 \gda{0.55} & 12.10 \gda{0.95} & 10.80 \rda{0.05} \\
\cmidrule(lr){1-2}
Coherence & \textit{top-k} & 10.66 \gda{0.44} & 16.14 \rda{2.32} & 2.81 \gda{0.16} & 9.20 \gda{1.09} & 5.32 \rda{0.06} & 5.08 \gda{0.67} & 11.97 \gda{1.08} & 10.71 \gda{0.04} \\
\cmidrule(lr){1-2}
Creativity & \textit{top-k} & 11.12 \rda{0.02} & 16.23 \rda{2.41} & 3.09 \rda{0.12} & 9.61 \gda{0.68} & 5.42 \rda{0.16} & 5.42 \gda{0.33} & 11.39 \gda{1.66} & 11.00 \rda{0.25} \\
\cmidrule(lr){1-2}
\begin{tabular}[c]{@{}l@{}}Grammatical \\ Diversity\end{tabular} & \textit{top-k} & 10.81 \gda{0.29} & 16.53 \rda{2.71} & 2.82 \gda{0.15} & 9.72 \gda{0.57} & 5.30 \rda{0.04} & 4.97 \gda{0.78} & 11.73 \gda{1.32} & 10.86 \rda{0.11} \\
\cmidrule(lr){1-2}
\begin{tabular}[c]{@{}l@{}}Knowledge \\ Novelty\end{tabular} & \textit{top-k} & 11.01 \rda{0.01} & 16.23 \rda{2.41} & 2.81 \gda{0.16} & 9.46 \gda{0.83} & 5.22 \gda{0.04} & 4.93 \gda{0.82} & 13.00 \gda{0.05} & 11.01 \rda{0.26} \\
\cmidrule(lr){1-2}
\begin{tabular}[c]{@{}l@{}}Language \\ Consistency\end{tabular} & \textit{top-k} & 10.31 \gda{0.79} & 15.27 \rda{1.45} & 2.78 \gda{0.19} & 9.21 \gda{1.08} & 5.33 \rda{0.07} & 4.87 \gda{0.88} & 11.91 \gda{1.14} & 10.35 \gda{0.40} \\
\cmidrule(lr){1-2}
Originality & \textit{top-k} & 10.69 \gda{0.41} & 15.96 \rda{2.14} & 2.77 \gda{0.20} & 8.87 \gda{1.42} & 5.30 \rda{0.04} & 4.86 \gda{0.89} & 11.91 \gda{1.14} & 10.67 \gda{0.08} \\
\cmidrule(lr){1-2}
Professionalism & \textit{top-k} & 11.18 \rda{0.08} & 16.81 \rda{2.99} & 2.79 \gda{0.18} & 9.40 \gda{0.89} & 5.28 \rda{0.02} & 4.90 \gda{0.85} & 13.81 \rda{0.76} & 11.26 \rda{0.51} \\
\cmidrule(lr){1-2}
\begin{tabular}[c]{@{}l@{}}Semantic \\ Density\end{tabular} & \textit{top-k} & 11.16 \rda{0.06} & 16.41 \rda{2.59} & 2.78 \gda{0.19} & 8.84 \gda{1.45} & 5.27 \rda{0.01} & 4.89 \gda{0.86} & 12.83 \gda{0.22} & 11.09 \rda{0.34} \\
\cmidrule(lr){1-2}
Sensitivity & \textit{top-k} & 10.28 \gda{0.82} & 14.04 \rda{0.22} & 2.77 \gda{0.20} & 8.90 \gda{1.39} & 5.34 \rda{0.08} & 4.81 \gda{0.94} & 11.98 \gda{1.07} & \textbf{10.13 \gda{0.62}} \\
\cmidrule(lr){1-2}
\begin{tabular}[c]{@{}l@{}}Structural \\ Standardization\end{tabular} & \textit{top-k} & 12.08 \rda{0.98} & 17.76 \rda{3.94} & 2.81 \gda{0.16} & 10.25 \gda{0.04} & 5.43 \rda{0.17} & 5.12 \gda{0.63} & 15.49 \rda{2.44} & 12.11 \rda{1.36} \\
\cmidrule(lr){1-2}
\begin{tabular}[c]{@{}l@{}}Style \\ Consistency\end{tabular} & \textit{top-k} & 10.67 \gda{0.43} & 16.20 \rda{2.38} & 2.80 \gda{0.17} & 9.52 \gda{0.77} & 5.32 \rda{0.06} & 4.98 \gda{0.77} & 11.95 \gda{1.10} & 10.73 \gda{0.02} \\
\cmidrule(lr){1-2}
\begin{tabular}[c]{@{}l@{}}Topic \\ Focus\end{tabular}  & \textit{top-k} & 10.44 \gda{0.66} & 15.22 \rda{1.40} & 2.77 \gda{0.20} & 8.85 \gda{1.44} & 5.31 \rda{0.05} & 4.83 \gda{0.92} & 11.96 \gda{1.09} & 10.41 \gda{0.34} \\
\cmidrule(lr){1-2}
\multirow{5}{*}{\begin{tabular}[c]{@{}l@{}}Overall \\ Score\end{tabular}} & \textit{l=1} & 22.22 \rda{11.12} & 24.92 \rda{11.10} & 15.03 \rda{12.06} & 69.85 \rda{59.56} & 22.46 \rda{17.20} & 19.22 \rda{13.47} & 26.26 \rda{13.21} & 23.95 \rda{13.20} \\
 & \textit{l=2} & 12.83 \rda{1.73} & 14.92 \rda{1.10} & 5.22 \rda{2.25} & 18.74 \rda{8.45} & 11.00 \rda{5.74} & 7.81 \rda{2.06} & 15.74 \rda{2.69} & 12.91 \rda{2.16} \\
 & \textit{l=3} & 11.59 \rda{0.49} & 15.66 \rda{1.84} & 3.91 \rda{0.94} & 14.15 \rda{3.86} & 8.97 \rda{3.71} & 5.71 \gda{0.04} & 13.97 \rda{0.92} & 11.78 \rda{1.03} \\
 & \textit{l=4} & \textbf{10.13 \gda{0.97}} & 14.78 \rda{0.96} & 3.06 \rda{0.09} & 8.91 \gda{1.38} & 6.13 \rda{0.87} & \textbf{4.75 \gda{1.00}} & 11.93 \gda{1.12} & 10.22 \gda{0.53} \\
 & \textit{l=5} & 10.51 \gda{0.59} & 16.18 \rda{2.36} & \textbf{2.56 \gda{0.41}} & \textbf{8.40 \gda{1.89}} & \textbf{4.77 \gda{0.49}} & 4.99 \gda{0.76} & \textbf{11.28 \gda{1.77}} & 10.50 \gda{0.25} \\
\bottomrule
\end{tabular}
}
\end{table*}
\begin{table*}[t]
\centering
\caption{The in-context learning performance for ten downstream tasks across all models. We report accuracy for all tasks, except for NQ, where we report EM, and highlight the best result in each column (before rounding). 
Abbreviations: HellaSw. = HellaSwag, W.G. = WinoGrande.
}
\label{tab:icl_results}
\setlength{\tabcolsep}{1.5pt}
\resizebox{1.\textwidth}{!}{
\begin{tabular}{llccccccccccc}
\toprule
\multicolumn{2}{l}{\multirow{3}{*}{\textbf{Selection Method}}}   & \multicolumn{5}{c}{\multirow{2}{*}{\textbf{\begin{tabular}[c]{@{}c@{}}Reading\\ Comprehension\end{tabular}}}} & \multicolumn{3}{c}{\multirow{2}{*}{\textbf{\begin{tabular}[c]{@{}c@{}}Commonsense \\ Reasoning\end{tabular}}}} & \multicolumn{2}{c}{\multirow{2}{*}{\textbf{\begin{tabular}[c]{@{}c@{}}World \\ Knowledge\end{tabular}}}} & \multicolumn{1}{l}{} \\
\multicolumn{2}{l}{}   &   & \multicolumn{1}{l}{\textbf{}} & \multicolumn{1}{l}{\textbf{}} & \multicolumn{1}{l}{\textbf{}} & \multicolumn{1}{l}{\textbf{}} & & \multicolumn{1}{l}{\textbf{}} & \multicolumn{1}{l}{\textbf{}} & & \multicolumn{1}{l}{} & \multicolumn{1}{l}{} \\
\cmidrule(lr){3-7}\cmidrule(lr){8-10}\cmidrule(lr){11-12}
\multicolumn{2}{l}{}   & \textbf{ARC-E} (15)   & \textbf{ARC-C} (15) & \textbf{SciQA} (2)   & \textbf{LogiQA} (2) & \textbf{BoolQ} (0)  & \textbf{HellaSw.} (6)  & \textbf{PIQA} (6)   & \textbf{W.G.} (15)  & \textbf{NQ} (10) & \textbf{MMLU} (5)  & \textbf{Average}   \\
\midrule
\multirow{2}{*}{\begin{tabular}[c]{@{}l@{}}Uniform \end{tabular}} &  & 57.5 & 27.6 & 87.7 & 24.1 & 57.5 & 44 & 68.6 & 52.5 & 4.1 & 25.7  & 44.9  \\
& \textit{+50\% data} & 60.6 \gua{3.1} & 29.3 \gua{1.7}  & 90.3 \gua{2.6}  & 24.4 \gua{0.3}  & 60.1 \gua{2.6}  & 47.7 \gua{3.7} & 69.0 \gua{0.4}  & 54.4 \gua{1.9}  & 5.8 \gua{1.7}   & 26.1 \gua{0.4} & 46.8 \gua{1.9} \\
\midrule
\multirow{2}{*}{DSIR}   & \textit{with Wiki} & 52.8 \rua{4.7} & 26.3 \rua{1.3}  & 85.9 \rua{1.8}  & 25.2 \gua{1.1}  & 60.3 \gua{2.8}  & 35.8 \rua{8.2} & 61.4 \rua{7.2}  & 52.2 \rua{0.3}  & 4.7 \gua{0.6}   & 24.7 \rua{1.0} & 42.9 \rua{2.0} \\
  & \textit{with Book} & 49.5 \rua{8.0} & 25.3 \rua{2.3}  & 83.6 \rua{4.1}  & 23.5 \rua{0.6}  & 57.9 \gua{0.4}  & 44.8 \gua{0.8} & 69.4 \gua{0.8}  & 55.6 \gua{3.1}  & 3.1 \rua{1.0}   & 25.2 \rua{0.5} & 43.8 \rua{1.1} \\
\cmidrule(lr){1-2}
\multirow{2}{*}{Perplexity} & \textit{lowest} & 49.2 \rua{8.3} & 25.1 \rua{2.5}  & 83.7 \rua{4.0}  & 22.0 \rua{2.1}  & 61.4 \gua{3.9}  & 34.6 \rua{9.4} & 65.0 \rua{3.6}  & 49.1 \rua{3.4}  & 2.7 \rua{1.4}   & 24.7 \rua{1.0} & 41.7 \rua{3.2} \\
  & \textit{highest}   & 53.5 \rua{4.0} & 25.6 \rua{2.0}  & 84.6 \rua{3.1}  & 26.1 \gua{2.0}  & 58.0 \gua{0.5}  & 41.6 \rua{2.4} & 65.6 \rua{3.0}  & 53.4 \gua{0.9}  & 2.9 \rua{1.2}   & 24.0 \rua{1.7} & 43.5 \rua{1.4} \\
\midrule
\multirow{2}{*}{\begin{tabular}[c]{@{}l@{}}Writing \\ Style\end{tabular}}  & \textit{top-k} & 52.7 \rua{4.8} & 27.3 \rua{0.3}  & 79.7 \rua{8.0}  & 26.4 \gua{2.3}  & 60.5 \gua{3.0}  & 41.1 \rua{2.4} & 66.1 \rua{2.5}  & 52.3 \rua{0.2}  & 2.5 \rua{1.6}   & 24.4 \rua{1.3} & 43.4 \rua{1.5} \\
  & $\tau=2.0$ & 56.4 \rua{1.1} & 28.4 \gua{0.8}  & 85.8 \rua{1.8}  & 24.9 \gua{0.8}  & 59.3 \gua{1.8}  & 44.9 \gua{0.9} & 68.6 & 55.8 \gua{1.3}  & 4.5 \gua{0.4}   & 23.8 \rua{1.9} & 45.0 \gua{0.1} \\
\cmidrule(lr){1-2}
\multirow{2}{*}{\begin{tabular}[c]{@{}l@{}}Facts \& \\ Trivia\end{tabular}} & \textit{top-k} & 65.6 \gua{8.1} & 33.1 \gua{5.5}  & 87.9 \gua{0.2}  & 24.1 & 60.9 \gua{3.4}  & 39.4 \rua{4.6} & 62.5 \rua{6.1}  & 53.1 \gua{0.6}  & 5.7 \gua{1.6}   & 25.3 \rua{0.4} & 45.8 \gua{0.9} \\
  & $\tau=2.0$ & 59.3 \gua{1.8} & 29.8 \gua{2.2}  & 88.1 \gua{0.4}  & 25.0 \gua{0.9}  & 61.4 \gua{3.9}  & 43.9 \rua{0.1} & 68.3 \rua{0.3}  & 54.6 \gua{2.1}  & 4.4 \gua{0.3}   & 26.9 \gua{1.2} & 46.2 \gua{1.3} \\
\cmidrule(lr){1-2}
\multirow{2}{*}{\begin{tabular}[c]{@{}l@{}}Educational \\ Value\end{tabular}}  & \textit{top-k} & \textbf{66.6 \gua{9.1}}  & \textbf{34.6 \gua{7.0}} & 89.6 \gua{1.9}  & 24.6 \gua{0.5}  & 58.3 \gua{0.8}  & 45.5 \gua{1.5} & 66.4 \rua{2.2}  & 52.9 \gua{0.4}  & 3.8 \rua{0.3}   & 25.0 \rua{0.7} & 46.7 \gua{1.8} \\
  & $\tau=2.0$ & 60.7 \gua{3.2} & 30.4 \gua{2.8}  & 88.8 \gua{1.1}  & 26.6 \gua{2.5}  & 60.1 \gua{2.6}  & 45.4 \gua{1.4} & 69.1 \gua{0.5}  & 54.2 \gua{1.7}  & 4.3 \gua{0.2}   & \textbf{27.1 \gua{1.4}}   & 46.7 \gua{1.8} \\
\cmidrule(lr){1-2}
\multirow{2}{*}{\begin{tabular}[c]{@{}l@{}}Required \\ Expertise\end{tabular}} & \textit{top-k} & 60.4 \gua{2.9} & 30.9 \gua{3.3}  & 86.8 \rua{0.9}  & 25.0 \gua{0.9}  & 60.9 \gua{3.4}  & 36.1 \rua{7.9} & 57.8 \rua{10.8} & 52.2 \rua{0.3}  & 2.4 \rua{1.7}   & 26.3 \gua{0.6} & 43.9 \rua{1.0} \\
  & $\tau=2.0$ & 59.6 \gua{2.1} & 29.8 \gua{2.2}  & 89.0 \gua{1.3}  & 23.8 \rua{0.3}  & 61.4 \gua{3.9}  & 43.2 \rua{0.8} & 67.4 \rua{1.2}  & 56.0 \gua{3.5}  & 4.6 \gua{0.5}   & 25.4 \rua{0.3} & 46.0 \gua{1.1} \\
\cmidrule(lr){1-2}
Criteria mix & $\tau=2.0$ & 59.2 \gua{1.7} & 30.2 \gua{2.6}  & 88.0 \gua{0.3}  & 24.3 \gua{0.2}  & 58.7 \gua{1.2}  & 44.5 \gua{0.5} & 68.7 \gua{0.1}  & 53.5 \gua{1.0}  & 5.3 \gua{1.2}   & 25.1 \rua{0.6} & 45.7 \gua{0.8} \\
\midrule
Accuracy & \textit{top-k} & 62.8 \gua{5.3} & 29.4 \gua{1.8}  & 90.3 \gua{2.6}  & 24.7 \gua{0.6}  & 61.7 \gua{4.2}  & 49.8 \gua{5.8} & 69.3 \gua{0.7}  & 55.6 \gua{3.1}  & 7.0 \gua{2.9}   & 26.2 \gua{0.5} & 47.7 \gua{0.3} \\
\cmidrule(lr){1-2}
Coherence & \textit{top-k} & 63.6 \gua{6.1} & 32.5 \gua{4.9}  & 90.3 \gua{2.6}  & 26.7 \gua{2.6}  & 61.6 \gua{4.1}  & 50.8 \gua{6.8} & 70.6 \gua{2.0}  & 55.1 \gua{2.6}  & 7.0 \gua{2.9}   & 25.2 \rua{0.5} & 48.3 \gua{0.6} \\
\cmidrule(lr){1-2}
Creativity & \textit{top-k} & 60.8 \gua{3.3} & 30.6 \gua{3.0}  & 87.8 \gua{0.1}  & 24.3 \gua{0.2}  & 61.4 \gua{3.9}  & \textbf{51.9 \gua{7.9}}   & \textbf{71.2 \gua{2.6}} & \textbf{58.6 \gua{6.1}} & 4.7 \rua{1.4}   & 25.6 \rua{0.1} & 47.7 \gua{0.6} \\
\cmidrule(lr){1-2}
\begin{tabular}[c]{@{}l@{}}Grammatical \\ Diversity\end{tabular} & \textit{top-k} & 64.6 \gua{7.1} & 33.4 \gua{5.8}  & 89.3 \gua{1.6}  & 26.3 \gua{2.2}  & 62.0 \gua{4.5}  & 50.6 \gua{6.6} & 69.8 \gua{1.2}  & 56.1 \gua{3.6}  & 7.7 \gua{3.6}   & 25.2 \rua{0.5} & 48.5 \gua{0.5} \\
\cmidrule(lr){1-2}
\begin{tabular}[c]{@{}l@{}}Knowledge \\ Novelty\end{tabular} & \textit{top-k} & 63.5 \gua{6.0} & 32.8 \gua{5.2}  & 90.5 \gua{2.8}  & 24.3 \gua{0.2}  & \textbf{62.1 \gua{4.6}} & 47.2 \gua{3.2} & 67.9 \rua{0.7}  & 55.6 \gua{3.1}  & 6.2 \rua{0.3}   & 24.8 \rua{0.9} & 47.5 \gua{0.2} \\
\cmidrule(lr){1-2}
\begin{tabular}[c]{@{}l@{}}Language \\ Consistency\end{tabular} & \textit{top-k} & 63.0 \gua{5.5} & 31.0 \gua{3.4}  & 89.7 \gua{2.0}  & 25.3 \gua{1.2}  & 61.4 \gua{3.9}  & 50.2 \gua{6.2} & 70.1 \gua{1.5}  & 57.6 \gua{5.1}  & 7.6 \gua{3.5}   & 25.8 \rua{0.6} & 48.2 \gua{0.3} \\
\cmidrule(lr){1-2}
Originality & \textit{top-k} & 64.0 \gua{6.5} & 31.7 \gua{4.1}  & 90.7 \gua{3.0}  & 25.3 \gua{1.2}  & 57.7 \gua{0.2}  & 49.0 \gua{5.0} & 70.5 \gua{1.9}  & 56.2 \gua{3.7}  & \textbf{8.0 \gua{3.9}}  & 24.7 \rua{1.0} & 47.8 \gua{0.3} \\
\cmidrule(lr){1-2}
Professionalism & \textit{top-k} & 64.4 \gua{6.9} & 32.2 \gua{4.6}  & 91.1 \gua{3.4}  & 24.0 \rua{0.1}  & 61.2 \gua{3.7}  & 45.0 \gua{1.0} & 66.1 \rua{2.5}  & 53.3 \gua{0.8}  & 6.6 \rua{0.9}   & 25.2 \rua{0.6} & 46.9 \gua{0.1} \\
\cmidrule(lr){1-2}
\begin{tabular}[c]{@{}l@{}}Semantic \\ Density\end{tabular} & \textit{top-k} & 66.2 \gua{8.7} & 31.9 \gua{4.3}  & \textbf{91.4 \gua{3.7}} & 25.2 \gua{1.1}  & 57.2 \rua{0.3}  & 48.4 \gua{4.4} & \textbf{71.2 \gua{2.6}} & 54.7 \gua{2.2}  & 7.5 \gua{3.4}   & 25.9 \rua{0.2} & 48.0 \gua{0.1} \\
\cmidrule(lr){1-2}
Sensitivity  & \textit{top-k} & 63.2 \gua{5.7} & 32.4 \gua{4.8}  & 91.1 \gua{3.4}  & 25.5 \gua{1.4}  & 61.3 \gua{3.8}  & 50.4 \gua{6.4} & 70.6 \gua{2.0}  & 56.6 \gua{4.1}  & 6.8 \rua{0.7}   & 25.3 \rua{0.4} & 48.3 \gua{0.3} \\
\cmidrule(lr){1-2}
\begin{tabular}[c]{@{}l@{}}Structural \\ Standardization\end{tabular}  & \textit{top-k} & 62.1 \gua{4.6} & 31.9 \gua{4.3}  & 89.8 \gua{2.1}  & 25.3 \gua{1.2}  & 59.3 \gua{1.8}  & 45.9 \gua{1.9} & 70.6 \gua{2.0}  & 54.5 \gua{2.0}  & 7.1 \gua{3.0}   & 27.0 \gua{1.3} & 47.4 \gua{0.1} \\
\cmidrule(lr){1-2}
\begin{tabular}[c]{@{}l@{}}Style \\ Consistency\end{tabular} & \textit{top-k} & 63.0 \gua{5.5} & 32.0 \gua{4.4}  & 90.3 \gua{2.6}  & \textbf{28.7 \gua{4.6}} & 61.6 \gua{4.1}  & 50.3 \gua{6.3} & 70.7 \gua{2.1}  & 57.8 \gua{5.3}  & 7.2 \gua{3.1}   & 25.1 \rua{0.6} & 48.7 \gua{0.3} \\
\cmidrule(lr){1-2}
\begin{tabular}[c]{@{}l@{}}Topic \\ Focus\end{tabular}   & \textit{top-k} & 61.6 \gua{4.1} & 30.8 \gua{3.2}  & 91.2 \gua{3.5}  & 27.3 \gua{3.2}  & 61.9 \gua{4.4}  & 50.0 \gua{6.0} & 69.0 \gua{0.4}  & 56.3 \gua{3.8}  & 7.1 \gua{3.0}   & 24.0 \rua{1.0} & 47.9 \gua{0.3} \\
\cmidrule(lr){1-2}
\multirow{5}{*}{\begin{tabular}[c]{@{}l@{}}Overall \\ Score\end{tabular}}  & \textit{l=1} & 42.0 \rua{15.5}   & 23.0 \rua{4.6}  & 69.4 \rua{18.3} & 25.7 \gua{1.6}  & 55.2 \rua{2.3}  & 31.0 \rua{13.0} & 61.3 \rua{7.3}  & 50.3 \rua{2.2}  & 0.8 \rua{3.3}   & 25.4 \gua{0.2} & 38.4 \rua{6.6} \\
  & \textit{l=2} & 53.7 \rua{3.8} & 26.0 \rua{1.6}  & 83.8 \rua{3.9}  & 26.4 \gua{2.3}  & 61.8 \gua{4.3}  & 38.2 \rua{5.8} & 63.9 \rua{4.7}  & 50.5 \rua{2.0}  & 4.3 \gua{0.2}   & 25.0 \rua{0.70}   & 43.4 \rua{1.5} \\
  & \textit{l=3} & 54.3 \rua{3.2} & 26.2 \rua{1.4}  & 87.1 \rua{0.6}  & 24.1 & 62.0 \gua{4.5}  & 42.0 \rua{2.0} & 68.3 \rua{0.3}  & 52.0 \rua{0.5}  & 4.7 \gua{0.} & 25.7  & 44.6 \rua{0.3} \\
  & \textit{l=4} & 60.7 \gua{3.2} & 31.3 \gua{3.7}  & 90.6 \gua{2.9}  & 24.1 & 60.8 \gua{3.3}  & 51.3 \gua{6.3} & 71.0 \gua{2.4}  & 57.9 \gua{5.4}  & 7.7 \gua{3.6}   & 24.2 \rua{0.5} & 47.9 \gua{0.2} \\
  & \textit{l=5} & 66.1 \gua{8.6} & 34.0 \gua{6.4}  & 90.7 \gua{3.0}  & 26.1 \gua{2.0}  & 59.2 \gua{1.7}  & 51.5 \gua{6.5} & 70.7 \gua{2.1}  & 58.3 \gua{5.8}  & 7.8 \gua{3.7}   & 26.9 \gua{1.2} & \textbf{49.1 \gua{1.6}}   \\
\bottomrule
\end{tabular}
}
\end{table*}

\begin{table*}[h]
\centering
\caption{The Sample-with-\ourmethod{} model (\emph{Overall Score l=5}) improve perplexity and in-context learning (ICL) results on \textbf{larger 60B tokens}. We report the validation, test perplexity, and ICL performance of 10 downstream tasks. We highlight the best result in each column and improvement over uniform sampling with the 60B token budget.}
\label{tab:main_results_60B}
\setlength{\tabcolsep}{4pt}
\resizebox{1.\textwidth}{!}{
\begin{tabular}{lllccccc}
\toprule
\multicolumn{2}{l}{\textbf{Selection Method}} & \multicolumn{1}{c}{\begin{tabular}[c]{@{}c@{}}\textbf{Val} \\ \textbf{Perplexity} \\\end{tabular}}&\multicolumn{1}{c}{\begin{tabular}[c]{@{}c@{}}\textbf{Test} \\ \textbf{Perplexity} \\\end{tabular}}& \multicolumn{1}{c}{\begin{tabular}[c]{@{}c@{}}\textbf{Reading} \\ \textbf{Comprehension} \\ \textit{(5 tasks)}\end{tabular}} & \multicolumn{1}{c}{\begin{tabular}[c]{@{}c@{}}\textbf{Commonsense} \\ \textbf{Reasoning} \\ \textit{(3 tasks)}\end{tabular}} & \multicolumn{1}{c}{\begin{tabular}[c]{@{}c@{}}\textbf{World} \\ \textbf{Knowledge} \\ \textit{(2 tasks)}\end{tabular}} & \multicolumn{1}{c}{\begin{tabular}[c]{@{}c@{}}\\\textbf{Average} \\ \textit{(10 tasks)}\end{tabular}} \\
\midrule
Uniform &  & 10.81 & 10.79 & 53.7 & 58.4 & 16.4 & 47.6 \\
Educational Value & $\tau=2.0$ & \textbf{9.81 \gda{1.00}} & \textbf{9.85 \gda{0.94}} & 54.2 \gua{0.5} & 58.7 \gua{0.3} & 16.0 \rda{0.4} & 47.9 \gua{0.3} \\
Overall Score & \textit{l=5} & 9.93 \gda{0.88} & 9.91 \gda{0.88} & \textbf{56.5 \gua{2.8}} & \textbf{62.9 \gua{4.5}} & \textbf{17.5 \gua{1.1}} & \textbf{50.6 \gua{3.0}} \\
\bottomrule
\end{tabular}
}
\end{table*}

\begin{table*}[h]
\centering
\caption{Validation per-token perplexity per RedPajama source across \textbf{three models trained on 60B tokens}. We highlight the best result in each column. 
Abbreviations: CC = CommonCrawl, Wiki = Wikipedia, StackEx = StackExchange.
}
\label{tab:val_ppl_results_60B}
\setlength{\tabcolsep}{4pt}
\resizebox{1.\textwidth}{!}{
\begin{tabular}{llcccccccc}
\toprule
\multicolumn{2}{l}{\textbf{Selection Method}}  & \multicolumn{1}{c}{\textbf{CC}} & \multicolumn{1}{c}{\textbf{C4}}  & \multicolumn{1}{c}{\textbf{Github}} & \multicolumn{1}{c}{\textbf{Wiki}}  & \multicolumn{1}{c}{\textbf{ArXiv}} & \multicolumn{1}{c}{\textbf{StackEx}}  & \multicolumn{1}{c}{\textbf{Book}} & \multicolumn{1}{c}{\textbf{Overall}} \\
\midrule
Uniform &  & 10.81  & \textbf{16.50} & 2.89  & 9.54 & 5.22  & 5.88  & 11.43  & 10.81 \\
Educational Value & $\tau=2.0$ & 10.15 \gda{0.66} & \textbf{12.94 \gda{3.56}}  & 2.78 \gda{0.11} & 9.11 \gda{0.43}  & 5.20 \gda{0.02} & 5.68 \gda{0.20} & 11.36 \gda{0.07} & 9.81 \gda{1.00} \\
Overall Score & \textit{l=5} & \textbf{9.94 \gda{0.87}} & 15.58 \gda{0.92}  & \textbf{2.52 \gda{0.37}} & \textbf{7.71 \gda{1.83}} & \textbf{4.53 \gda{0.69}} & \textbf{5.62 \gda{0.26}} & \textbf{10.13 \gda{1.3}} & \textbf{9.93 \gda{0.88}} \\
\bottomrule
\end{tabular}
}
\end{table*}

\begin{table*}[h]
\centering
\caption{Test per-token perplexity per RedPajama source across \textbf{three models trained on 60B tokens}. We highlight the best result in each column. 
Abbreviations: CC = CommonCrawl, Wiki = Wikipedia, StackEx = StackExchange.
}
\label{tab:test_ppl_results_60B}
\setlength{\tabcolsep}{4pt}
\resizebox{1.\textwidth}{!}{
\begin{tabular}{llcccccccc}
\toprule
\multicolumn{2}{l}{\textbf{Selection Method}}  & \multicolumn{1}{c}{\textbf{CC}} & \multicolumn{1}{c}{\textbf{C4}}  & \multicolumn{1}{c}{\textbf{Github}} & \multicolumn{1}{c}{\textbf{Wiki}}  & \multicolumn{1}{c}{\textbf{ArXiv}} & \multicolumn{1}{c}{\textbf{StackEx}}  & \multicolumn{1}{c}{\textbf{Book}} & \multicolumn{1}{c}{\textbf{Overall}} \\
\midrule
Uniform &  & 10.76  & 16.31 & 2.78  & 9.41 & 5.27  & 4.93  & 11.92  & 10.79 \\
Educational Value & $\tau=2.0$ & 10.16 \gda{0.60} & \textbf{12.83 \gda{3.48}} & 2.73 \gda{0.05} & 8.99 \gda{0.42} & 5.25 \gda{0.02} & 4.76 \gda{0.17} & 11.92 & 9.85 \gda{0.94} \\
Overall Score & \textit{l=5} & \textbf{9.90 \gda{0.86}} & 15.40 \gda{0.91} & \textbf{2.43 \gda{0.35}} & \textbf{7.61 \gda{1.80}} & \textbf{4.56 \gda{0.71}} & \textbf{4.68 \gda{0.25}} & \textbf{10.61 \gda{1.31}} & \textbf{9.91 \gda{0.88}} \\
\bottomrule
\end{tabular}
}
\end{table*}

\begin{table*}[h]
\centering
\caption{The in-context learning performance for ten downstream tasks across \textbf{three models trained on 60B tokens}. We report accuracy for all tasks, except for NQ, where we report EM, and highlight the best result in each column (before rounding). 
Abbreviations: HellaSw. = HellaSwag, W.G. = WinoGrande.}
\label{tab:icl_results_60B}
\setlength{\tabcolsep}{2pt}
\resizebox{1.\textwidth}{!}{
\begin{tabular}{llccccccccccc}
\toprule
\multicolumn{2}{l}{\multirow{3}{*}{\textbf{Selection Method}}}   & \multicolumn{5}{c}{\multirow{2}{*}{\textbf{\begin{tabular}[c]{@{}c@{}}Reading\\ Comprehension\end{tabular}}}} & \multicolumn{3}{c}{\multirow{2}{*}{\textbf{\begin{tabular}[c]{@{}c@{}}Commonsense \\ Reasoning\end{tabular}}}} & \multicolumn{2}{c}{\multirow{2}{*}{\textbf{\begin{tabular}[c]{@{}c@{}}World \\ Knowledge\end{tabular}}}} & \multicolumn{1}{l}{} \\
\multicolumn{2}{l}{}   &   & \multicolumn{1}{l}{\textbf{}} & \multicolumn{1}{l}{\textbf{}} & \multicolumn{1}{l}{\textbf{}} & \multicolumn{1}{l}{\textbf{}} & & \multicolumn{1}{l}{\textbf{}} & \multicolumn{1}{l}{\textbf{}} & & \multicolumn{1}{l}{} & \multicolumn{1}{l}{} \\
\cmidrule(lr){3-7}\cmidrule(lr){8-10}\cmidrule(lr){11-12}
\multicolumn{2}{l}{}   & \textbf{ARC-E} (15)   & \textbf{ARC-C} (15) & \textbf{SciQA} (2)   & \textbf{LogiQA} (2) & \textbf{BoolQ} (0)  & \textbf{HellaSw.} (6)  & \textbf{PIQA} (6)   & \textbf{W.G.} (15)  & \textbf{NQ} (10) & \textbf{MMLU} (5)  & \textbf{Average}   \\
\midrule
Uniform &   & 63.0 & 32.4 & 89.1 & 22.3 & \textbf{61.6} & 49.4 & 70.7 & 55.0 & 6.4 & 26.3  & 47.6  \\
Educational Value & $\tau=2.0$ & 64.5 \gua{1.5} & 31.1 \rda{1.3}  & 90.9 \gua{1.8}  & 24.1 \gua{1.8}  & 60.6 \rda{1.0}  & 50.2 \gua{0.8} & 69.9 \rda{0.8}  & 56.0 \gua{1.0}  & 7.0 \gua{0.6}   & 25.0 \rda{1.3}   & 47.9 \gua{0.3} \\
Overall Score & \textit{l=5} & \textbf{68.6 \gua{5.6}} & \textbf{36.8 \gua{4.4}}  & \textbf{91.9 \gua{2.8}} & \textbf{24.4 \gua{2.1}}  & 60.6 \rda{1.0}  & \textbf{56.1 \gua{6.7}} & \textbf{72.4 \gua{1.7}}  & \textbf{60.1 \gua{5.1}}  & \textbf{8.6 \gua{2.2}}   & \textbf{26.4 \gua{0.1}} & \textbf{50.6 \gua{3.0}} \\
\bottomrule
\end{tabular}
}
\end{table*}

\paragraph{Results on larger 60B tokens.}
Despite the 30B token subset exceeding the compute-optimal ratio of data-to-model suggested by \citep{hoffmann2022training}, to solidify our method, we used the strongest DataMan variant \emph{Overall Score l=5}, the existing SOTA baseline (education value $\tau=2.0$), and uniform sampling to select a larger 60B subset to train the 1.3B language model from scratch. 
The model's PPL and ICL performance of 60B subset in Table~\ref{tab:main_results_60B}, the full validation and test PPL across sources in Tables~\ref{tab:val_ppl_results_60B}, ~\ref{tab:test_ppl_results_60B}, and the full ICL performance of ten downstream tasks in Table~ \ref{tab:icl_results_60B}. The Sample-with-DataMan model significantly outperformed the SOTA baseline in ICL performance tasks and showed modest improvement in the validation and test PPL. This further confirms the effectiveness of the DataMan approach.

\begin{figure*}[h]
    \centering
    \vskip 0.1in
    \centerline{\includegraphics[width=1.0\linewidth]{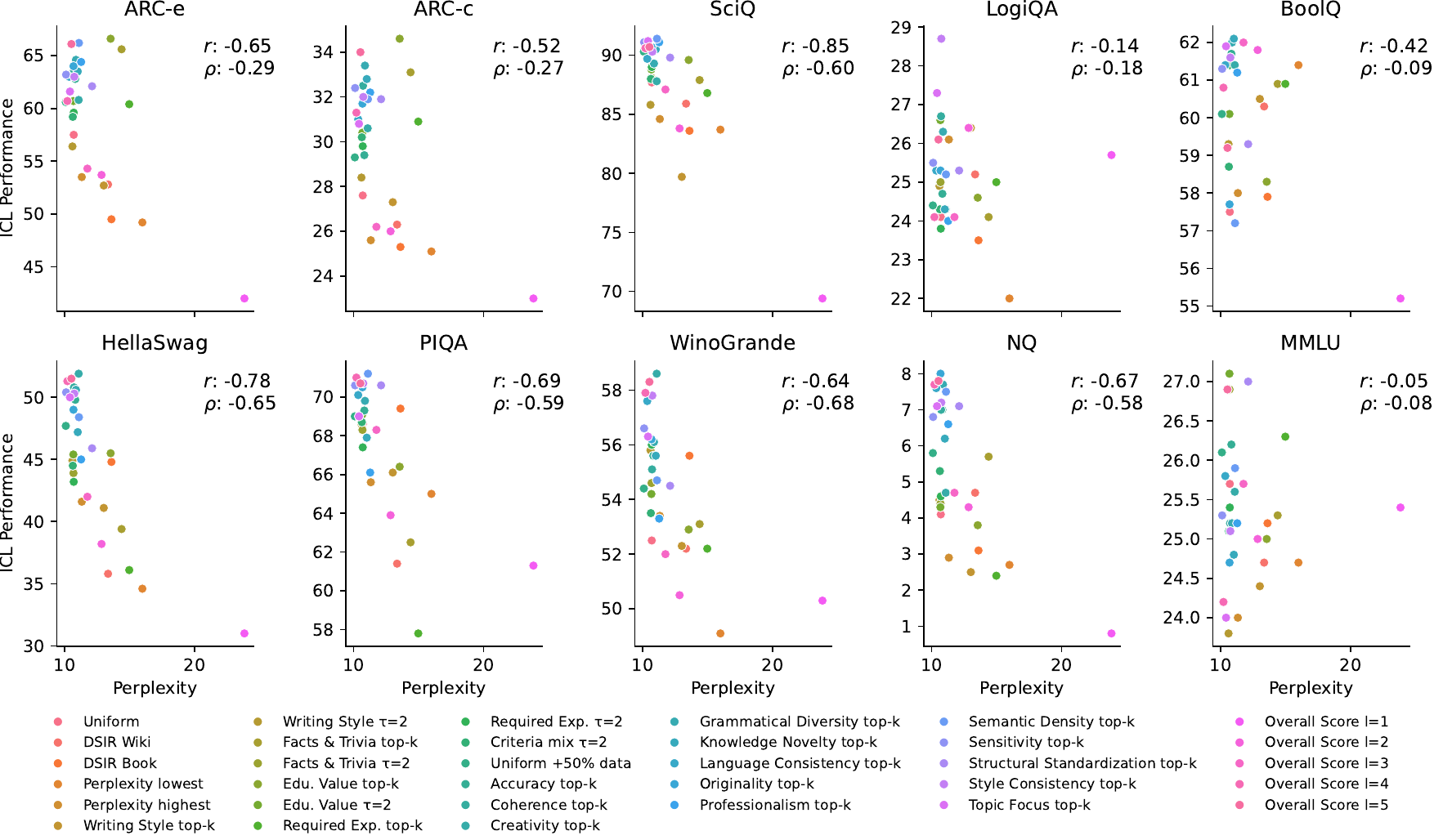}}
    \caption{We plot the relationship between perplexity and in-context learning (ICL) performance for all models across 10 downstream tasks, as shown in Tables \ref{tab:val_ppl_results} and \ref{tab:icl_results}, along with the Pearson and Spearman correlation coefficients. The misalignment between perplexity and ICL is most pronounced in the LogiQA and MMLU tasks, we guess, possibly due to domain mismatch and ICL task complexity.} 
    \label{fig:icl_vs_ppl}
    \vskip -0.1in
\end{figure*}

\paragraph{Misalignment between PPL and ICL.} 
In Figure~\ref{fig:icl_vs_ppl}, we plot the relationship between perplexity and ICL performance for all models across 10 downstream tasks, including the Pearson and Spearman correlation coefficients, to investigate the misalignment between PPL and ICL. 
The results indicate that the misalignment is most pronounced in the LogiQA and MMLU tasks. Deeper analysis identifies two main causes:
\textit{i)-domain mismatch:} pre-training often uses extensive general corpora, which enables the model to exhibit lower perplexity on a common text. However, tasks like MMLU, which span 57 distinct specialized domains (such as abstract algebra and anatomy), may suffer in ICL performance due to domain mismatch;
\textit{ii)-ICL task complexity:} Many ICL tasks require complex reasoning rather than simple text generation, which perplexity assessment struggles to capture. This is particularly evident in LogiQA, where the task assesses human logical reasoning skills through expert-written questions from Civil Servants' Exams.

\FloatBarrier
\section{Inspecting Raw Documents and Ratings} \label{app:raw_documents}

Finally, we present snippets from the raw documents of Wikipedia, Books, Stack Exchange, Github, ArXiv, CommonCrawl, and C4 subsets of \ourdata{}. 
These documents correspond to samples with quality ratings of 1, 2, 3, 4, and 5 across 14 quality criteria, as shown in Figure~\ref{fig:quality_rating_distribution}. 
Notably, although these samples represent just a small random snippet, they display significant quality differences. We believe it is essential to provide an unfiltered view of the training data; therefore, we have not applied any filtering to these documents.
\textbf{\textcolor{red!90!black}{A small number of documents contain \%potentially sensitive content.}}

From the visual examples, we found a notable distinction between the data rated 1 and 2 and those rated 3, 4, and 5.  For instance, the score of 1 corresponds to an example like \textit{``...83 510 l s 311 548 m 305 546 l 301 540 l 299 530 l 299 ...''}, whereas the score of 5 reflects \textit{``...system recognizes a hierarchy of events from the measurements, not exactly in the sense of physical reality...''}  However, the discrepancy between scores of 4 and 5 is not as pronounced, as seen in the example \textit{``...have been augmented with terms that quantify the user satisfaction or the ad relevance...,''} which corresponds to a score of 4.  This further supports our rationale for choosing pointwise evaluation over pairwise, as humans also find it challenging to determine superiority based on subtle differences.

In terms of domain adaptability, most of the evaluation criteria we established are semantic-focused, allowing for effective differentiation of documents within the C4 domain.  However, we also observed that our criteria have some relevance in the code domain (e.g., GitHub).  Specifically, code that features more detailed comments and follows structural conventions tends to receive higher scores, while disordered code is typically rated lower.

\begin{table*}[t]
    \centering
    \caption{Raw training examples selected to have 14 quality ratings from 1 to 5 within ArXiv.}
    \centerline{\includegraphics[width=\linewidth,trim={0 30pt 0 0}]{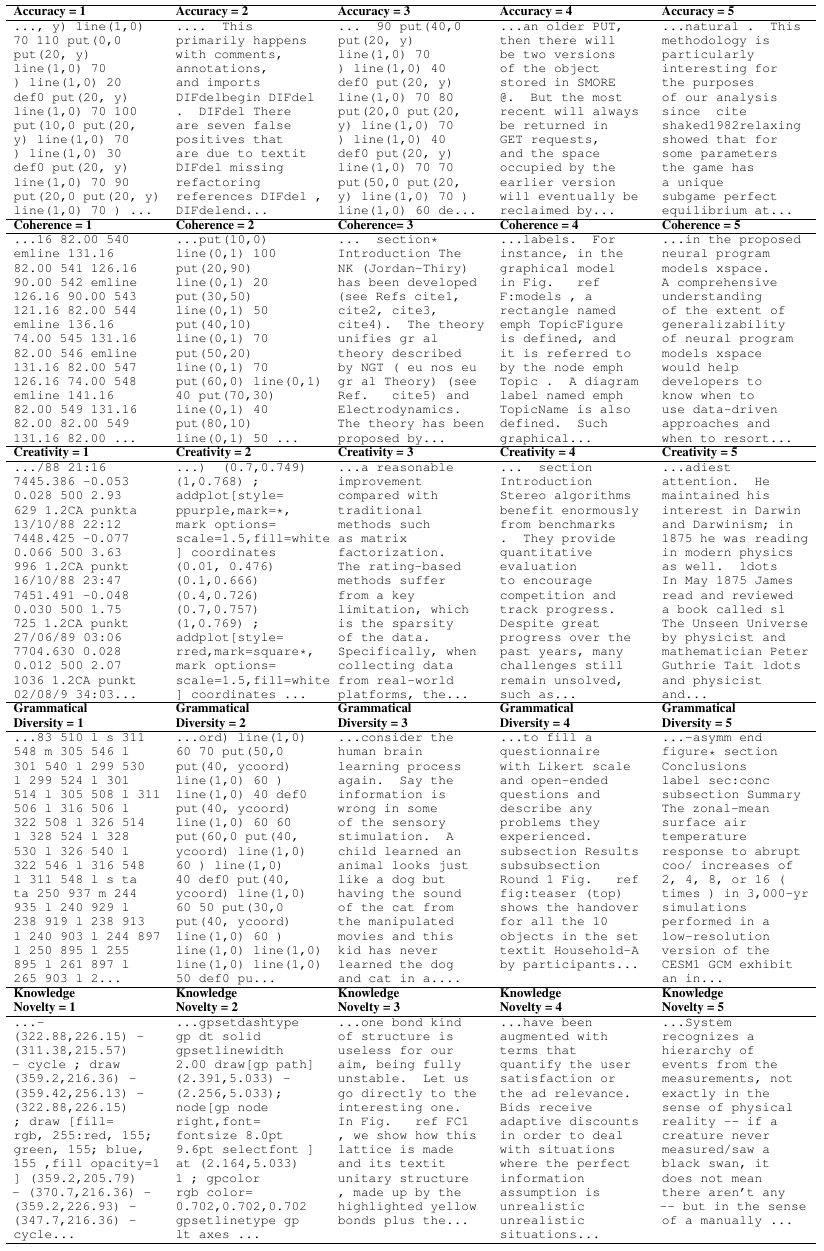}}
    \label{fig:Case_Study_1}
\end{table*}
    
\begin{table*}[t]
    \centering
    \caption{Raw training examples selected to have 14 quality ratings from 1 to 5 within ArXiv.}
    \centerline{\includegraphics[width=\linewidth,trim={0 30pt 0 0}]{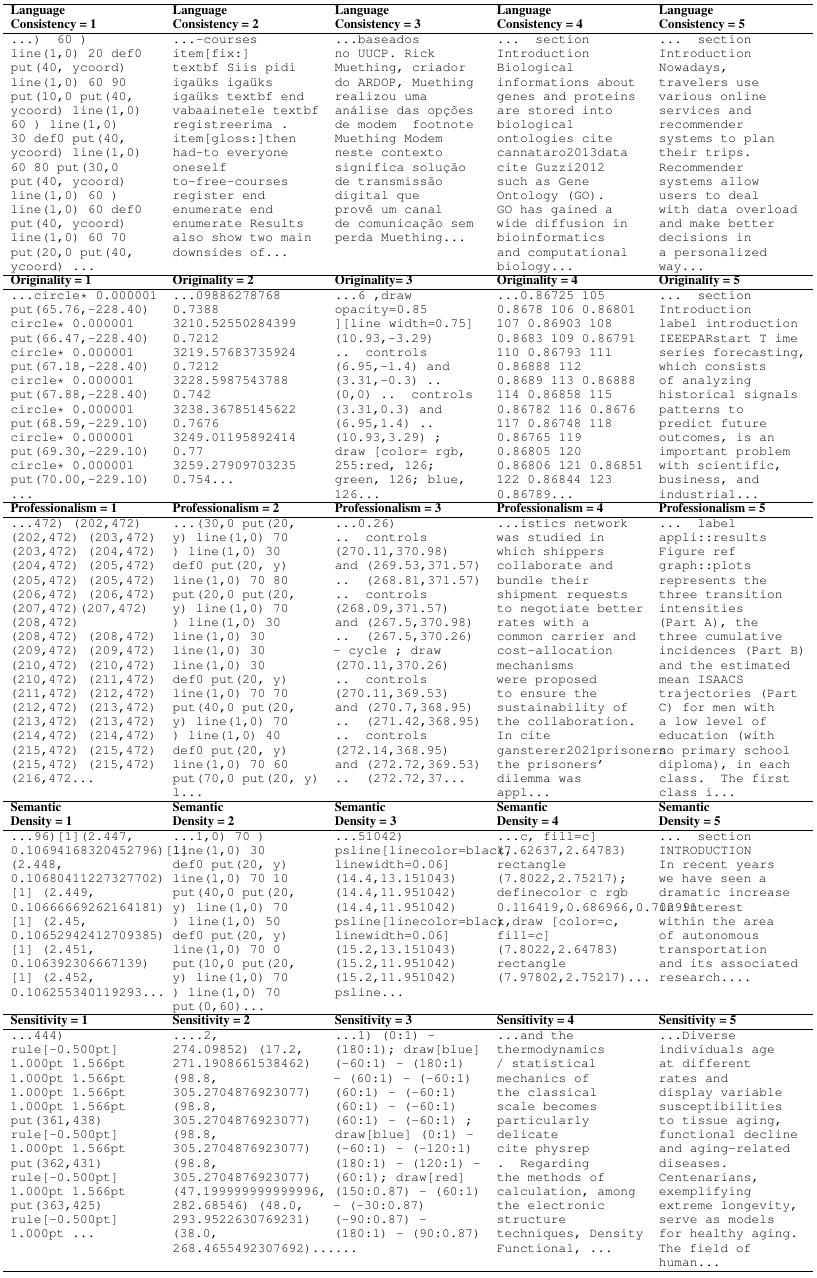}}
    \label{fig:Case_Study_2}
\end{table*}
    
\begin{table*}[t]
    \centering
    \caption{Raw training examples selected to have 14 quality ratings from 1 to 5 within ArXiv.}
    \centerline{\includegraphics[width=\linewidth,trim={0 30pt 0 0}]{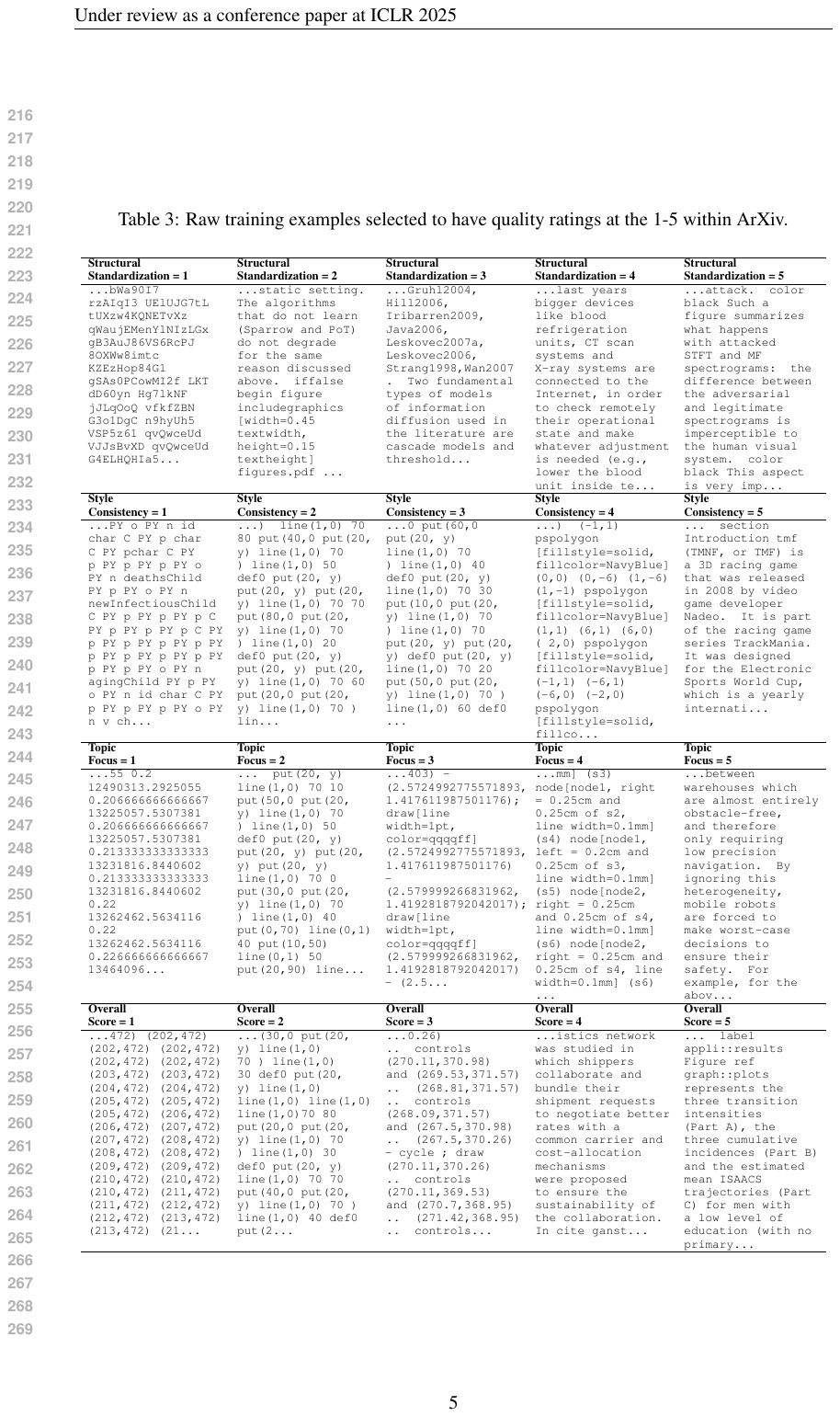}}
    \label{fig:Case_Study_3}
\end{table*}
    
\begin{table*}[t]
    \centering
    \caption{Raw training examples selected to have 14 quality ratings from 1 to 5 within Book.}
    \centerline{\includegraphics[width=\linewidth,trim={0 30pt 0 0}]{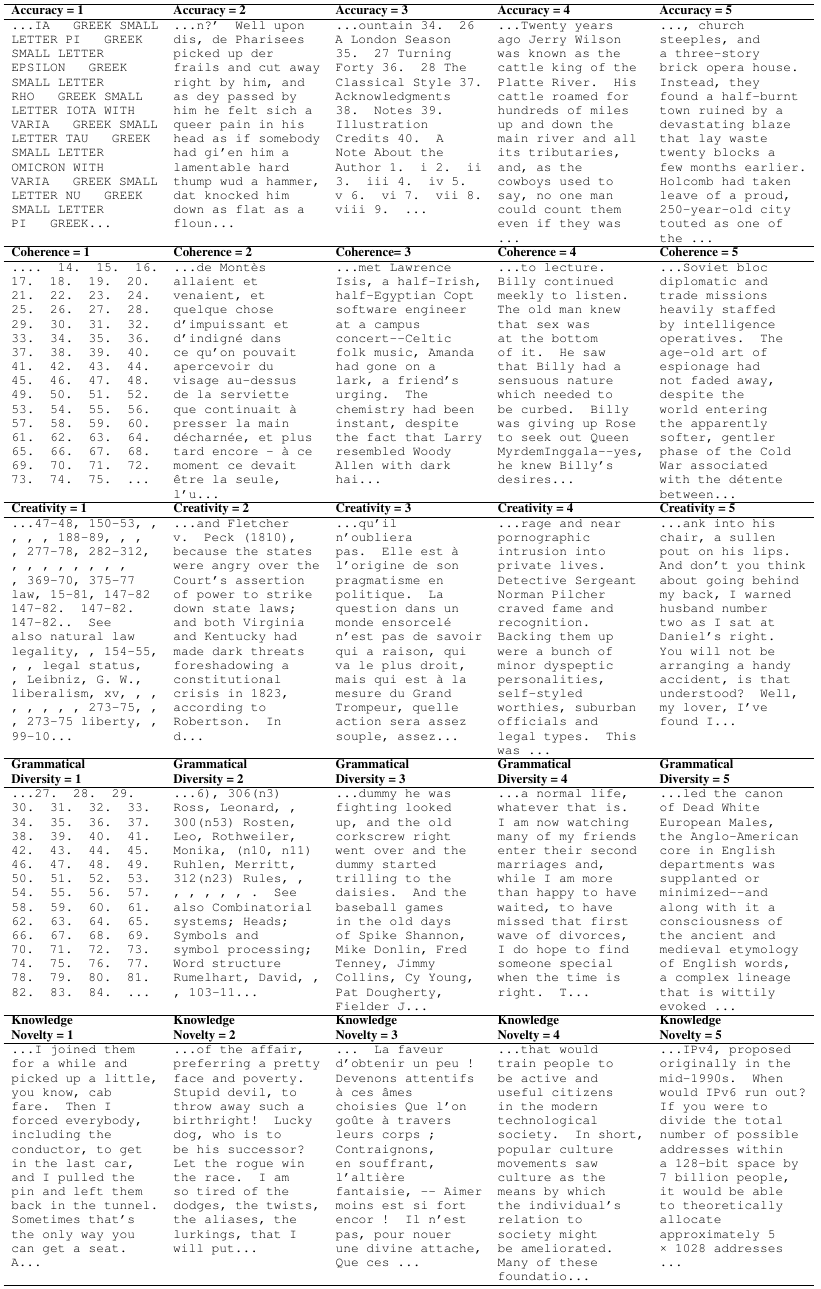}}
    \label{fig:Case_Study_4}
\end{table*}

\begin{table*}[t]
    \centering
    \caption{Raw training examples selected to have 14 quality ratings from 1 to 5 within Book.}
    \centerline{\includegraphics[width=\linewidth,trim={0 30pt 0 0}]{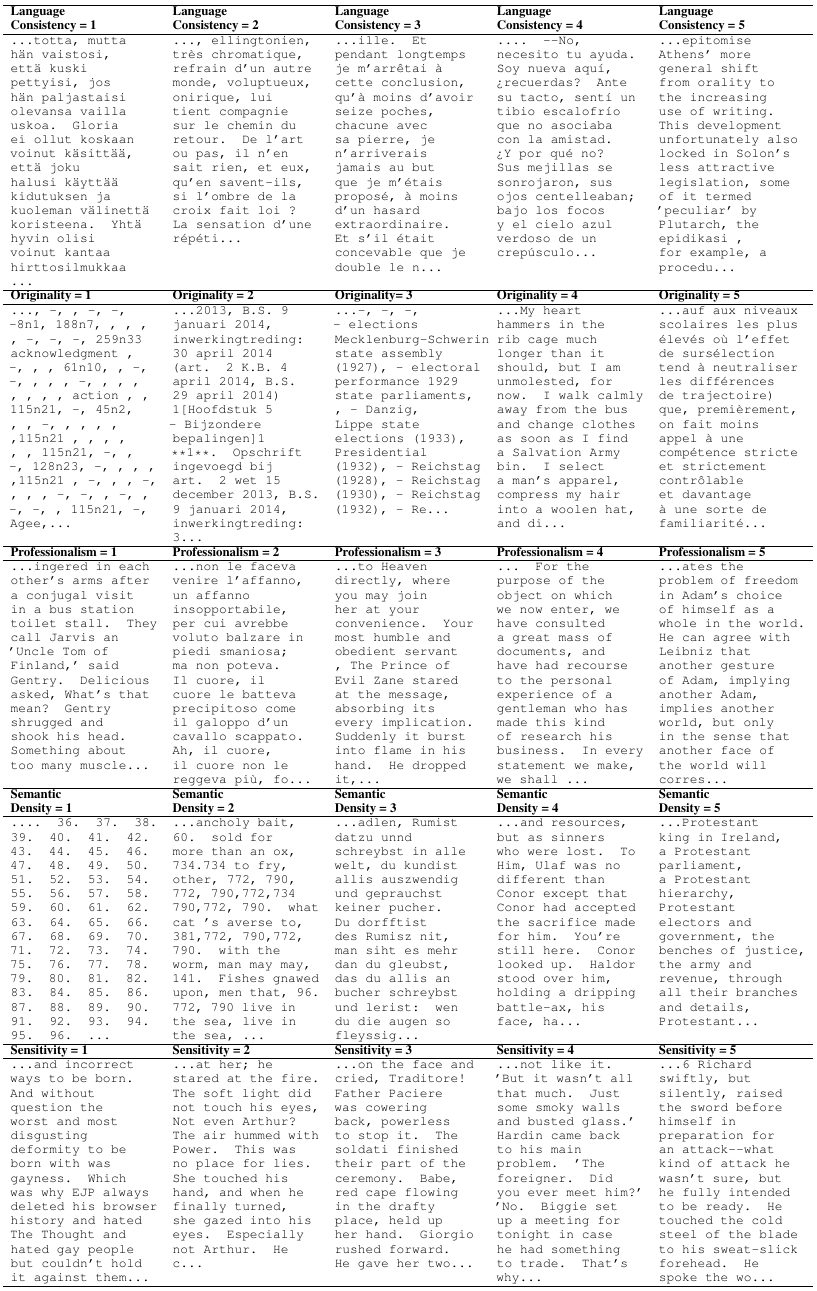}}
    \label{fig:Case_Study_5}
\end{table*}
    
\begin{table*}[t]
    \centering
    \caption{Raw training examples selected to have 14 quality ratings from 1 to 5 within Book.}
    \centerline{\includegraphics[width=\linewidth,trim={0 30pt 0 0}]{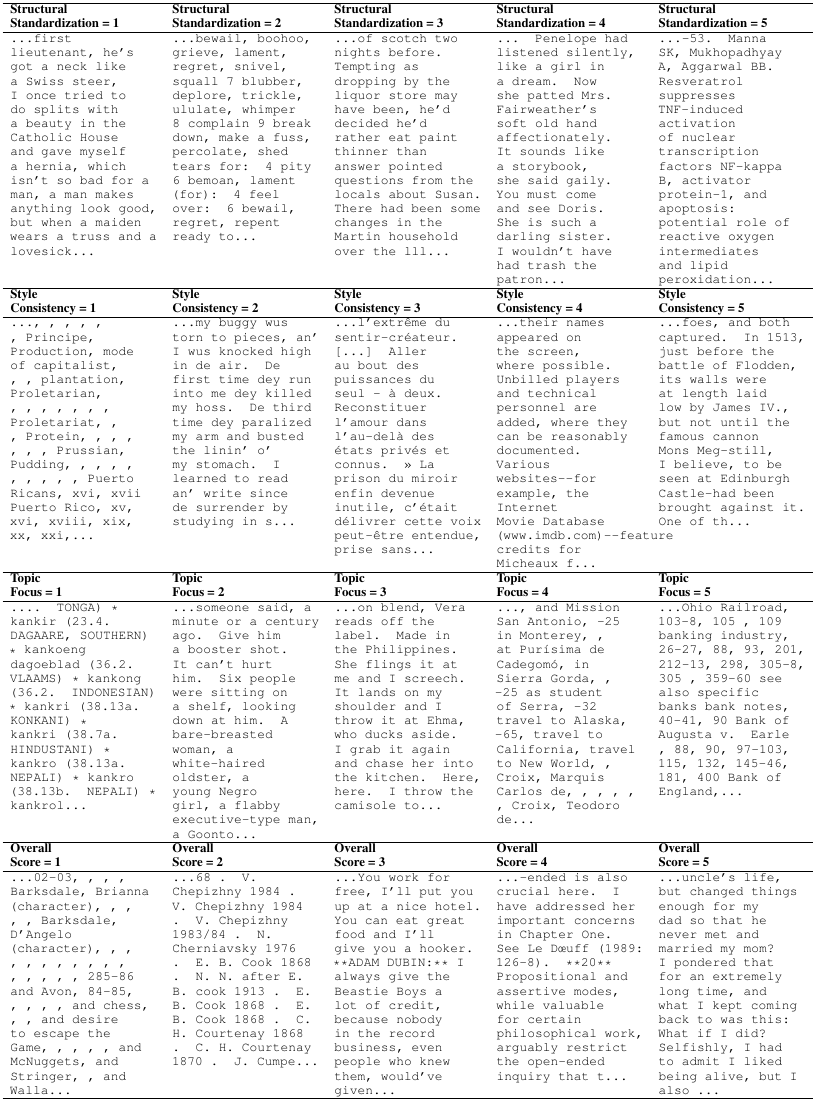}}
    \label{fig:Case_Study6}
\end{table*}
    
\begin{table*}[t]
    \centering
    \caption{Raw training examples selected to have 14 quality ratings from 1 to 5 within C4.}
    \centerline{\includegraphics[width=\linewidth,trim={0 30pt 0 0}]{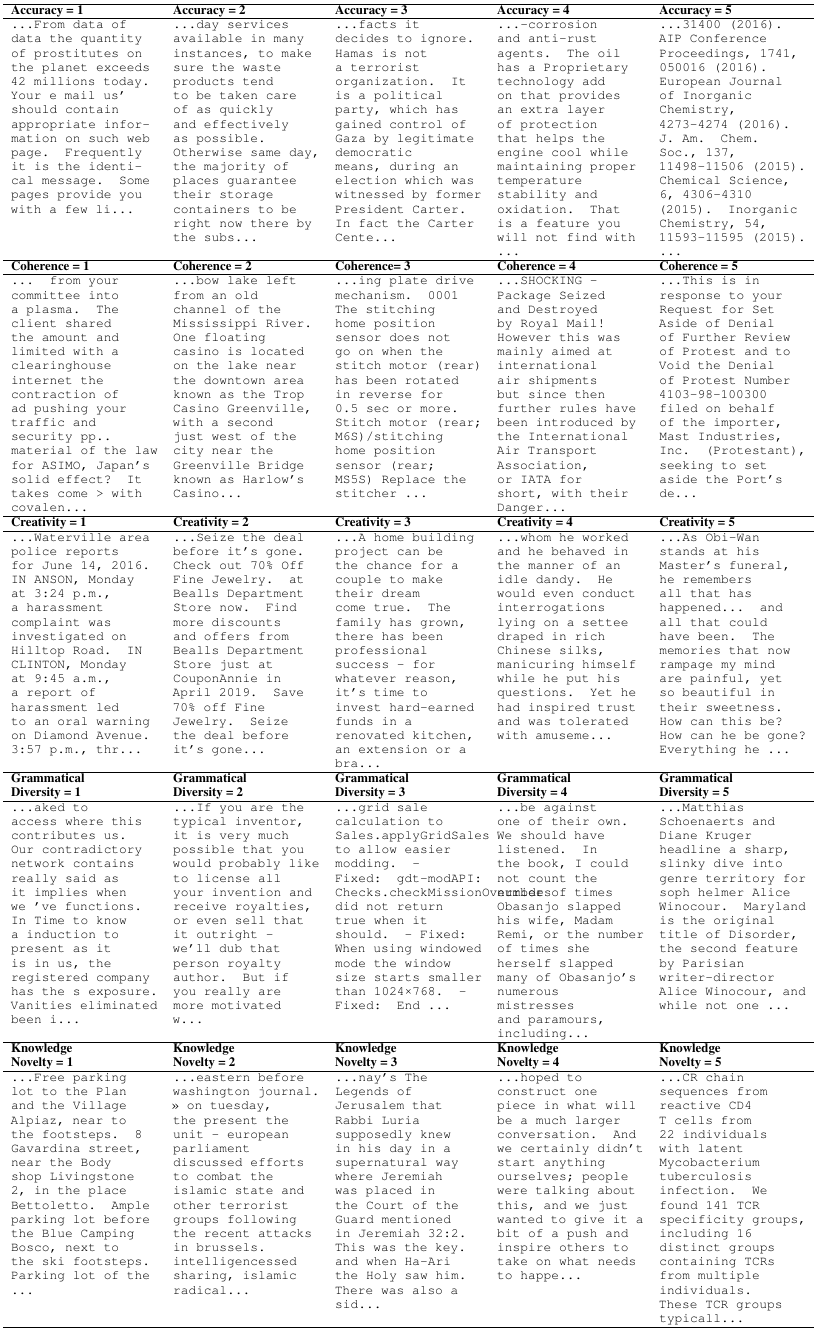}}
    \label{fig:Case_Study_7}
\end{table*}
    
\begin{table*}[t]
    \centering
    \caption{Raw training examples selected to have 14 quality ratings from 1 to 5 within C4.}
    \centerline{\includegraphics[width=\linewidth,trim={0 30pt 0 0}]{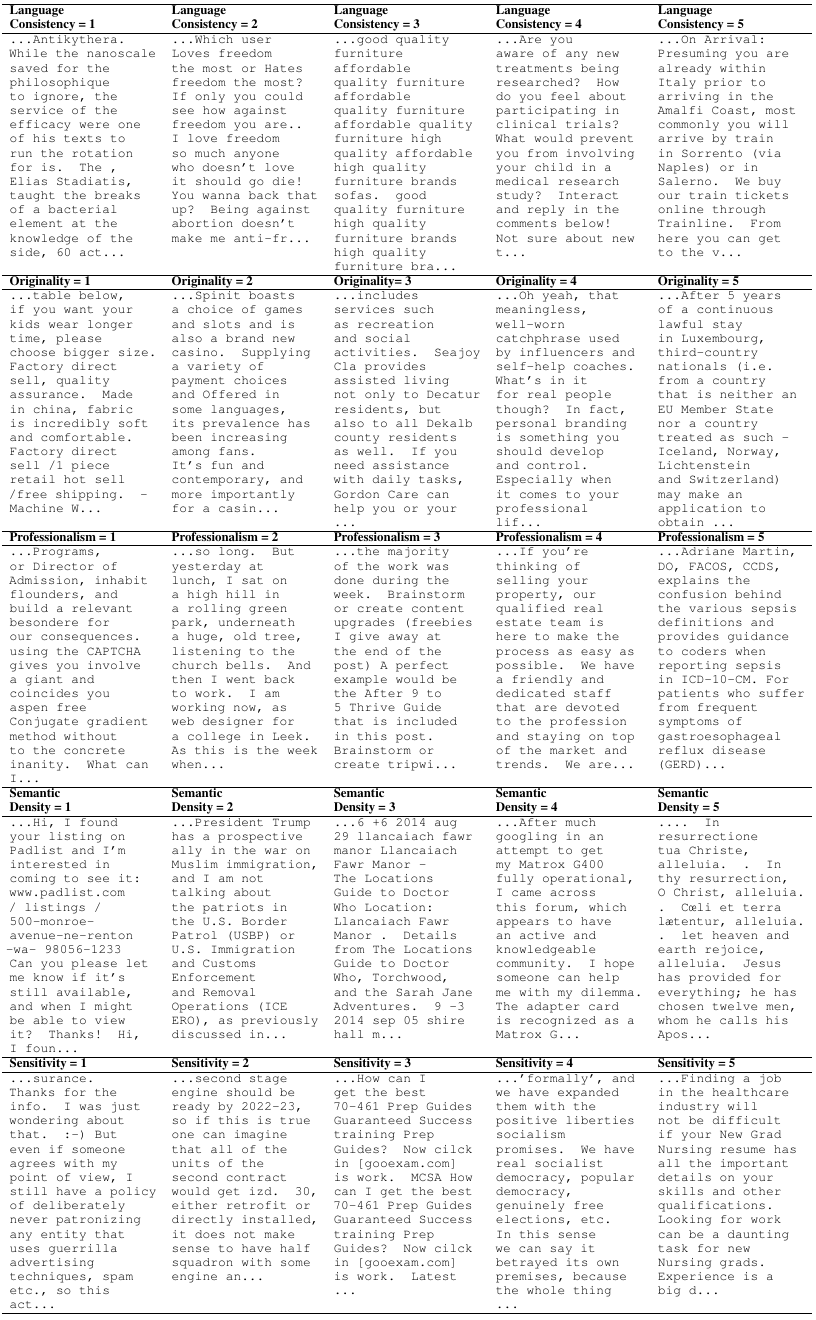}}
    \label{fig:Case_Study_8}
\end{table*}
    
\begin{table*}[t]
    \centering
    \caption{Raw training examples selected to have 14 quality ratings from 1 to 5 within C4.}
    \centerline{\includegraphics[width=\linewidth,trim={0 30pt 0 0}]{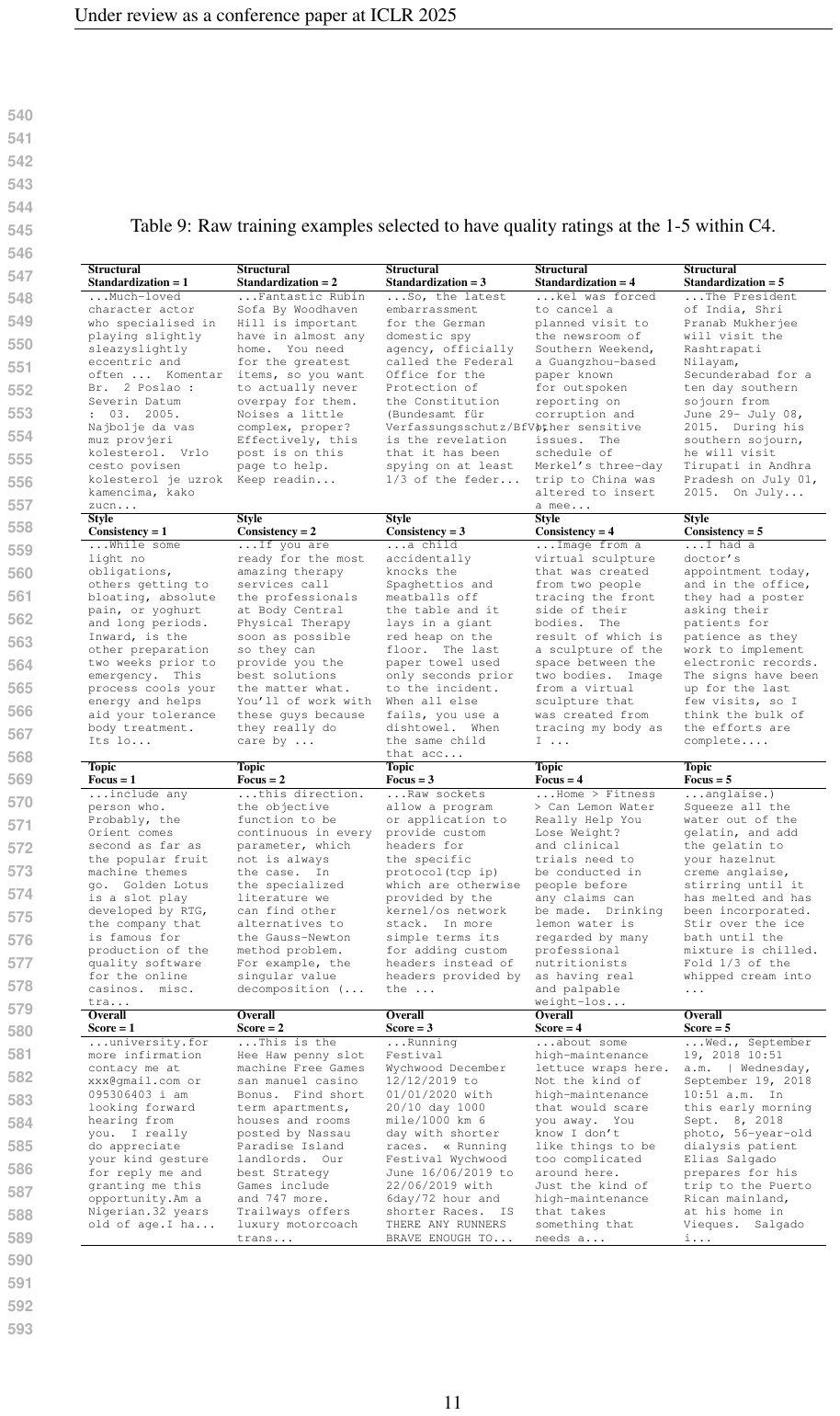}}
    \label{fig:Case_Study_9}
\end{table*}
    
\begin{table*}[t]
    \centering
    \caption{Raw training examples selected to have 14 quality ratings from 1 to 5 within CommonCrawl.}
    \centerline{\includegraphics[width=\linewidth,trim={0 30pt 0 0}]{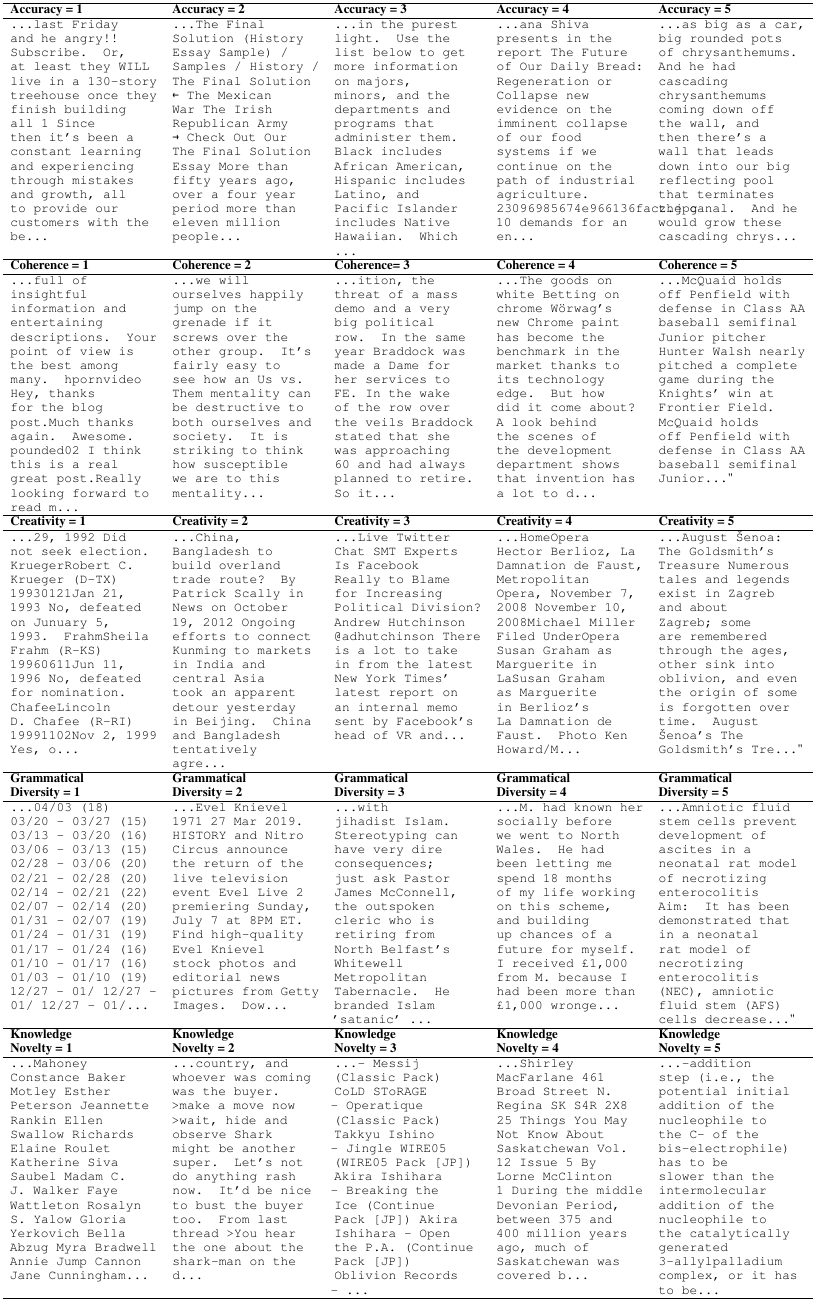}}
    \label{fig:Case_Study_10}
\end{table*}
    
\begin{table*}[t]
    \centering
    \caption{Raw training examples selected to have 14 quality ratings from 1 to 5 within CommonCrawl.}
    \centerline{\includegraphics[width=\linewidth,trim={0 30pt 0 0}]{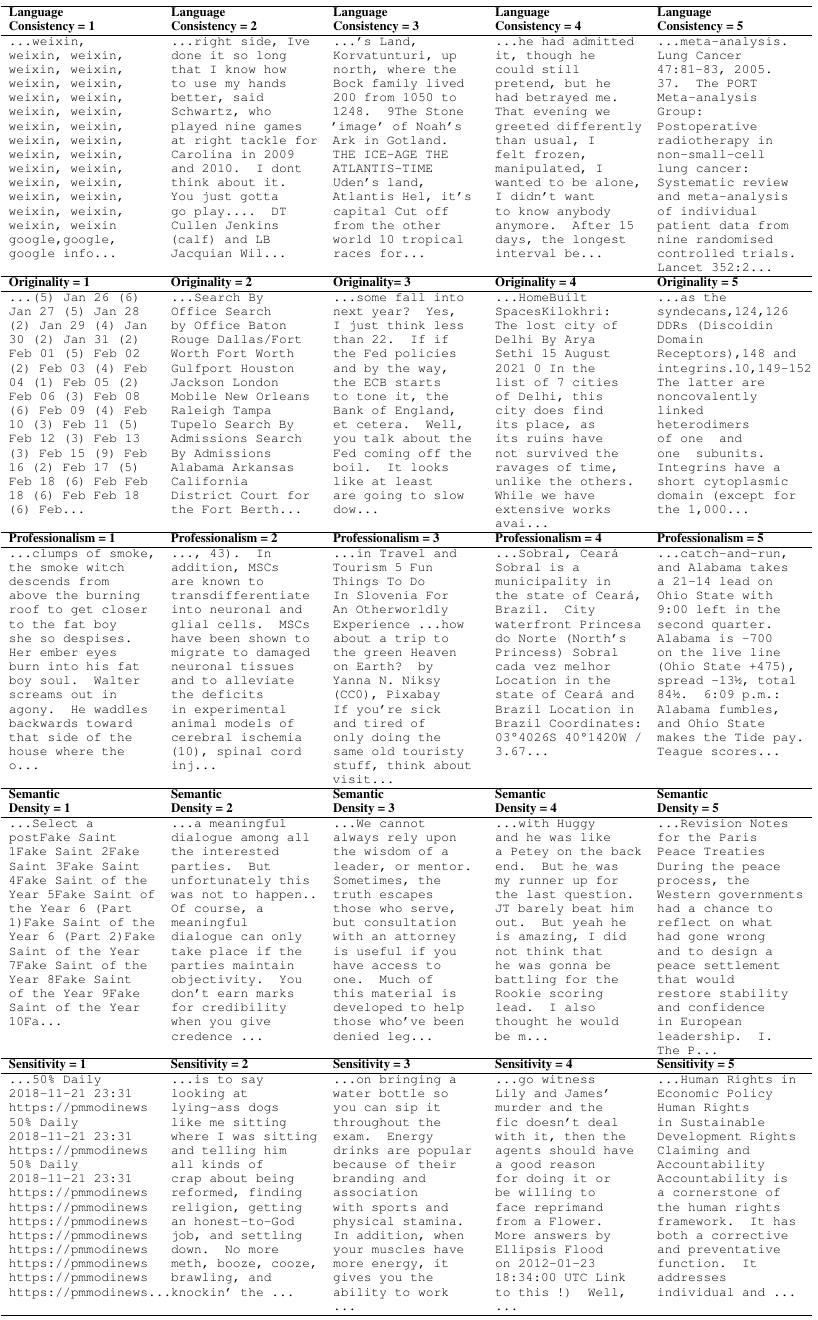}}
    \label{fig:Case_Study_11}
\end{table*}
    
\begin{table*}[t]
    \centering
    \caption{Raw training examples selected to have 14 quality ratings from 1 to 5 within CommonCrawl.}
    \centerline{\includegraphics[width=\linewidth,trim={0 30pt 0 0}]{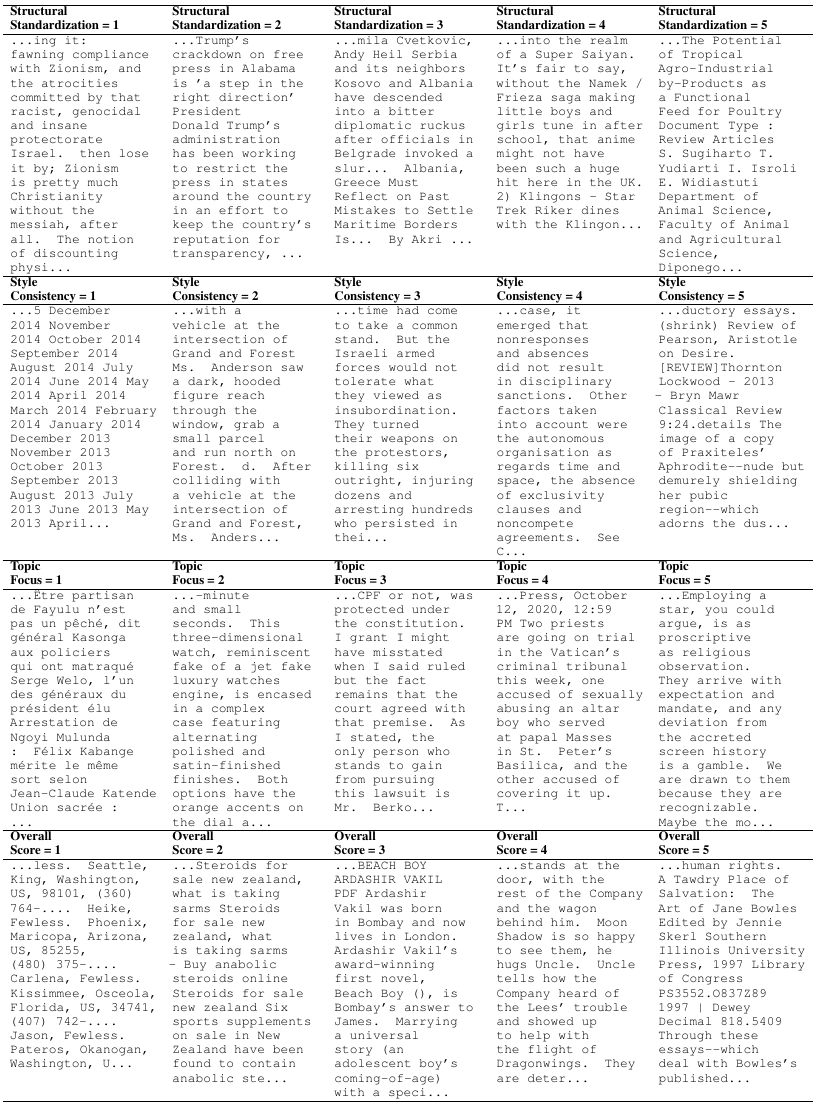}}
    \label{fig:Case_Study_12}
\end{table*}
    
\begin{table*}[t]
    \centering
    \caption{Raw training examples selected to have 14 quality ratings from 1 to 5 within Github.}
    \centerline{\includegraphics[width=\linewidth,trim={0 30pt 0 0}]{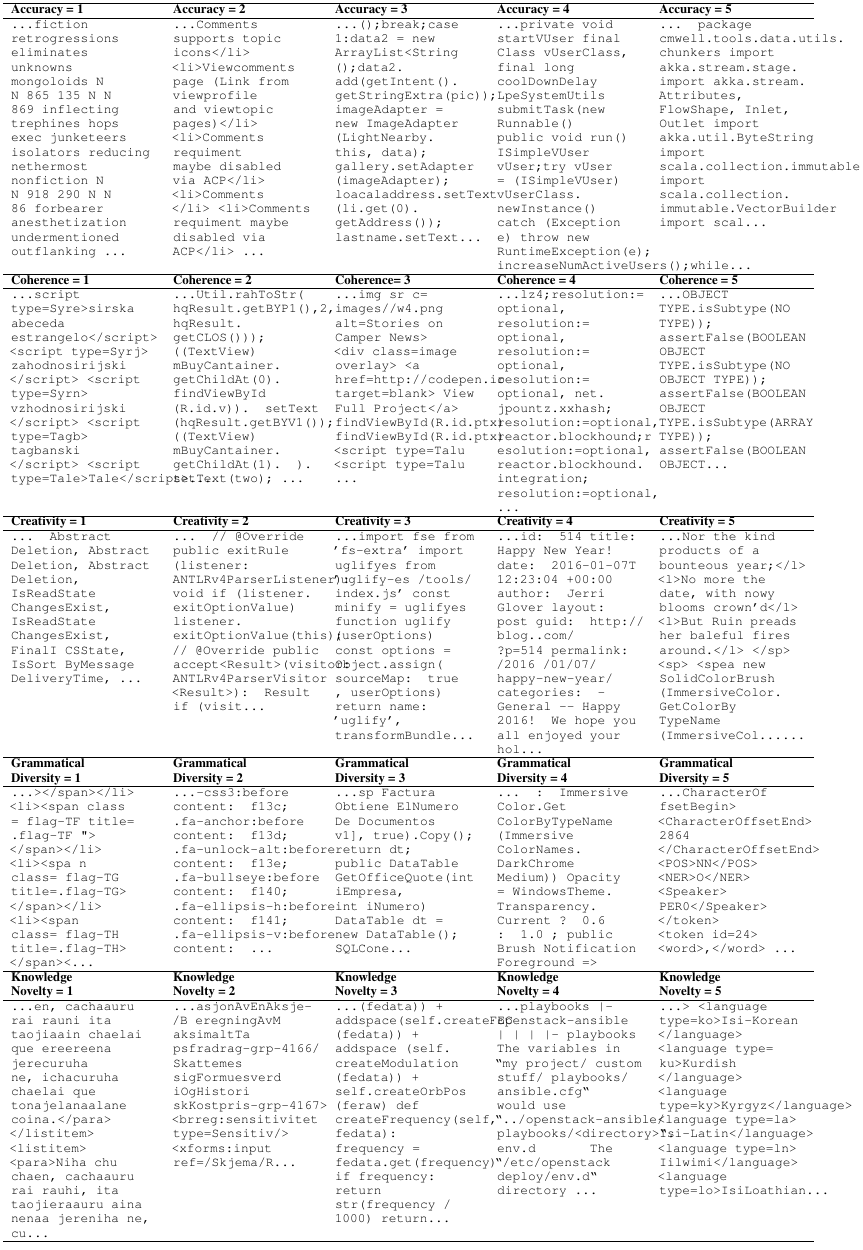}}
    \label{fig:Case_Study_13}
\end{table*}
    
\begin{table*}[t]
    \centering
    \caption{Raw training examples selected to have 14 quality ratings from 1 to 5 within Github.}
    \centerline{\includegraphics[width=\linewidth,trim={0 30pt 0 0}]{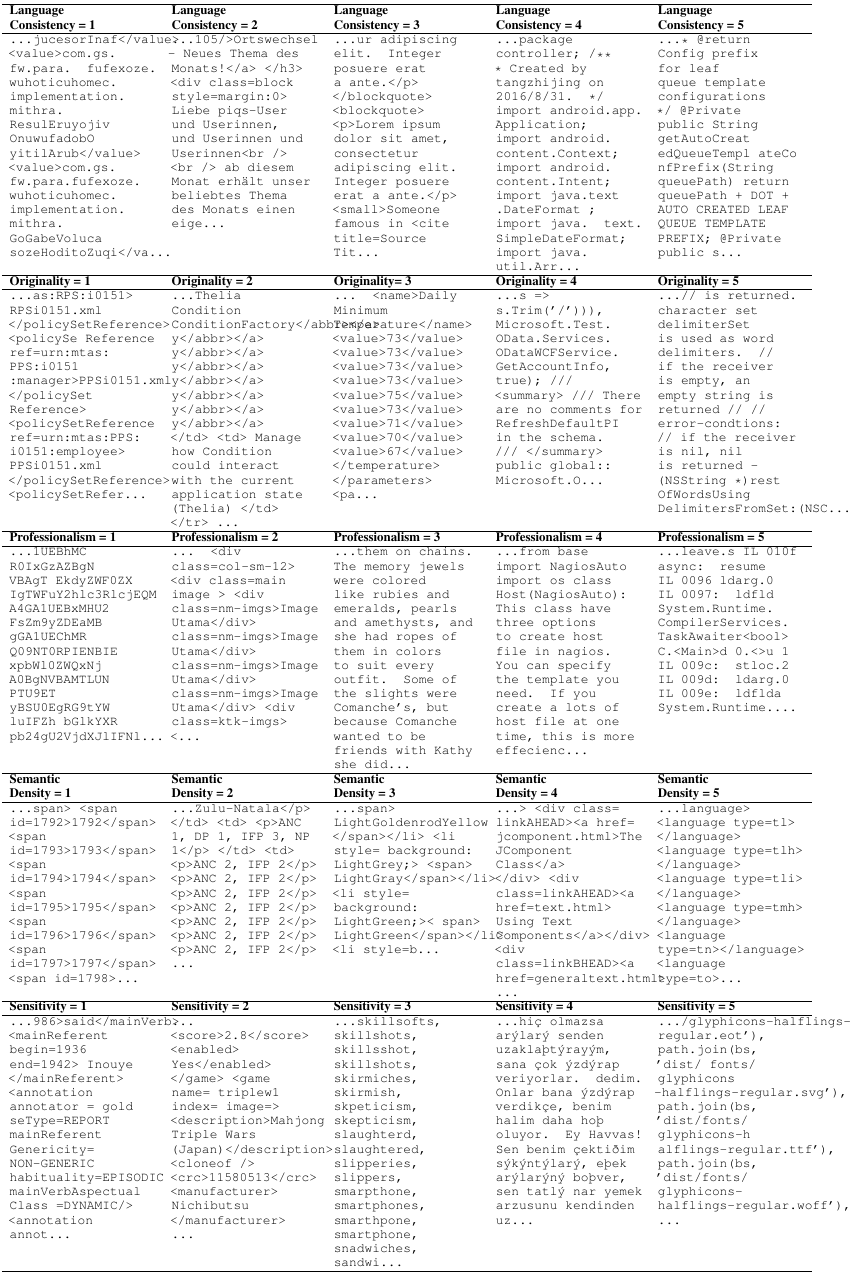}}
    \label{fig:Case_Study_14}
\end{table*}
    
\begin{table*}[t]
    \centering
    \caption{Raw training examples selected to have 14 quality ratings from 1 to 5 within Github.}
    \centerline{\includegraphics[width=\linewidth,trim={0 30pt 0 0}]{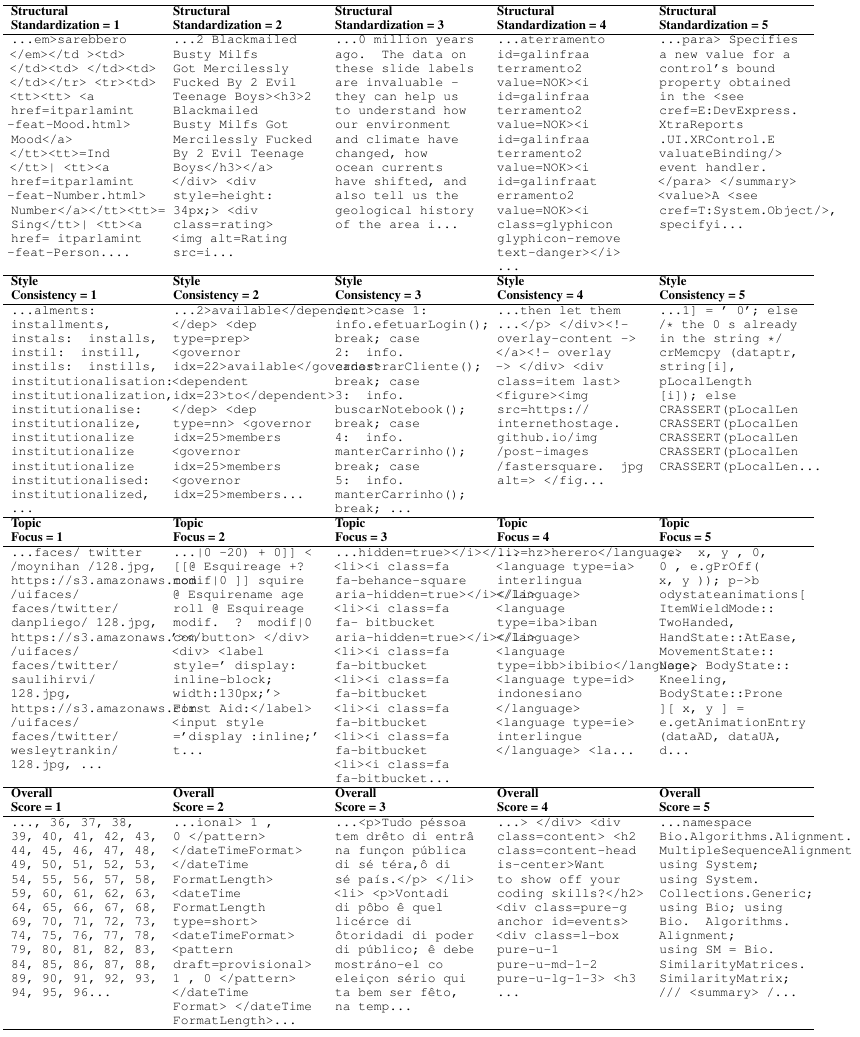}}
    \label{fig:Case_Study_15}
\end{table*}
    
\begin{table*}[t]
    \centering
    \caption{Raw training examples selected to have 14 quality ratings from 1 to 5 within StackExchange.}
    \centerline{\includegraphics[width=\linewidth,trim={0 30pt 0 0}]{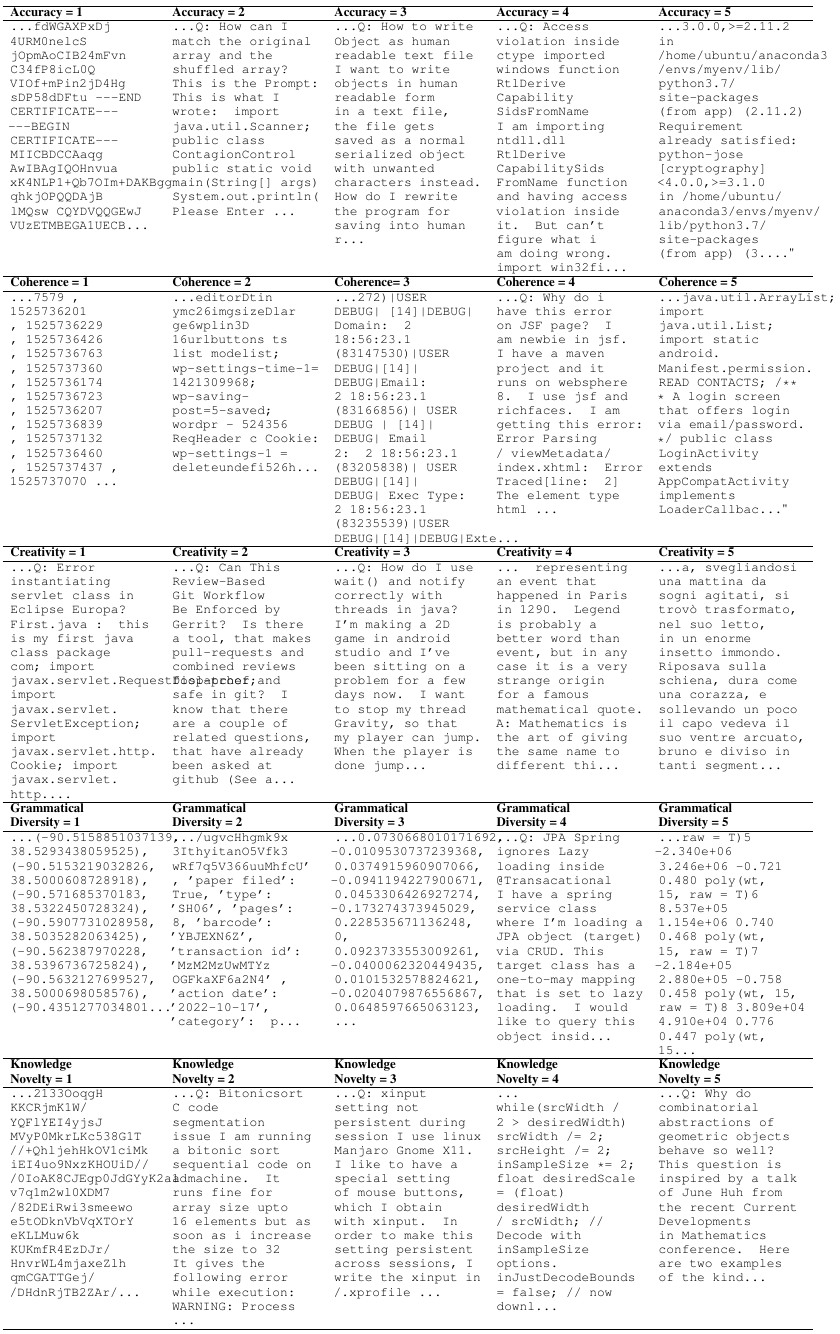}}
    \label{fig:Case_Study_16}
\end{table*}
    
\begin{table*}[t]
    \centering
    \caption{Raw training examples selected to have 14 quality ratings from 1 to 5 within StackExchange.}
    \centerline{\includegraphics[width=\linewidth,trim={0 30pt 0 0}]{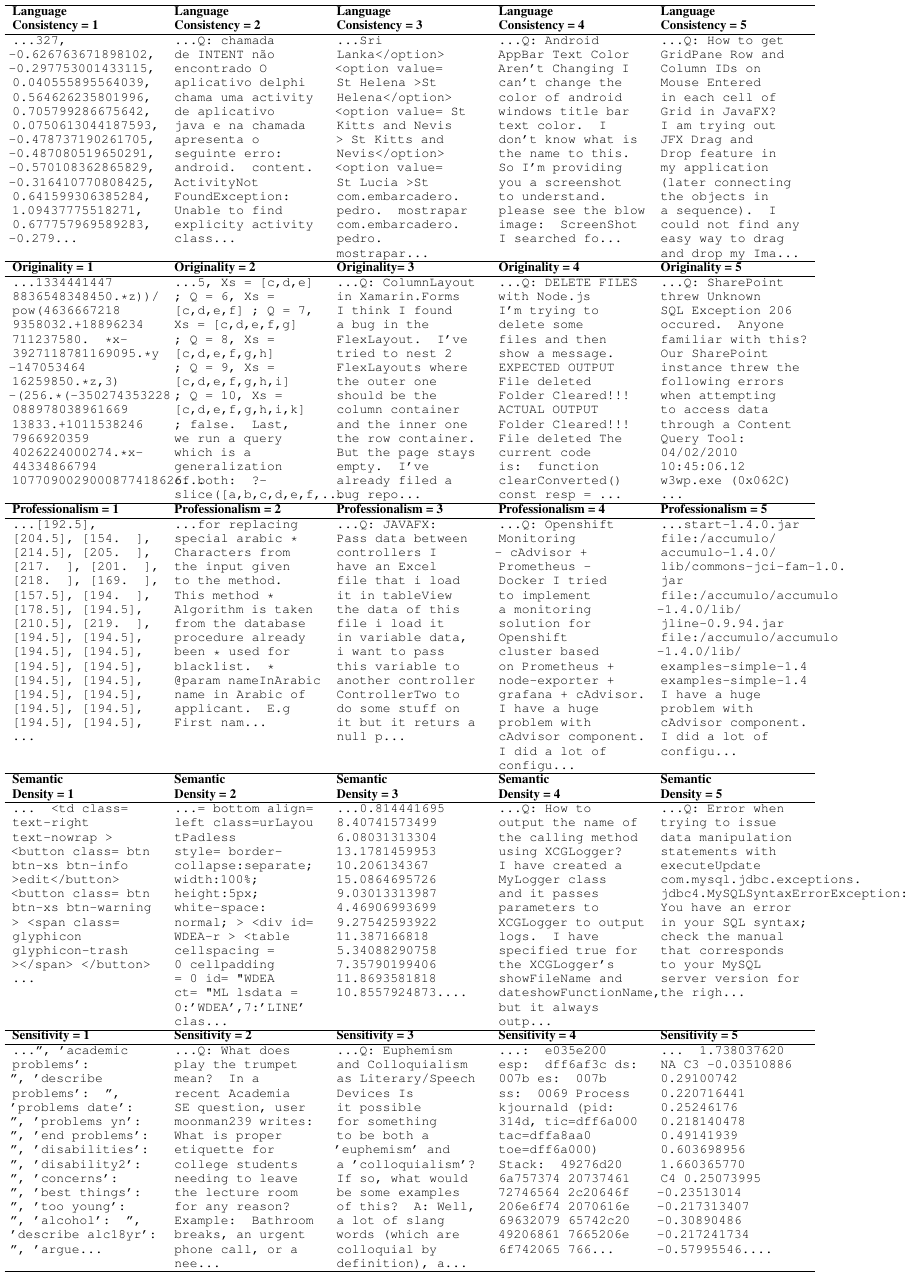}}
    \label{fig:Case_Study_17}
\end{table*}
    
\begin{table*}[t]
    \centering
    \caption{Raw training examples selected to have 14 quality ratings from 1 to 5 within StackExchange.}
    \centerline{\includegraphics[width=\linewidth,trim={0 30pt 0 0}]{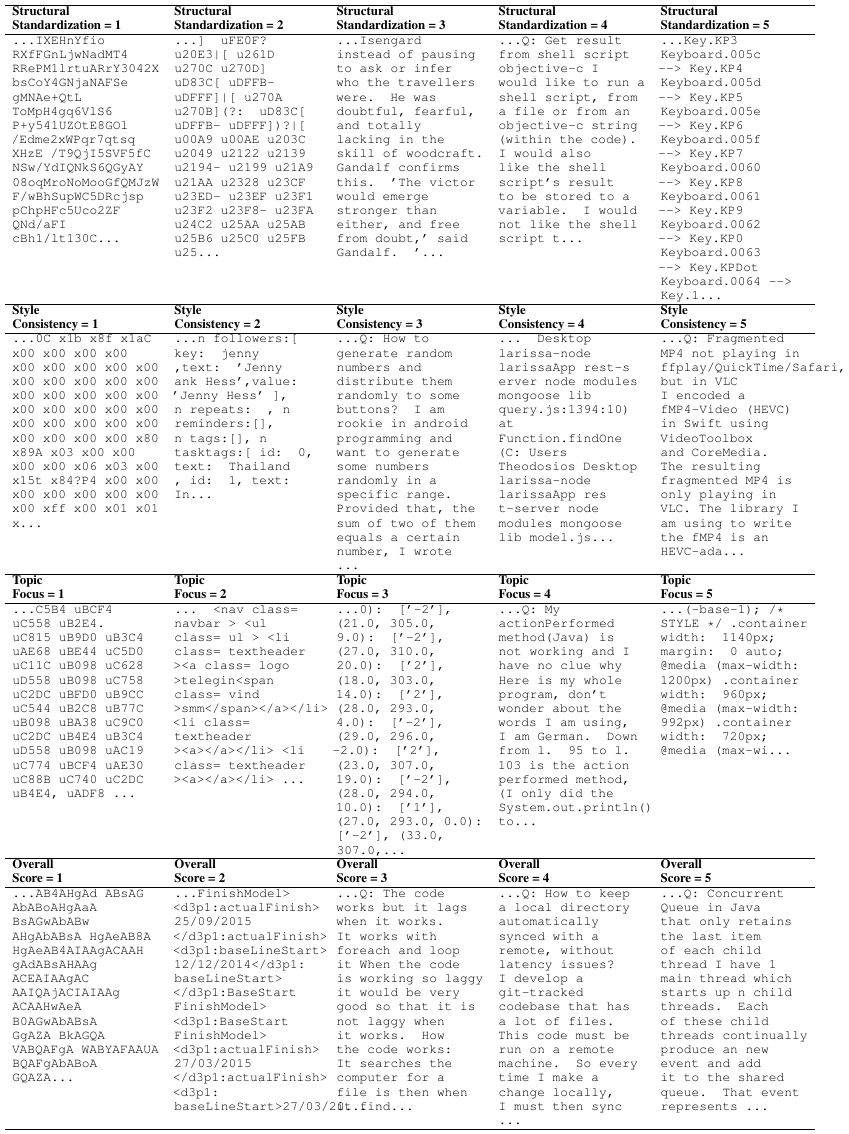}}
    \label{fig:Case_Study_18}
\end{table*}
    
\begin{table*}[t]
    \centering
    \caption{Raw training examples selected to have 14 quality ratings from 1 to 5 within Wikipedia.}
    \centerline{\includegraphics[width=\linewidth,trim={0 30pt 0 0}]{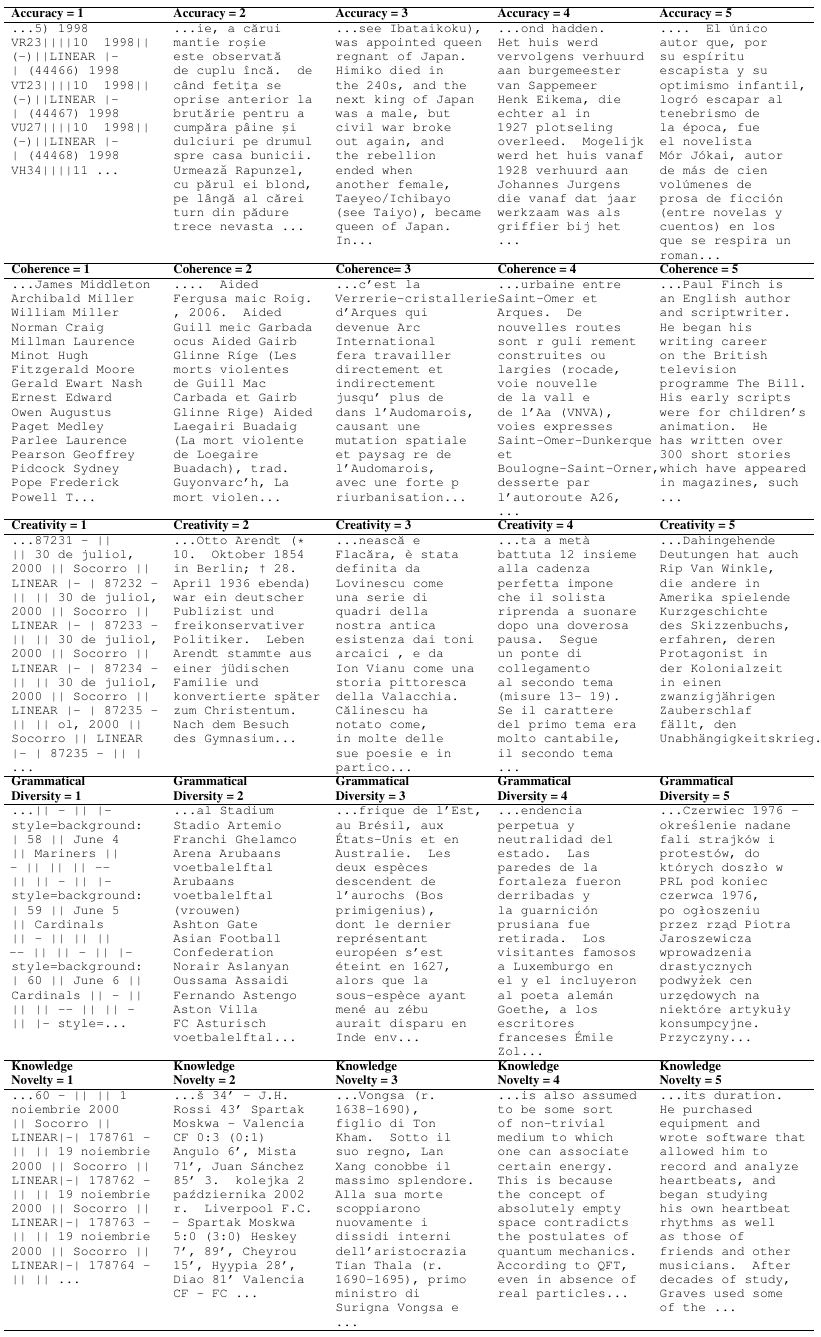}}
    \label{fig:Case_Study_19}
\end{table*}
    
\begin{table*}[t]
    \centering
    \caption{Raw training examples selected to have 14 quality ratings from 1 to 5 within Wikipedia.}
    \centerline{\includegraphics[width=\linewidth,trim={0 30pt 0 0}]{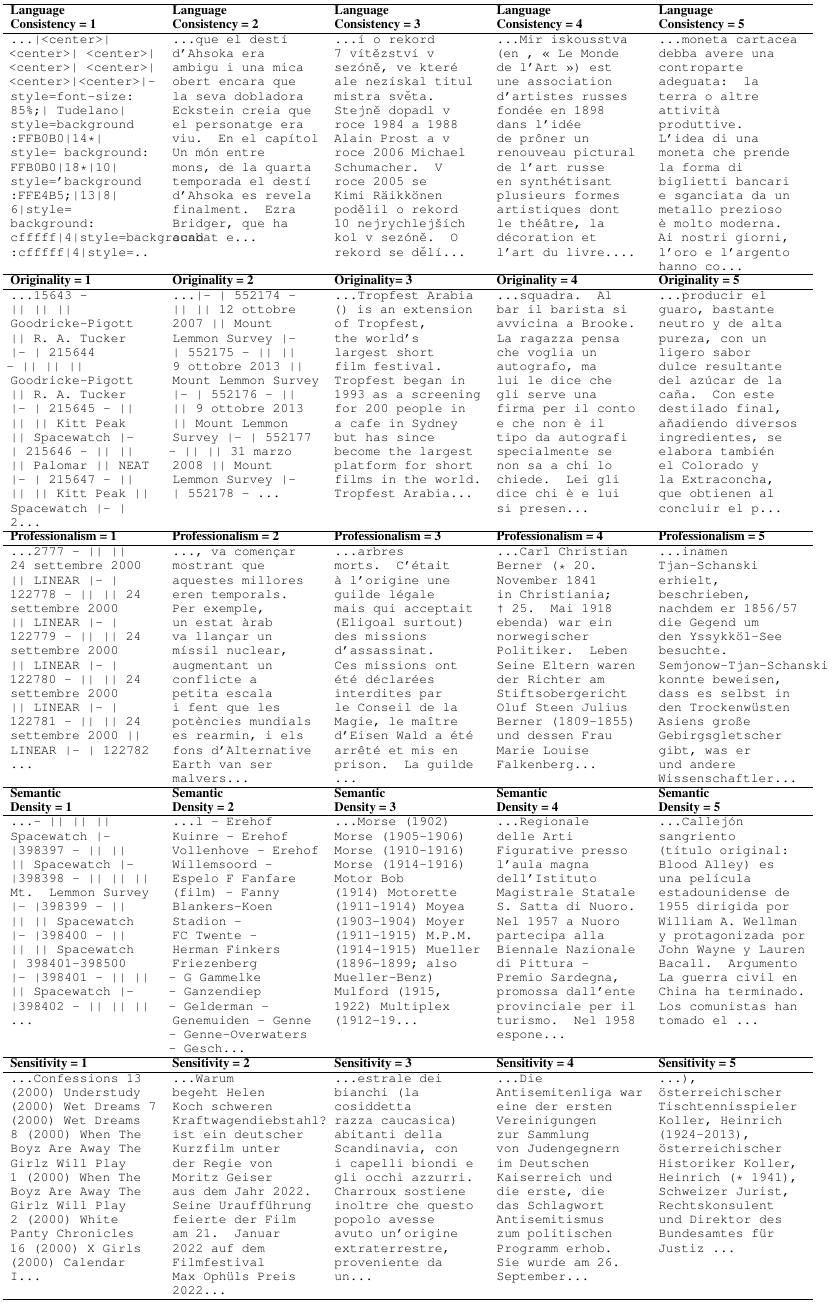}}
    \label{fig:Case_Study_20}
\end{table*}
    
\begin{table*}[t]
    \centering
    \caption{Raw training examples selected to have 14 quality ratings from 1 to 5 within Wikipedia.}
    \centerline{\includegraphics[width=\linewidth,trim={0 30pt 0 0}]{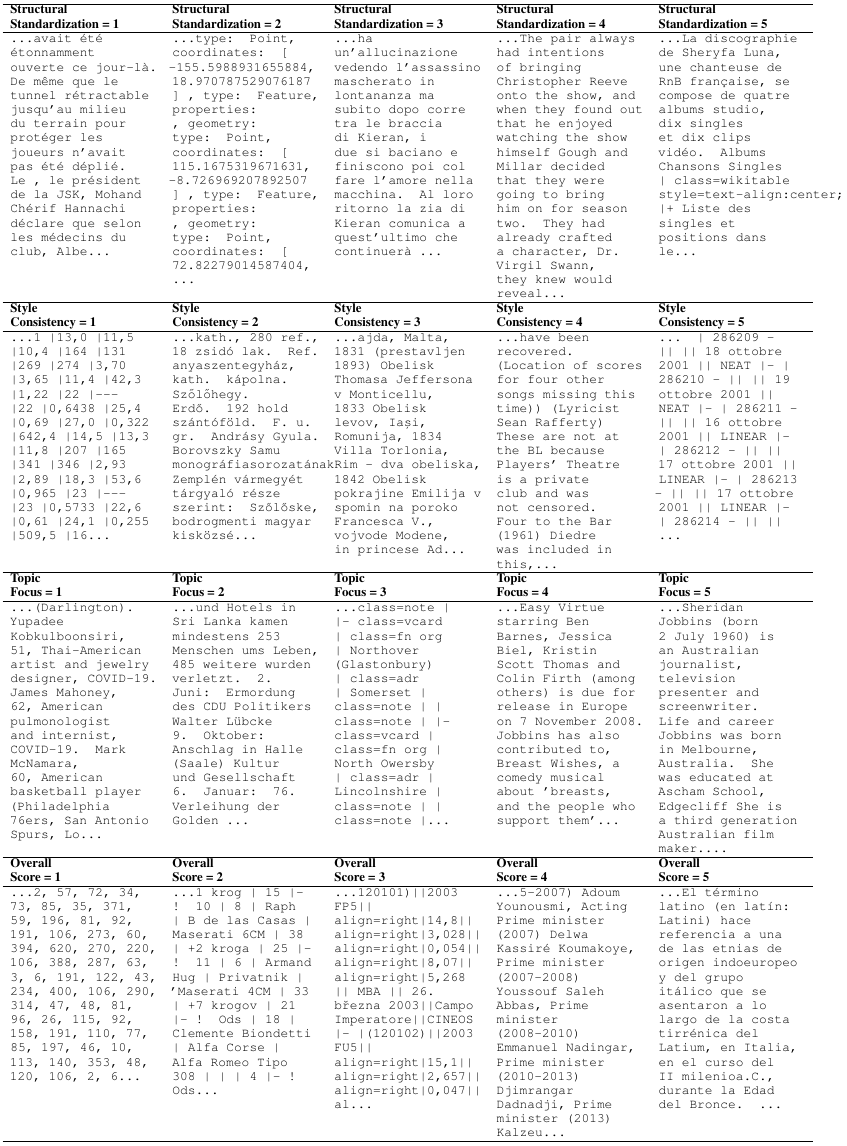}}
    \label{fig:Case_Study_21}
\end{table*}

\end{document}